\documentclass[review]{elsarticle}

\usepackage{lineno}
\modulolinenumbers[5]

\usepackage{amsmath,amssymb,amsfonts,bm}
\usepackage{caption}
\usepackage{multirow}
\usepackage{subfigure}
\usepackage{booktabs}
\usepackage[ruled]{algorithm2e}

\usepackage[colorlinks, linkcolor=blue, anchorcolor=blue, citecolor=blue]{hyperref}
\usepackage{geometry}
\journal{Journal}

\begin{document}
\captionsetup[figure]{labelfont={bf},labelformat={default},labelsep=period,name={Fig.}}

\begin{frontmatter}

\title{Deep Adaptive Arbitrary Polynomial Chaos Expansion: A Mini-data-driven Semi-supervised Method for Uncertainty Quantification}

\author[mymainaddress,mysecondaryaddress]{Wen Yao}

\author[mymainaddress,mysecondaryaddress]{Xiaohu Zheng\corref{mycorrespondingauthor}}
\cortext[mycorrespondingauthor]{Corresponding author}
\ead{zhengboy320@163.com}

\author[mysecondaryaddress]{Jun Zhang}

\author[mysecondaryaddress]{Ning Wang}

\author[mysecondaryaddress]{Guijian Tang}

\address[mymainaddress]{College of Aerospace Science and Engineering, National University of Defense Technology, No. 109, Deya Road, Changsha 410073, China}
\address[mysecondaryaddress]{Defense Innovation Institute, Chinese Academy of Military Science, No. 53, Fengtai East Street, Beijing 100071, China}

\begin{abstract}
The surrogate model-based uncertainty quantification method has drawn much attention in many engineering fields. Polynomial chaos expansion (PCE) and deep learning (DL) are powerful methods for building a surrogate model. However, PCE needs to increase the expansion order to improve the accuracy of the surrogate model, which causes more labeled data to solve the expansion coefficients, and DL also requires a lot of labeled data to train the deep neural network (DNN). First of all, this paper proposes the adaptive arbitrary polynomial chaos (aPC) and proves two properties about the adaptive expansion coefficients. Based on the adaptive aPC, a semi-supervised deep adaptive arbitrary polynomial chaos expansion (Deep aPCE) method is proposed to reduce the training data cost and improve the surrogate model accuracy. For one hand, the Deep aPCE method uses two properties of the adaptive aPC to assist in training the DNN based on only a small amount of labeled data and many unlabeled data, significantly reducing the training data cost. On the other hand, the Deep aPCE method adopts the DNN to fine-tune the adaptive expansion coefficients dynamically, improving the Deep aPCE model accuracy with lower expansion order. Besides, the Deep aPCE method can directly construct accurate surrogate models of the high dimensional stochastic systems without complex dimension-reduction and model decomposition operations. Five numerical examples and an actual engineering problem are used to verify the effectiveness of the Deep aPCE method.
\end{abstract}


\end{frontmatter}

\linenumbers

\section{Introduction}\label{sec1}
In many engineering fields like mechanical engineering\cite{Lee2020, Liu2021}, aerospace engineering\cite{Yao2011, Yao2013}, electric power engineering\cite{Sun2021Machine}, and civil engineering\cite{Xu2020, Zhang2021, Hong2021Reliability}, some uncertainty factors such as environmental changes and manufacturing defects of mechanical structures will lead to uncertainties in the physical parameters, external loads, boundary conditions\cite{Zhang2021}, etc. The above uncertainties will significantly influence the performance of the system. Therefore, uncertainty quantification is essential for analyzing the effects of uncertainties on system performance and minimizing system failure risk\cite{Zheng2019, Zheng2020, Fan2018Uncertainty}. 

In recent years, the uncertainty quantification methods mainly include the sampling-based simulation method (e.g., Monte Carlo simulation \cite{Rubinstein2016}), the surrogate model-based method (e.g., polynomial chaos expansion \cite{Xiu2003}), the Taylor series expansion-based method (e.g., the first-order and second-order reliability method \cite{Yao2019}), etc. For the surrogate model-based method, it constructs a simple and explicit surrogate model to replace the original complicated and computationally expensive model like finite element analysis (FEA). Once the explicit surrogate model is constructed, the researcher can straightforwardly perform the Monte Carlo simulation (MCS) on the surrogate model with a lower computational cost. Then, any probabilistic quantity of interest can be obtained according to the results of MCS. Thus, some surrogate model-based methods like the polynomial chaos expansion (PCE) \cite{Oladyshkin2012, Wan2020}, deep learning method \cite{Tripathy2018, Zhang2019}, response surface method \cite{Wong1985, Roussouly2013}, Kriging method \cite{Kaymaz2005, Echard2011}, support vector machine \cite{Li2006, Bourinet2011} have drawn a lot of attention for quantifying uncertainty in recent years. By expanding the stochastic system's output response onto a basis composed of orthogonal polynomials, PCE becomes a versatile and powerful method for building a surrogate model.

PCE expands the stochastic system model into a polynomial model based on the polynomial orthogonal basis of random input variables. By employing the Hermite polynomials, Wiener \cite{Wiener1938} introduced the original Wiener–Hermite expansion firstly in 1938. The original Wiener–Hermite expansion is a very effective method for the stochastic system model with normal random input variables. However, if the stochastic system model's random input variables are non-normal variables, the original Wiener–Hermite expansion shows a prolonged convergence rate. To solve this problem, Xiu et al. \cite{Xiu2003} proposed generalized polynomial chaos (gPC) by employing orthogonal polynomial functional from the Askey scheme. The gPC makes PCE suitable for the stochastic system model whose random input variables obey parametric statistical distributions like Gamma, Beta, Uniform, etc. Unfortunately, information about data distribution is very limited in practical engineering applications, which will make the parametric statistical distributions of random input variables unknown. Thereby, the gPC is not suitable for this situation. To make PCE suitable for a larger spectrum of distributions, Oladyshkin et al. \cite{Oladyshkin2012} proposed the arbitrary polynomial chaos expansion (aPC) method based on a finite number of moments of random input variables. The main feature of aPC method is that it allows the researchers to select freely technical constraints the shapes of their statistical assumptions. After constructing the PCE model by the above three methods, the expansion coefficients can be solved by Galerkin projection method \cite{Xiu2003, Ghanem1993, Matthies2005}, collocation method \cite{Isukapalli1998, Li2007, Shi2009}, etc. \textbf{Generally, the PCE model's accuracy can be improved by increasing the expansion order. However, this increases the number of samples for solving the expansion coefficients, which will result in higher computational costs in constructing the PCE model.}

Besides, the PCE method suffers the problem known as the ‘curse of dimensionality’ for high dimensional stochastic systems, i.e., the number of expansion coefficients increases dramatically with the dimension of random input variables. To overcome this problem, Xu et al. \cite{Xu2020} decomposed the original performance function to be a summation of several component functions by contribution-degree analysis, and then the gPC is employed to reconstruct these component functions. Pan et al. \cite{Pan2017} adopted the sliced inverse regression technique to achieve a dimension reduction, then the sparse PCE was used to construct the surrogate model. Blatman et al. \cite{Blatman2010} proposed an adaptive regression-based algorithm that can automatically detect the significant coefficients of the PCE. As a consequence, a relatively small number of expansion coefficients is eventually retained. Besides, an adaptive algorithm based on least angle regression was also proposed for automatically detecting the significant coefficients of PCE by Blatman et al. \cite{Blatman2011}. \textbf{For the dimension-reduction steps or the significant coefficients detecting in the above methods, however, it complicates the process of building a PCE model and will affect the accuracy of the PCE model.}

To simulate a high dimensional stochastic system, the deep learning (DL) \cite{Schmidhuber2015} is applied to construct a surrogate model, and the deep neural network (DNN) is a classical DL model \cite{Goodfellow2016}. According to the researches by Hornik et al. \cite{Hornik1989} and Cybenko \cite{Cybenko1989}, the surrogate model with any desired degree of accuracy can be obtained by DNN. Thus, DNN can handle the 'curse of dimensionality' by multiple nonlinear activation functions \cite{Shahane2019}. In recent years, DNN has been studied in many fields for constructing surrogate models \cite{Shahane2019, Czl2014, Lye2021, ZhangXin2021}. By dividing the hot wall into multiple domains, Shahane et al. \cite{Shahane2019} used deterministic forwarding simulations to train DNN. Based on DNN, Lye et al. \cite{Lye2021} proposed an iterative surrogate model optimization method whose key feature is the iterative selection of training data. By blending different fidelity information, Zhang et al. \cite{ZhangXin2021} constructed a high-accuracy multi-fidelity surrogate model for designing aircraft aerodynamic shapes based on DNN. Refer to the above methods, DNN can build a high-precision surrogate model. \textbf{However, DL's main defect is that a large number of labeled training data need to be used to train DNN, which will lead to unacceptable computational costs.} To solve this problem, many researchers used some physical knowledge to train DNN. For example, Raissi et al. \cite{Raissi2019} proposed the physics-informed neural network to solve the forward and inverse problems involved in nonlinear partial differential equations. Zheng et al. \cite{Zheng2022,Gongzq2021} used the Laplace equation and the boundary condition to train the DNN for reconstructing the temperature field of the satellite system. Combined with the aPC, Zhang et al. \cite{Zhang2019} adopted the physics-informed neural network to construct the NN-aPC model. \textbf{However, some physical knowledge, like the partial differential equation, etc., is usually hard to acquire for some systems.}

According to the above analysis, PCE and DL have the following three shortcomings: Firstly, the PCE model's accuracy can be improved by increasing the expansion order, but the computational cost will increase accordingly; Secondly, for higher dimensional stochastic systems, the existing PCE methods not only complicate the PCE modeling process but also affect the PCE model's accuracy; Thirdly, the DL can construct a high-precision surrogate model but needs a lot of labeled training data to train DNN. \textbf{To overcome the above three shortcomings, this paper proposes the adaptive arbitrary polynomial chaos (aPC) and proves two properties about the adaptive expansion coefficients. Based on the adaptive aPC and the DNN, this paper proposes a semi-supervised deep arbitrary polynomial chaos expansion (Deep aPCE) method to improve the PCE model's accuracy and reduce the number of samples for solving the expansion coefficients. The proposed Deep aPCE method adopts a small amount of labeled training data and many unlabeled training data to learn the shared parameters of the Deep aPCE model and the DNN. Compared with the existing PCE method, the proposed Deep aPCE method requires fewer calling times of computationally expensive models to build a higher accuracy surrogate model. What's more, due to the application of DL, the proposed Deep aPCE method can directly construct accurate surrogate models of the high dimensional stochastic systems without complex dimension-reduction and model decomposition operations.} After training the Deep aPCE model, the researchers can straightforwardly perform MCS on the Deep aPCE model to quantify the uncertainty of the stochastic system.

Unlike the traditional PCE methods \cite{Xiu2003, Oladyshkin2012, Wiener1938}, the proposed Deep aPCE method can fine-tune the adaptive expansion coefficients dynamically by the DNN for different random inputs. Besides, the NN-aPC method \cite{Zhang2019} also used the DNN to solve the expansion coefficients. Pay special attention, the technical route of the proposed Deep aPCE method is completely different from the NN-aPC method's technical route. The NN-aPC method mainly uses physical knowledge (partial differential equations and boundary conditions) to learn the parameters of DNN. It is apparent that the NN-aPC method firstly demands researchers to obtain the related physical knowledge of stochastic systems. Then, the NN-aPC method can be used to quantify uncertainty. Unfortunately, if the stochastic system is a black-box problem, the NN-aPC method will no longer apply. However, the proposed Deep aPCE method uses the properties (detailed in section \ref{sec32}) of adaptive aPC to assist in training the DNN, reducing the number of labeled training data. This point is essentially different from NN-aPC method. Although the stochastic system is a black box problem, the proposed Deep aPCE method can still quantify the uncertainty of the stochastic system. Therefore, the proposed Deep aPCE method of this paper is fundamentally different from NN-aPC method. 

In summary, the main innovations of this paper are presented in the following: \textbf{1) This paper proposes the adaptive aPC and proves two properties about the adaptive expansion coefficients; 2) This paper uses two properties of the adaptive aPC to assist in training the DNN, reducing the number of samples for solving the adaptive expansion coefficients; 3) The adaptive expansion coefficients dynamically fine-tuned by the DNN can improve the accuracy of the Deep aPCE model.} The rest of this paper is organized as follows. In section \ref{sec2}, the theoretical bases including DNN, aPC and error estimates are introduced. The adaptive aPC and its two properties are proposed in section \ref{sec3}. Then, the proposed Deep aPCE method are presented in section \ref{sec4}, including the proposed cost function and the model iterative training method. In section \ref{sec5}, five numerical examples are used to verify the effectiveness of the proposed Deep aPCE method. Finally, the proposed Deep aPCE method is applied to the uncertainty analysis for the first-order frequency of micro-satellite TianTuo-3 frame structure in section \ref{sec6}.

\section{Theoretical bases}\label{sec2}
\subsection{Deep neural network}\label{sec21}
\begin{figure}[htbp]
	\centering
	{\includegraphics[scale=1]{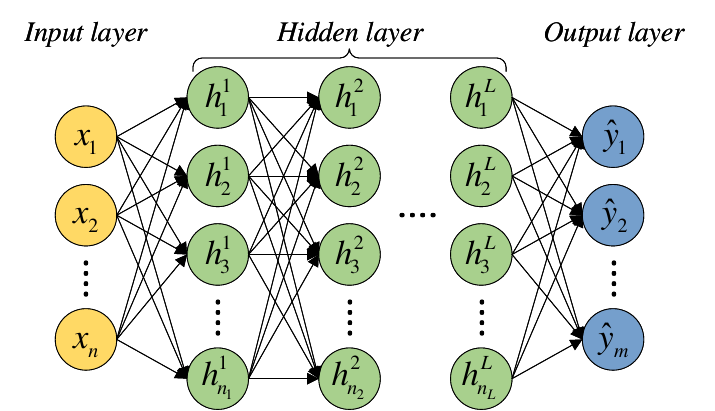}}
	\caption{A DNN model with $(L+2)$ layers}\label{DNN}
\end{figure}

\paragraph{\textbf{Definition of DNN}} DNN \cite{Goodfellow2016} consists of an input layer, multiple hidden layers, and an output layer. Except for the input nodes, each node is a neuron. Besides, each node in the hidden layer uses a nonlinear activation function \cite{Xu2015}. As shown in Fig.\ref{DNN}, it is a $(L+2)$ layers DNN.

According to the definition of DNN \cite{Goodfellow2016}, the values ${{{\bm{\hat{y}}}}_{1\times m}}$ of neurons in the output layer are calculated by
\begin{equation}\label{MLP_output}
\begin{aligned}
& {{\bm{H}}_{1\times {{n}_{1}}}}=g\left( {{\bm{x}}_{1\times n}}{{\bm{W}}_{n\times {{n}_{1}}}}+{{\bm{b}}_{1\times {{n}_{1}}}} \right), \\ 
& {{\bm{H}}_{1\times {{n}_{L}}}}=g\left\{ g\left[ \cdots g\left( {{\bm{H}}_{1\times {{n}_{1}}}}{{\bm{W}}_{{{n}_{1}}\times {{n}_{2}}}}+{{\bm{b}}_{1\times {{n}_{2}}}} \right)\cdots  \right]{{\bm{W}}_{{{n}_{L-1}}\times {{n}_{L}}}}+{{\bm{b}}_{1\times {{n}_{L}}}} \right\}, \\
& {{{\bm{\hat{y}}}}_{1\times m}}={{\bm{H}}_{1\times {{n}_{L}}}}{{\bm{W}}_{{{n}_{L}}\times m}}+{{\bm{b}}_{1\times m}}, \\ 
\end{aligned}
\end{equation}
where $g$ is the nonlinear activation function (eg. $ReLU(x)$, $GELU(x)$, $sigmoid(x)$, $tanh(x)$, etc.), $\bm{H}$ is the value of neuron in the hidden layer, $\bm{W}$ is the weight, and $\bm{b}$ is the bias.

The goal of DNN is to approximate some function $\bm{y}=f(\bm{x})$. Given the input $\bm x$, DNN will define a mapping $\hat{\bm{y}}=\hat{f}(\bm{x};\bm{\theta})$ that can approximate $\bm{y}=f(\bm{x})$ as accurately as possible by learning the parameters $\bm{\theta}=\left( \bm{W},\bm{b}\right)$.

\paragraph{\textbf{Parameters learning}} In forward propagation, the $l\text{th}$ ($l=1,2,\cdots ,N$) input $\bm{x}_l$ is propagated to neurons in each layer, and the corresponding final output $\hat{\bm y}_l$ of DNN can be obtained. Thus, the error $\mathcal{L}\left( {{\bm{x}}_{l}},{{\bm{y}}_{l}};\bm{\theta } \right)$ between the estimated value $\hat{\bm y}_l$ and the true value $\bm{y}_l$ is measured by the $L_1\text{-norm}$ $ {\left\| \cdot \right \|}_{1} $ or the $L_2\text{-norm}$ $L_1\text{-norm}$ $ {\left\| \cdot \right \|}_{2} $, i.e.,
\begin{equation}\label{ab_error}
\mathcal{L}\left( {{\bm{x}}_{l}},{{\bm{y}}_{l}};\bm{\theta } \right)= {\left({\left \| {{{\bm{\hat{y}}}}_{l}}-{{\bm{y}}_{l}} \right \|}_{r}\right)}^{r} ={\left[{\left\| \hat{f}\left( {{\bm{x}}_{l}};\bm{\theta } \right)-{{\bm{y}}_{l}} \right \|}_{r}\right]}^{r},
\end{equation}
where $ r=1 $ or $ r=2 $. Based on $N$ training data $\left\{ \left( {{\bm{x}}_{l}},{{\bm{y}}_{l}} \right)\left| l=1,2,\cdots ,N \right. \right\}$, the cost function $\mathcal{J}\left(\theta \right)$ can be calculated by
\begin{equation}\label{cost_fun_MLP}
\mathcal{J}\left( \bm{\theta } \right)=\frac{1}{N}\sum\limits_{l=1}^{N}{\mathcal{L}\left( {{\bm{x}}_{l}},{{\bm{y}}_{l}};\bm{\theta } \right)}.
\end{equation}
Therefore, the gradient $\mathcal{G}\left( \bm{\theta } \right)$ of the cost function $\mathcal{J}\left(\theta \right)$ is 
\begin{equation}\label{cost_grad}
\mathcal{G}\left( \bm{\theta } \right)=\frac{\partial \mathcal{J}\left( \bm{\theta } \right)}{\partial \bm{\theta }}=\frac{1}{N}\sum\limits_{l=1}^{N}{\frac{\partial \mathcal{L}\left( {{\bm{x}}_{l}},{{\bm{y}}_{l}};\bm{\theta } \right)}{\partial \bm{\theta }}}.
\end{equation}
Apparently, the training objective is to minimize the cost function $\mathcal{J}\left(\theta \right)$. During the training process, the model parameters $ \bm \theta$ will be updated iteratively as follows:
\begin{equation}\label{Param_update_MLP}
\bm{\theta }\leftarrow \bm{\theta }+\Delta \left[ \eta ,\mathcal{G}\left( \bm{\theta } \right) \right],
\end{equation}
where $\eta $ is the learning rate, $\Delta \left( \cdot \right)$ is the calculation update operator which is determined by the deep neural network optimization algorithm like Adam algorithm \cite{Kingma2014}, AdaGrad algorithm \cite{Duchi2011}, SGD algorithm \cite{Ruder2017}, etc.

\subsection{Arbitrary polynomial chaos expansion}\label{sec22}
\paragraph{\textbf{aPC model}} It is defined that model $Y=f\left( \bm{\xi } \right)$ is a stochastic model with the random variable $\bm{\xi }$ and the model output $Y$, where the random variable $\bm{\xi }=\left\{ {{\xi }_{k}}\left| k=1,2,\cdots ,d \right. \right\}$. For the stochastic analysis of $Y$ , the model $f\left( \bm{\xi } \right)$ can be approximated by the $p$-order aPC model \cite{Oladyshkin2012, Wan2020}, i.e.,
\begin{equation}\label{aPCE}
Y\approx {{y}^{(p)}}\left( \bm{\xi } \right)=\sum\limits_{i=1}^{M}{{{c}_{i}}{{\Phi }_{i}}\left( \bm{\xi } \right)},
\end{equation}
where $c_i$ are the expansion coefficients that can be obtained by Galerkin projection method \cite{Xiu2003, Ghanem1993, Matthies2005} or collocation method \cite{Isukapalli1998, Li2007, Shi2009}, the number of the expansion coefficients $c_i$ is $M={\left( d+p \right)!}/{\left( d!p! \right)}\;$, and ${{\Phi }_{i}}\left( \bm{\xi } \right)$ are the multi-dimensional polynomials forming the multi-dimensional orthogonal basis $\left\{ {{\Phi }_{1}}\left( \bm{\xi } \right),\cdots ,{{\Phi }_{M}}\left( \bm{\xi } \right) \right\}$. For the $i\text{th}$ multi-dimensional polynomial ${{\Phi }_{i}}\left( \bm{\xi } \right)$, it is the product of the univariate polynomials of random variables ${{\xi }_{k}}$ ($k=1,2,\cdots ,d$), i.e.,
\begin{equation}\label{multi_basis}
\begin{aligned}
& {{\Phi }_{i}}\left( \bm{\xi } \right)=\prod\limits_{k=1}^{d}{\phi _{k}^{(s_{i}^{k})}\left( {{\xi }_{k}} \right)}, \\ 
& \sum\limits_{k=1}^{d}{s_{i}^{k}}\le p, \quad i=1,2,\cdots ,M, \\ 
\end{aligned}
\end{equation}
where $ s_{i}^{k} $ is a multivariate index that contains the individual univariate basis combinatoric information \cite{Oladyshkin2012}, and $\phi _{k}^{(s_{i}^{k})}\left( {{\xi }_{k}} \right)$ are the univariate polynomials forming the basis $\left\{ \phi _{k}^{(0)}\left( {{\xi }_{k}} \right),\phi _{k}^{(1)}\left( {{\xi }_{k}} \right),\cdots ,\phi _{k}^{(p)}\left( {{\xi }_{k}} \right) \right\}$ that is orthogonal with respect to the probability measure $\mathit{\mathit{\Gamma}} $\cite{Mircea2002}, i.e.,
\begin{equation}\label{orth_measure}
\int_{{{\xi }_{k}}\in \mathit{\mathit{\Omega}} }{\phi _{k}^{(j)}\left( {{\xi }_{k}} \right)\phi _{k}^{({j}')}\left( {{\xi }_{k}} \right)}d\mathit{\mathit{\Gamma}} \left( {{\xi }_{k}} \right)=\delta_{j{j}'},\quad j=0,1,\cdots ,p,
\end{equation}
where $\mathit{\mathit{\Omega}}$ is the space of events \cite{Mircea2002}, $\delta_{j{j}'}$ is the Kronecker delta function. Different with the Wiener PCE \cite{Wiener1938, Meecham1964, Siegel2004} and the gPC \cite{Xiu2003, Choi2004}, the univariate orthogonal basis $\left\{ \phi _{k}^{(0)}\left( {{\xi }_{k}} \right),\phi _{k}^{(1)}\left( {{\xi }_{k}} \right),\cdots ,\phi _{k}^{(p)}\left( {{\xi }_{k}} \right) \right\}$ can be constructed by the raw moment of the random variable ${\xi}_k$, where the probability measure $\mathit{\mathit{\Gamma}} $ can be arbitrary form.

\paragraph{\textbf{Constructing orthogonal basis}} For the $k\text{th}$ random variable ${\xi}_k$, the $j$-degree ($j=0,1,\cdots ,p$) polynomial $\phi _{k}^{(j)}\left( {{\xi }_{k}} \right)$ is defined to be
\begin{equation}\label{phi_xi_k}
\phi _{k}^{(j)}\left( {{\xi }_{k}} \right)=\sum\limits_{m=0}^{j}{a_{m}^{(j)}(\xi _{k})^{m}}.
\end{equation}
where $a_{m}^{(j)}$ ($j=0,1,\cdots ,p$) are the coefficients of $\phi _{k}^{(j)}\left( {{\xi }_{k}} \right)$. According to Eqs.(\ref{orth_measure}) and (\ref{phi_xi_k}), the coefficients $a_{m}^{(j)}$ can be calculated by the following equation (The detailed derivation process can be found in the reference \cite{Oladyshkin2012}.):
\begin{equation}\label{mu_mat}
\left[ \begin{matrix}
\mu _{{{\xi }_{k}}}^{(0)} & \mu _{{{\xi }_{k}}}^{(1)} & \cdots  & \mu _{{{\xi }_{k}}}^{(j)}  \\
\mu _{{{\xi }_{k}}}^{(1)} & \mu _{{{\xi }_{k}}}^{(2)} & \cdots  & \mu _{{{\xi }_{k}}}^{(j+1)}  \\
\vdots  & \vdots  & \vdots  & \vdots   \\
\mu _{{{\xi }_{k}}}^{(j-1)} & \mu _{{{\xi }_{k}}}^{(j)} & \cdots  & \mu _{{{\xi }_{k}}}^{(2j-1)}  \\
0 & 0 & \cdots  & 1  \\
\end{matrix} \right]\left[ \begin{matrix}
a_{0}^{(j)}  \\
a_{1}^{(j)}  \\
\vdots   \\
a_{j-1}^{(j)}  \\
a_{j}^{(j)}  \\
\end{matrix} \right]=\left[ \begin{matrix}
0  \\
0  \\
\vdots   \\
0  \\
1  \\
\end{matrix} \right],
\end{equation}
where $\mu _{{{\xi }_{k}}}^{(j)}$ is the $j\text{th}$ raw moment of random variable $\xi_{k}$, i.e.
\begin{equation}\label{j_moment}
\mu _{{{\xi }_{k}}}^{(j)}=\int_{{{\xi }_{k}}\in \bm{\mathit{\Omega} }}{(\xi _{k})^{j}}d\mathit{\Gamma} \left( {{\xi }_{k}} \right),
\end{equation}
where $ {(\xi _{k})^{j}} $ denotes $ {\xi _{k}} $ to the power $j$.

To simplify the explicit form of $a_{m}^{(j)}$ ($j=0,1,\cdots ,p$), the random variable $\xi_{k}$ will be normalized, i.e.,
\begin{equation}\label{normlize}
{{{\xi }'}_{k}}=\frac{{{\xi }_{k}}-{{\mu }_{k}}}{{{\sigma }_{k}}},
\end{equation}
where ${\mu }_{k}$ and ${\sigma }_{k}$ are the mean and the standard deviation of random variable $\xi_{k}$, respectively. Apparently, normalization will make all moments to be centralize and standardize. To simplify the formula notation, the following content of this paper assumes that $\bm{\xi }=\left\{ {{\xi }_{k}}\left| k=1,2,\cdots ,d \right. \right\}$ are already a normalized random variable. 

Based on Eqs. (\ref{mu_mat}) and (\ref{normlize}), the coefficients for polynomials with different degrees can be obtained. For example, the coefficients for polynomials of the 0, 1st, 2nd and 3rd degree are shown in Table \ref{coeff_table}. According to Eqs.(\ref{multi_basis}) and (\ref{phi_xi_k}), the multi-dimensional orthogonal basis $\left\{ {{\Phi }_{1}}\left( \bm{\xi } \right),{{\Phi }_{2}}\left( \bm{\xi } \right),\cdots ,{{\Phi }_{M}}\left( \bm{\xi } \right) \right\}$  can be constructed. In section \ref{sec32}, a simple method is proposed for constructing the multi-dimensional orthogonal basis.

\begin{table}[htbp]
	\centering
	\caption{Coefficients for polynomials of the 0, 1st, 2nd and 3rd degree}
	\begin{tabular}{cl}
		\toprule
		Degree & \multicolumn{1}{c}{Coefficients} \\
		\midrule
		0     & $a_{0}^{(0)}=1$ \\
		1     & $a_{0}^{(1)}=0$ \qquad $a_{1}^{(1)}=1$ \\
		2     & $a_{0}^{(2)}=-1$ \qquad $a_{1}^{(2)}=-\mu _{{{\xi }_{k}}}^{(3)}$ \qquad $a_{2}^{(2)}=1$ \\
		3     &$a_{0}^{(3)}={{\left[ \mu _{{{\xi }_{k}}}^{(3)} \right]}^{2}}-{{\left[ \mu _{{{\xi }_{k}}}^{(3)} \right]}^{3}}+\mu _{{{\xi }_{k}}}^{(3)}\mu _{{{\xi }_{k}}}^{(4)}-\mu _{{{\xi }_{k}}}^{(5)}$ \\
		& $a_{1}^{(3)}=-\mu _{{{\xi }_{k}}}^{(3)}\mu _{{{\xi }_{k}}}^{(5)}+{{\left[ \mu _{{{\xi }_{k}}}^{(3)} \right]}^{2}}-\mu _{{{\xi }_{k}}}^{(4)}+\mu _{{{\xi }_{k}}}^{(3)}\mu _{{{\xi }_{k}}}^{(4)}$ \\
		& $a_{2}^{(3)}=-\mu _{{{\xi }_{k}}}^{(3)}\mu _{{{\xi }_{k}}}^{(4)}+\mu _{{{\xi }_{k}}}^{(5)}-\mu _{{{\xi }_{k}}}^{(3)}$ \quad $a_{3}^{(3)}=1-\mu _{{{\xi }_{k}}}^{(3)}+{{\left[ \mu _{{{\xi }_{k}}}^{(3)} \right]}^{2}}$ \\
		\bottomrule
	\end{tabular}
	\label{coeff_table}
\end{table}

\subsection{Error estimates}
In this paper, the well-known determination coefficient $R^2$ and the error $e$ are calculated respectively to measure the accuracy of the surrogate model, i.e.,
\begin{equation}\label{error_R2}
\begin{aligned}
& {{R}^{2}}=1-\frac{{{\varepsilon }_{E}}}{D\left( {{y}^{test}} \right)} \\ 
& E\left( {{y}^{test}} \right)=\frac{1}{{{N}_{test}}}\sum\limits_{{{l}_{t}}=1}^{{{N}_{test}}}{y_{{{l}_{t}}}^{test}} \\ 
& D\left( {{y}^{test}} \right)=\frac{1}{{{N}_{test}}-1}\sum\limits_{{{l}_{t}}=1}^{{{N}_{test}}}{{{\left[ y_{{{l}_{t}}}^{test}-E\left( {{y}^{test}} \right) \right]}^{2}}} \\ 
& {{\varepsilon }_{E}}=\frac{1}{{{N}_{test}}}\sum\limits_{{{l}_{t}}=1}^{{{N}_{test}}}{{{\left[ y_{{{l}_{t}}}^{test}-\hat{y}_{{{l}_{t}}}^{test}\left( \bm{\xi }_{{{l}_{t}}}^{test} \right) \right]}^{2}}}, \\ 
\end{aligned}
\end{equation}
and
\begin{equation}\label{error_e}
e=\sqrt{\frac{\sum\limits_{{{l}_{t}}=1}^{{{N}_{test}}}{{{\left[ y_{{{l}_{t}}}^{test}-\hat{y}_{{{l}_{t}}}^{test}\left( \bm{\xi }_{{{l}_{t}}}^{test} \right) \right]}^{2}}}}{\sum\limits_{{{l}_{t}}=1}^{{{N}_{test}}}{{{\left( y_{{{l}_{t}}}^{test} \right)}^{2}}}}}.
\end{equation}
where $E(\cdot)$ and $D(\cdot)$ respectively are the functions to calculate the mean and variance of input variable, $\bm{\xi }_{{{l}_{t}}}^{test}$ (${{l}_{t}}=1,2,\cdots ,{{N}_{test}}$) are the random inputs of test data, $y_{{{l}_{t}}}^{test}$ are the corresponding true outputs of $\bm{\xi }_{{{l}_{t}}}^{test}$, ${N}_{test}$ is the number of test data, and $ \hat{y}_{{{l}_{t}}}^{test}\left( \bm{\xi }_{{{l}_{t}}}^{test} \right) $ are the corresponding estimated outputs of $\bm{\xi }_{{{l}_{t}}}^{test}$ by the surrogate model. The situation ${{R}^{2}}=1$ or $e=0$ corresponds to a perfect fit. Thus, ${{R}^{2}}\to 1$ or $e\to 0$ means that the surrogate model can fit the stochastic model perfectly.

\section{Proposed adaptive aPC}\label{sec3}
\subsection{Definition of adaptive aPC}\label{sec31}
Defined that a function $\bm{\mathcal{C}} \left (\bm{\xi};\bm{\theta}\right ) $ with the input $ \bm{\xi}\in \bm{\Omega}^d$ has $ M $ outputs $ \left \{ {\mathcal{C}}_i \left (\bm{\xi};\bm{\theta}\right )\mid i=1,2,\cdots , M \right \} $, i.e.,
\begin{equation}\label{C_d_M}
\bm{\Omega}^d\longrightarrow \bm{\Omega}^M:{{\left[ 
\begin{matrix}
{{\mathcal{C}}_{1}}\left( \mathbf{\xi };\mathbf{\theta } \right) & {{\mathcal{C}}_{2}}\left( \mathbf{\xi };\mathbf{\theta } \right) & \cdots  & {{\mathcal{C}}_{M}}\left( \mathbf{\xi };\mathbf{\theta } \right)  \\
\end{matrix}
\right]}^{\text{T}}}=\bm{\mathcal{C}} \left ( \bm{\xi};\bm{\theta} \right ),
\end{equation}
where $\bm{\theta }$ is the parameter of function $\bm{\mathcal{C}}\left ( \bm{\xi};\bm{\theta} \right )$. Based on the outputs $ \left \{ {\mathcal{C}}_i \left (\bm{\xi};\bm{\theta}\right )\mid i=1,2,\cdots , M \right \} $ and the multi-dimensional orthogonal basis $\left\{ {{\Phi }_{1}}\left( \bm{\xi } \right),{{\Phi }_{2}}\left( \bm{\xi } \right),\cdots ,{{\Phi }_{M}}\left( \bm{\xi } \right) \right\}$, a surrogate model of the stochastic system $Y=f\left( \bm{\xi } \right)$ ($\forall \bm{\xi}\in \bm{\Omega}^d$) is constructed to be 
\begin{equation}\label{adaptive_PCE}
{{\hat{y}}^{(p)}}\left( \bm{\xi } \right)=\sum\limits_{i=1}^{M}{{{\mathcal{C}}_{i}}\left( \bm{\xi } ;\bm{\theta} \right){{\Phi }_{i}}\left( \bm{\xi } \right)}.
\end{equation}
Supposed that $\bm{\theta }^{*}$ is the optimal parameter. For $\bm{\theta} \to \bm{\theta}^{*}$, if $ {{\hat{y}}^{(p)}}\left( \bm{\xi } \right) $ and $ \left \{ {\mathcal{C}}_i \left (\bm{\xi};\bm{\theta}\right )\mid i=1,2,\cdots , M \right \} $ meet the following two conditions, i.e.,
\begin{equation}\label{C_xi_Y}
\left |{{\hat{y}}^{(p)}}\left( \bm{\xi } \right)- Y \right | <  {\varepsilon}_1 , \quad \left (\forall {\varepsilon}_1 >0 \right )
\end{equation}
\begin{equation}\label{C_constant}
\left |\mathcal{C}_i \left (\bm{\xi};\bm{\theta}\right ) - E\left[ \mathcal{C}_i \left (\bm{\xi};\bm{\theta}\right ) \right] \right | <  {\varepsilon}_2 , \quad \left (\forall {\varepsilon}_2 >0; \; i=1,2,\cdots,M. \right ),
\end{equation}
Eq.(\ref{adaptive_PCE}) is called the adaptive aPC of the stochastic system $Y=f\left( \bm{\xi } \right)$, and $\left \{ \mathcal{C}_i \left (\bm{\xi};\bm{\theta}\right )\mid i=1,2,\cdots , M \right \} $ are called the adaptive expansion coefficients.

\begin{figure}[!htb]
	\centering
	{\includegraphics[scale=0.9]{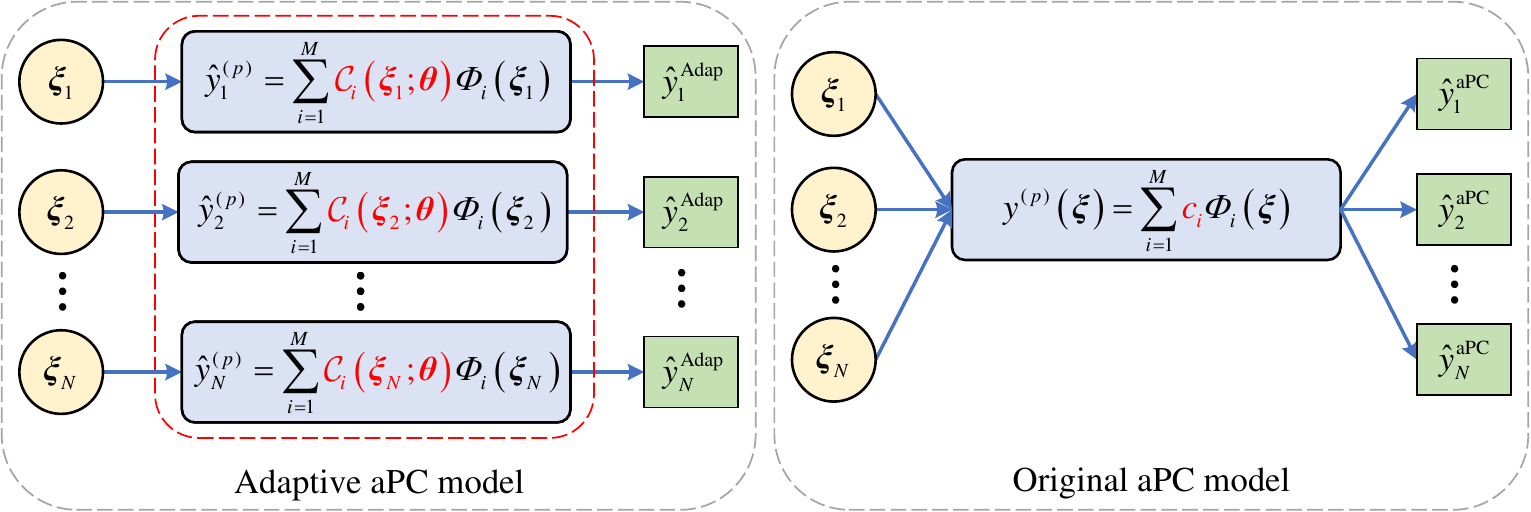}}
	\caption{The prediction processes of the adaptive PCE method and the existing PCE method.}\label{coeff_order}
\end{figure}

Unlike the original aPC model \cite{Oladyshkin2012} in which expansion coefficients are constants, the adaptive expansion coefficients $\left \{ \mathcal{C}_i \left (\bm{\xi};\bm{\theta}\right )\mid i=1,2,\cdots , M \right \} $ are fine-tuned dynamically by the function $\bm{\mathcal{C}} \left (\bm{\xi};\bm{\theta}\right )$ for different random input $\bm{\xi}$, as shown in Fig.\ref{coeff_order}. 

\subsection{Properties of adaptive aPC}\label{sec32}
By the orthonormality of the multi-dimensional orthogonal basis $\left\{ {{\Phi }_{1}}\left( \bm{\xi } \right),{{\Phi }_{2}}\left( \bm{\xi } \right),\cdots ,{{\Phi }_{M}}\left( \bm{\xi } \right) \right\}$, the following four equations can be obtained, i.e.,
\begin{equation}\label{orth_1}
\int_{\bm{\xi}\in \bm{\Omega}^d} {{\Phi }_{1}}\left( \bm{\xi } \right)d\mathit{\Gamma} \left( \bm{\xi } \right)=1
\end{equation}
\begin{equation}\label{orth_2}
\int_{\bm{\xi}\in \bm{\Omega}^d} {{\Phi }_{i}}\left( \bm{\xi } \right)d\mathit{\Gamma} \left( \bm{\xi } \right)=0 \quad (i=2,3,\cdots,M.)
\end{equation}
\begin{equation}\label{orth_3}
\int_{\bm{\xi}\in \bm{\Omega}^d} {{\Phi }^2_{i}}\left( \bm{\xi } \right)d\mathit{\Gamma} \left( \bm{\xi } \right)=1 \quad (i=1,2,\cdots,M.)
\end{equation}
\begin{equation}\label{orth_4}
\int_{\bm{\xi}\in \bm{\Omega}^d} {{\Phi }_{i}}\left( \bm{\xi } \right){{\Phi }_{j}}\left( \bm{\xi } \right)d\mathit{\Gamma} \left( \bm{\xi } \right)=0 \quad (i \ne j;\; i=1,2,\cdots,M; \; j=1,2,\cdots,M.)
\end{equation}

According to Eq.(\ref{adaptive_PCE}), the mean of ${{\hat{y}}^{(p)}}\left( \bm{\xi } \right)$ is
\begin{equation}\label{proof_Ey1}
\begin{aligned}
& E\left[{{\hat{y}}^{(p)}}\left( \bm{\xi } \right)\right]=E\left\{\sum\limits_{i=1}^{M}{{{\mathcal{C}}_{i}}\left( \bm{\xi } ;\bm{\theta} \right){{\Phi }_{i}}\left( \bm{\xi } \right)}\right\} \\
& \qquad \qquad \;\;\;\; = \sum\limits_{i=1}^{M}{E\left[{{\mathcal{C}}_{i}}\left( \bm{\xi } ;\bm{\theta} \right){{\Phi }_{i}}\left( \bm{\xi } \right)\right]} \\
& \qquad \qquad \;\;\;\; = E\left[{{\mathcal{C}}_{1}}\left( \bm{\xi };\bm{\theta} \right){{\Phi }_{1}}\left( \bm{\xi } \right)\right] + \sum\limits_{i=2}^{M}{E\left[{{\mathcal{C}}_{i}}\left( \bm{\xi } ;\bm{\theta} \right){{\Phi }_{i}}\left( \bm{\xi } \right)\right]}.
\end{aligned}
\end{equation}
Refer to Eq.(\ref{C_constant}), ${{\mathcal{C}}_{i}}\left( \bm{\xi } ;\bm{\theta} \right)= E\left[{{\mathcal{C}}_{i}}\left( \bm{\xi } ;\bm{\theta} \right)\right] + {\delta}_i^{\bm{\xi}}$, where ${\delta}_i^{\bm{\xi}}$ is an infinitesimal for $\bm{\theta} \to \bm{\theta}^{*}$. Thus, Eq.(\ref{proof_Ey1}) can be
\begin{equation}\label{proof_Ey2}
\begin{aligned}
& E\left[{{\hat{y}}^{(p)}}\left( \bm{\xi } \right)\right]=E\left[{{\mathcal{C}}_{1}}\left( \bm{\xi };\bm{\theta} \right){{\Phi }_{1}}\left( \bm{\xi } \right)\right] + \sum\limits_{i=2}^{M}{E\left[{{\mathcal{C}}_{i}}\left( \bm{\xi } ;\bm{\theta} \right){{\Phi }_{i}}\left( \bm{\xi } \right)\right]} \\
& \qquad \qquad \;\;\;\; = E\left\{\left[E\left[{{\mathcal{C}}_{1}}\left( \bm{\xi } ;\bm{\theta} \right)\right] + {\delta}_1^{\bm{\xi}}\right]{{\Phi }_{1}}\left( \bm{\xi } \right) \right\} + \sum\limits_{i=2}^{M}{E\left\{\left[E\left[{{\mathcal{C}}_{i}}\left( \bm{\xi } ;\bm{\theta} \right)\right] + {\delta}_i^{\bm{\xi}}\right]{{\Phi }_{i}}\left( \bm{\xi } \right) \right\}} \\
& \qquad \qquad \;\;\;\; = E\left\{E\left[{{\mathcal{C}}_{1}}\left( \bm{\xi } ;\bm{\theta} \right)\right]{{\Phi }_{1}}\left( \bm{\xi } \right)\right\} + E\left[{\delta}_1^{\bm{\xi}}{{\Phi }_{1}}\left( \bm{\xi } \right)\right] + \sum\limits_{i=2}^{M}{E\left\{E\left[{{\mathcal{C}}_{i}}\left( \bm{\xi } ;\bm{\theta} \right)\right]{{\Phi }_{1}}\left( \bm{\xi } \right)\right\}} + \sum\limits_{i=2}^{M}{E\left[{\delta}_i^{\bm{\xi}}{{\Phi }_{i}}\left( \bm{\xi } \right)\right]} \\
& \qquad \qquad \;\;\;\; = \int_{\bm{\xi}\in \bm{\Omega}^d} E\left[{{\mathcal{C}}_{1}}\left( \bm{\xi } ;\bm{\theta} \right)\right]{{\Phi }_{1}}\left( \bm{\xi } \right) d\mathit{\Gamma} \left( \bm{\xi } \right) + \sum\limits_{i=2}^{M}{\left[\int_{\bm{\xi}\in \bm{\Omega}^d} E\left[{{\mathcal{C}}_{i}}\left( \bm{\xi } ;\bm{\theta} \right)\right]{{\Phi }_{i}}\left( \bm{\xi } \right) d\mathit{\Gamma} \left( \bm{\xi } \right)\right]} d\mathit{\Gamma} \left( \bm{\xi } \right) \\
& \qquad \qquad \qquad \; + E\left[{\delta}_1^{\bm{\xi}}{{\Phi }_{1}}\left( \bm{\xi } \right)\right] + \sum\limits_{i=2}^{M}{E\left[{\delta}_i^{\bm{\xi}}{{\Phi }_{i}}\left( \bm{\xi } \right)\right]} \\
& \qquad \qquad \;\;\;\; = E\left[{{\mathcal{C}}_{1}}\left( \bm{\xi } ;\bm{\theta} \right)\right]\int_{\bm{\xi}\in \bm{\Omega}^d}{{\Phi }_{1}}\left( \bm{\xi } \right) d\mathit{\Gamma} \left( \bm{\xi } \right) + \sum\limits_{i=2}^{M}{\left[E\left[{{\mathcal{C}}_{i}}\left( \bm{\xi } ;\bm{\theta} \right)\right]\int_{\bm{\xi}\in \bm{\Omega}^d}{{\Phi }_{i}}\left( \bm{\xi } \right) d\mathit{\Gamma} \left( \bm{\xi } \right)\right]} \\
& \qquad \qquad \qquad \; + \sum\limits_{i=1}^{M}{E\left[{\delta}_i^{\bm{\xi}}{{\Phi }_{i}}\left( \bm{\xi } \right)\right]} \\
& \qquad \qquad \;\;\;\; = E\left[{{\mathcal{C}}_{1}}\left( \bm{\xi } ;\bm{\theta} \right)\right]\int_{\bm{\xi}\in \bm{\Omega}^d}{{\Phi }_{1}}\left( \bm{\xi } \right) d\mathit{\Gamma} \left( \bm{\xi } \right) + \sum\limits_{i=2}^{M}{\left[E\left[{{\mathcal{C}}_{i}}\left( \bm{\xi } ;\bm{\theta} \right)\right]\int_{\bm{\xi}\in \bm{\Omega}^d}{{\Phi }_{i}}\left( \bm{\xi } \right) d\mathit{\Gamma} \left( \bm{\xi } \right)\right]} \\
& \qquad \qquad \qquad \; + \sum\limits_{i=1}^{M}{\left[\frac{1}{N}\sum_{l=1}^{N}{{\delta}_i^{\bm{\xi}_l}{{\Phi }_{i}}\left( \bm{\xi }_l \right)}\right]}, \\ 
\end{aligned}
\end{equation}
where $N$ is a finite positive integer. Due to ${\delta}_i^{\bm{\xi}_l}$ ($l=1,2,\cdots, N$) are infinitesimals, ${\delta}_i^{\bm{\xi}_l}{{\Phi }_{i}}\left( \bm{\xi }_l \right)$ is an infinitesimal for $\bm{\theta} \to \bm{\theta}^{*}$. Therefore, 
\begin{equation}\label{sum_delta}
{\delta}^{\bm{\xi}} = \sum\limits_{i=1}^{M}{\left[\frac{1}{N}\sum_{l=1}^{N}{{\delta}_i^{\bm{\xi}_l}{{\Phi }_{i}}\left( \bm{\xi }_l \right)}\right]}
\end{equation}
is an infinitesimal for $\bm{\theta} \to \bm{\theta}^{*}$. On the basis of Eqs.(\ref{orth_1}), (\ref{orth_2}) and (\ref{sum_delta}), Eq.(\ref{proof_Ey2}) can be

\begin{equation}\label{proof_Ey3}
\begin{aligned}
& E\left[{{\hat{y}}^{(p)}}\left( \bm{\xi } \right)\right]=E\left[{{\mathcal{C}}_{1}}\left( \bm{\xi } ;\bm{\theta} \right)\right]\int_{\bm{\xi}\in \bm{\Omega}^d}{{\Phi }_{1}}\left( \bm{\xi } \right) d\mathit{\Gamma} \left( \bm{\xi } \right) + \sum\limits_{i=2}^{M}{\left[E\left[{{\mathcal{C}}_{i}}\left( \bm{\xi } ;\bm{\theta} \right)\right]\int_{\bm{\xi}\in \bm{\Omega}^d}{{\Phi }_{i}}\left( \bm{\xi } \right) d\mathit{\Gamma} \left( \bm{\xi } \right)\right]} \\
& \qquad \qquad \qquad \; + \sum\limits_{i=1}^{M}{\left[\frac{1}{N}\sum_{l=1}^{N}{{\delta}_i^{\bm{\xi}_l}{{\Phi }_{i}}\left( \bm{\xi }_l \right)}\right]} \\ 
& \qquad \qquad \;\;\;\; = E\left[{{\mathcal{C}}_{1}}\left( \bm{\xi } ;\bm{\theta} \right)\right] + {\delta}^{\bm{\xi}}. \\ 
\end{aligned}
\end{equation}
Thus, the mean of ${{\hat{y}}^{(p)}}\left( \bm{\xi } \right)$ approaches the mean of the first adaptive expansion coefficient ${\mathcal{C}_{1}}\left( \bm{\xi};\bm{\theta} \right)$ for $\bm{\theta} \to \bm{\theta}^{*}$, i.e.,
\begin{equation}\label{aaPC_proper_I}
\left | E\left[ {{{\hat{y}}}^{(p)}}\left( \bm{\xi } \right) \right] - E\left[ {\mathcal{C}_{1}}\left( \bm{\xi};\bm{\theta} \right) \right] \right | <  {\varepsilon}_3 , \quad \left (\forall {\varepsilon}_3 >0 \right ).
\end{equation}

According to Eq.(\ref{adaptive_PCE}), the variance of ${{\hat{y}}^{(p)}}\left( \bm{\xi } \right)$ is
\begin{equation}\label{proof_Vary1}
\begin{aligned}
& D\left[{{\hat{y}}^{(p)}}\left( \bm{\xi } \right)\right] = E\left\{\left[\sum\limits_{i=1}^{M}{{{\mathcal{C}}_{i}}\left( \bm{\xi } ;\bm{\theta} \right){{\Phi }_{i}}\left( \bm{\xi } \right)}\right]^2\right\} - E^2\left[{{\hat{y}}^{(p)}}\left( \bm{\xi } \right)\right] \\
& \qquad \qquad \;\;\;\; = E\left\{\sum\limits_{i=1}^{M}{{{\mathcal{C}}^2_{i}}\left( \bm{\xi } ;\bm{\theta} \right){{\Phi }^2_{i}}\left( \bm{\xi } \right)} + \sum\limits_{j=1}^{M}{\sum\limits_{i=1, i \ne j}^{M}{{{\mathcal{C}}_{i}}\left( \bm{\xi } ;\bm{\theta} \right){{\mathcal{C}}_{j}}\left( \bm{\xi } ;\bm{\theta} \right){{\Phi }_{i}}\left( \bm{\xi } \right){{\Phi }_{j}}\left( \bm{\xi } \right)}} \right\} - E^2\left[{{\hat{y}}^{(p)}}\left( \bm{\xi } \right)\right] \\
& \qquad \qquad \;\;\;\; = \sum\limits_{i=1}^{M}{E\left[{{\mathcal{C}}^2_{i}}\left( \bm{\xi } ;\bm{\theta} \right){{\Phi }^2_{i}}\left( \bm{\xi } \right)\right]} + \sum\limits_{j=1}^{M}{\sum\limits_{i=1, i \ne j}^{M}{E\left[{{\mathcal{C}}_{i}}\left( \bm{\xi } ;\bm{\theta} \right){{\mathcal{C}}_{j}}\left( \bm{\xi } ;\bm{\theta} \right){{\Phi }_{i}}\left( \bm{\xi } \right){{\Phi }_{j}}\left( \bm{\xi } \right)\right]}} - E^2\left[{{\hat{y}}^{(p)}}\left( \bm{\xi } \right)\right].
\end{aligned}
\end{equation}
Refer to Eq.(\ref{C_constant}), 
\begin{equation}\label{Ci_2}
\begin{aligned}
& {{\mathcal{C}}^2_{i}}\left( \bm{\xi };\bm{\theta} \right) = \left\{E\left[{{\mathcal{C}}_{i}}\left( \bm{\xi } ;\bm{\theta} \right)\right] + {\delta}_i^{\bm{\xi}}\right\}^2 = E^2\left[{{\mathcal{C}}_{i}}\left( \bm{\xi } ;\bm{\theta} \right)\right] + 2{\delta}_i^{\bm{\xi}} E\left[{{\mathcal{C}}_{i}}\left( \bm{\xi } ;\bm{\theta} \right)\right] + \left({\delta}_i^{\bm{\xi}}\right)^2 \\
& \qquad \quad \;\; = E^2\left[{{\mathcal{C}}_{i}}\left( \bm{\xi } ;\bm{\theta} \right)\right] + \hat{\delta}_i^{\bm{\xi}},
\end{aligned}
\end{equation}
\begin{equation}\label{Ci_ij}
\begin{aligned}
& {{\mathcal{C}}_{i}}\left( \bm{\xi };\bm{\theta} \right){{\mathcal{C}}_{j}}\left( \bm{\xi };\bm{\theta} \right) = \left\{E\left[{{\mathcal{C}}_{i}}\left( \bm{\xi } ;\bm{\theta} \right)\right] + {\delta}_i^{\bm{\xi}}\right\}\left\{E\left[{{\mathcal{C}}_{j}}\left( \bm{\xi } ;\bm{\theta} \right)\right] + {\delta}_j^{\bm{\xi}}\right\} \\
& \qquad \qquad \qquad \;\;\;\, = E\left[{{\mathcal{C}}_{i}}\left( \bm{\xi } ;\bm{\theta} \right)\right] E\left[{{\mathcal{C}}_{j}}\left( \bm{\xi } ;\bm{\theta} \right)\right] + {\delta}_i^{\bm{\xi}} E\left[{{\mathcal{C}}_{j}}\left( \bm{\xi } ;\bm{\theta} \right)\right] + {\delta}_j^{\bm{\xi}} E\left[{{\mathcal{C}}_{i}}\left( \bm{\xi } ;\bm{\theta} \right)\right] + {\delta}_i^{\bm{\xi}}{\delta}_j^{\bm{\xi}} \\
& \qquad \qquad \qquad \;\;\;\, = E\left[{{\mathcal{C}}_{i}}\left( \bm{\xi } ;\bm{\theta} \right)\right] E\left[{{\mathcal{C}}_{j}}\left( \bm{\xi } ;\bm{\theta} \right)\right] + \hat{\delta}_{ij}^{\bm{\xi}},
\end{aligned}
\end{equation}
where both $\hat{\delta}_i^{\bm{\xi}} = 2{\delta}_i^{\bm{\xi}} E\left[{{\mathcal{C}}_{i}}\left( \bm{\xi } ;\bm{\theta} \right)\right] + \left({\delta}_i^{\bm{\xi}}\right)^2$ and $\hat{\delta}_{ij}^{\bm{\xi}}={\delta}_i^{\bm{\xi}} E\left[{{\mathcal{C}}_{j}}\left( \bm{\xi } ;\bm{\theta} \right)\right] + {\delta}_j^{\bm{\xi}} E\left[{{\mathcal{C}}_{i}}\left( \bm{\xi } ;\bm{\theta} \right)\right] + {\delta}_i^{\bm{\xi}}{\delta}_j^{\bm{\xi}}$ are infinitesimals for $\bm{\theta} \to \bm{\theta}^{*}$. Therefore, Eq.(\ref{proof_Vary1}) can be
\begin{equation}\label{proof_Vary2}
\begin{aligned}
& D\left[{{\hat{y}}^{(p)}}\left( \bm{\xi } \right)\right] =\sum\limits_{i=1}^{M}{E\left[{{\mathcal{C}}^2_{i}}\left( \bm{\xi } ;\bm{\theta} \right){{\Phi }^2_{i}}\left( \bm{\xi } \right)\right]} + \sum\limits_{j=1}^{M}{\sum\limits_{i=1, i \ne j}^{M}{E\left[{{\mathcal{C}}_{i}}\left( \bm{\xi } ;\bm{\theta} \right){{\mathcal{C}}_{j}}\left( \bm{\xi } ;\bm{\theta} \right){{\Phi }_{i}}\left( \bm{\xi } \right){{\Phi }_{j}}\left( \bm{\xi } \right)\right]}} - E^2\left[{{\hat{y}}^{(p)}}\left( \bm{\xi } \right)\right] \\
& \qquad \qquad \qquad = \sum\limits_{i=1}^{M}{E\left\{\left[E^2\left[{{\mathcal{C}}_{i}}\left( \bm{\xi } ;\bm{\theta} \right)\right] + \hat{\delta}_i^{\bm{\xi}}\right]{{\Phi }^2_{i}}\left( \bm{\xi } \right)\right\}} + \sum\limits_{j=1}^{M}{\sum\limits_{i=1, i \ne j}^{M}{E\left\{\left\{E\left[{{\mathcal{C}}_{i}}\left( \bm{\xi } ;\bm{\theta} \right)\right] E\left[{{\mathcal{C}}_{j}}\left( \bm{\xi } ;\bm{\theta} \right)\right] + \hat{\delta}_{ij}^{\bm{\xi}}\right\}{{\Phi }_{i}}\left( \bm{\xi } \right){{\Phi }_{j}}\left( \bm{\xi } \right)\right\}}} \\
& \qquad \qquad \qquad \;\;\;\; - E^2\left[{{\hat{y}}^{(p)}}\left( \bm{\xi } \right)\right] \\
& \qquad \qquad \qquad = \sum\limits_{i=1}^{M}{E\left\{E^2\left[{{\mathcal{C}}_{i}}\left( \bm{\xi } ;\bm{\theta} \right)\right]{{\Phi }^2_{i}}\left( \bm{\xi } \right)\right\}} + \sum\limits_{j=1}^{M}{\sum\limits_{i=1, i \ne j}^{M}{E\left\{E\left[{{\mathcal{C}}_{i}}\left( \bm{\xi } ;\bm{\theta} \right)\right]E\left[{{\mathcal{C}}_{j}}\left( \bm{\xi } ;\bm{\theta} \right)\right]{{\Phi }_{i}}\left( \bm{\xi } \right){{\Phi }_{j}}\left( \bm{\xi } \right)\right\}}} \\
& \qquad \qquad \qquad \;\;\;\;  - E^2\left[{{\hat{y}}^{(p)}}\left( \bm{\xi } \right)\right] + \sum\limits_{i=1}^{M}{E\left[\hat{\delta}_i^{\bm{\xi}}{{\Phi }^2_{i}}\left( \bm{\xi } \right)\right]} + \sum\limits_{j=1}^{M}{\sum\limits_{i=1, i \ne j}^{M}{E\left[\hat{\delta}_{ij}^{\bm{\xi}}{{\Phi }_{i}}\left( \bm{\xi } \right){{\Phi }_{j}}\left( \bm{\xi } \right)\right]}} \\
& \qquad \qquad \qquad = \sum\limits_{i=1}^{M}{E^2\left[{{\mathcal{C}}_{i}}\left( \bm{\xi } ;\bm{\theta} \right)\right]E\left[{{\Phi }^2_{i}}\left( \bm{\xi } \right)\right]} + \sum\limits_{j=1}^{M}{\sum\limits_{i=1, i \ne j}^{M}{E\left[{{\mathcal{C}}_{i}}\left( \bm{\xi } ;\bm{\theta} \right)\right]E\left[{{\mathcal{C}}_{j}}\left( \bm{\xi } ;\bm{\theta} \right)\right]E\left[{{\Phi }_{i}}\left( \bm{\xi } \right){{\Phi }_{j}}\left( \bm{\xi } \right)\right]}} - E^2\left[{{\hat{y}}^{(p)}}\left( \bm{\xi } \right)\right] \\
& \qquad \qquad \qquad \;\;\;\; + \sum\limits_{i=1}^{M}{E\left[\hat{\delta}_i^{\bm{\xi}}{{\Phi }^2_{i}}\left( \bm{\xi } \right)\right]} + \sum\limits_{j=1}^{M}{\sum\limits_{i=1, i \ne j}^{M}{E\left[\hat{\delta}_{ij}^{\bm{\xi}}{{\Phi }_{i}}\left( \bm{\xi } \right){{\Phi }_{j}}\left( \bm{\xi } \right)\right]}} \\
& \qquad \qquad \qquad = \sum\limits_{i=1}^{M}{E^2\left[{{\mathcal{C}}_{i}}\left( \bm{\xi } ;\bm{\theta} \right)\right]\int_{\bm{\xi}\in \bm{\Omega}^d} {{\Phi }^2_{i}}\left( \bm{\xi } \right)d\mathit{\Gamma} \left( \bm{\xi } \right)} + \sum\limits_{j=1}^{M}{\sum\limits_{i=1, i \ne j}^{M}{E\left[{{\mathcal{C}}_{i}}\left( \bm{\xi } ;\bm{\theta} \right)\right]E\left[{{\mathcal{C}}_{j}}\left( \bm{\xi } ;\bm{\theta} \right)\right]\int_{\bm{\xi}\in \bm{\Omega}^d} {{\Phi }_{i}}\left( \bm{\xi } \right){{\Phi }_{j}}\left( \bm{\xi } \right)d\mathit{\Gamma} \left( \bm{\xi } \right)}} \\
& \qquad \qquad \qquad \;\;\;\;  - E^2\left[{{\hat{y}}^{(p)}}\left( \bm{\xi } \right)\right] + \sum\limits_{i=1}^{M}{\left[\frac{1}{N}\sum_{l=1}^{N}{\hat{\delta}_i^{\bm{\xi}_l}{{\Phi }^2_{i}}\left( \bm{\xi }_l \right)}\right]} + \sum\limits_{j=1}^{M}{\sum\limits_{i=1, i \ne j}^{M}{\left[\frac{1}{N}\sum_{l=1}^{N}{\hat{\delta}_{ij}^{\bm{\xi}_l}{{\Phi }_{i}}\left( \bm{\xi }_l \right){{\Phi }_{j}}\left( \bm{\xi }_l \right)}\right]}}, \\
\end{aligned}
\end{equation}
where $N$ is a finite positive integer. Due to $\hat{\delta}_{i}^{\bm{\xi}_l}$ and $\hat{\delta}_{ij}^{\bm{\xi}_l}$ ($l=1,2,\cdots, N$) are infinitesimals, both $\hat{\delta}_i^{\bm{\xi}_l}{{\Phi }^2_{i}}\left( \bm{\xi }_l \right)$ and $\hat{\delta}_{ij}^{\bm{\xi}_l}{{\Phi }_{i}}\left( \bm{\xi }_l \right){{\Phi }_{j}}\left( \bm{\xi }_l \right)$ are infinitesimals for $\bm{\theta} \to \bm{\theta}^{*}$. Therefore, 
\begin{equation}\label{sum_hat_delta}
\hat{\delta}^{\bm{\xi}} = \sum\limits_{i=1}^{M}{\left[\frac{1}{N}\sum_{l=1}^{N}{\hat{\delta}_i^{\bm{\xi}_l}{{\Phi }^2_{i}}\left( \bm{\xi }_l \right)}\right]} + \sum\limits_{j=1}^{M}{\sum\limits_{i=1, i \ne j}^{M}{\left[\frac{1}{N}\sum_{l=1}^{N}{\hat{\delta}_{ij}^{\bm{\xi}_l}{{\Phi }_{i}}\left( \bm{\xi }_l \right){{\Phi }_{j}}\left( \bm{\xi }_l \right)}\right]}}
\end{equation}
is an infinitesimal for $\bm{\theta} \to \bm{\theta}^{*}$. On the basis of Eqs.(\ref{orth_3}), (\ref{orth_4}), (\ref{proof_Ey3}) and (\ref{sum_hat_delta}), Eq.(\ref{proof_Vary2}) can be
\begin{equation}\label{proof_Vary3}
\begin{aligned}
& D\left[{{\hat{y}}^{(p)}}\left( \bm{\xi } \right)\right] =\sum\limits_{i=1}^{M}{E^2\left[{{\mathcal{C}}_{i}}\left( \bm{\xi } ;\bm{\theta} \right)\right]\int_{\bm{\xi}\in \bm{\Omega}^d} {{\Phi }^2_{i}}\left( \bm{\xi } \right)d\mathit{\Gamma} \left( \bm{\xi } \right)} + \sum\limits_{j=1}^{M}{\sum\limits_{i=1, i \ne j}^{M}{E\left[{{\mathcal{C}}_{i}}\left( \bm{\xi } ;\bm{\theta} \right)\right]E\left[{{\mathcal{C}}_{j}}\left( \bm{\xi } ;\bm{\theta} \right)\right]\int_{\bm{\xi}\in \bm{\Omega}^d} {{\Phi }_{i}}\left( \bm{\xi } \right){{\Phi }_{j}}\left( \bm{\xi } \right)d\mathit{\Gamma} \left( \bm{\xi } \right)}} \\
& \qquad \qquad \qquad \;\;\;\;  - E^2\left[{{\hat{y}}^{(p)}}\left( \bm{\xi } \right)\right] + \sum\limits_{i=1}^{M}{\left[\frac{1}{N}\sum_{l=1}^{N}{\hat{\delta}_i^{\bm{\xi}_l}{{\Phi }^2_{i}}\left( \bm{\xi }_l \right)}\right]} + \sum\limits_{j=1}^{M}{\sum\limits_{i=1, i \ne j}^{M}{\left[\frac{1}{N}\sum_{l=1}^{N}{\hat{\delta}_{ij}^{\bm{\xi}_l}{{\Phi }_{i}}\left( \bm{\xi }_l \right){{\Phi }_{j}}\left( \bm{\xi }_l \right)}\right]}} \\
& \qquad \qquad \qquad = \sum\limits_{i=1}^{M}{E^2\left[{{\mathcal{C}}_{i}}\left( \bm{\xi } ;\bm{\theta} \right)\right]} - \left\{E^2\left[{{\mathcal{C}}_{1}}\left( \bm{\xi };\bm{\theta} \right)\right] + {\delta}^{\bm{\xi}}\right\} + \hat{\delta}^{\bm{\xi}} \\
& \qquad \qquad \qquad = \sum\limits_{i=1}^{M}{E^2\left[{{\mathcal{C}}_{i}}\left( \bm{\xi } ;\bm{\theta} \right)\right]} - E^2\left[{{\mathcal{C}}_{1}}\left( \bm{\xi };\bm{\theta} \right)\right] + \left( \hat{\delta}^{\bm{\xi}} - {\delta}^{\bm{\xi}}\right) \\
& \qquad \qquad \qquad = \sum\limits_{i=2}^{M}{E^2\left[{{\mathcal{C}}_{i}}\left( \bm{\xi } ;\bm{\theta} \right)\right]} + \bar{\delta}^{\bm{\xi}}, \\
\end{aligned}
\end{equation}
where $\bar{\delta}^{\bm{\xi}} = \hat{\delta}^{\bm{\xi}} - {\delta}^{\bm{\xi}}$ is an infinitesimal for $\bm{\theta} \to \bm{\theta}^{*}$. Thus, in addition to the first adaptive expansion coefficient ${{\mathcal{C}}_{1}}\left( \bm{\xi };\bm{\theta} \right)$, the variance of ${{\hat{y}}^{(p)}}\left( \bm{\xi } \right)$ approaches the sum of squares of the adaptive expansion coefficients for $\bm{\theta} \to \bm{\theta}^{*}$, i.e.,
\begin{equation}\label{aaPC_proper_II}
\left | D\left[ {{{\hat{y}}}^{(p)}}\left( \bm{\xi } \right) \right] - \sum\limits_{i=2}^{M}{{{\left\{ E\left[ {\mathcal{C}_{i}}\left( \bm{\xi };\bm{\theta} \right) \right] \right\}}^{2}}} \right | <  {\varepsilon}_4 , \quad \left (\forall {\varepsilon}_4 >0 \right )
\end{equation}

In summary, based on the orthonormality of the multi-dimensional orthogonal basis $\left\{ {{\Phi }_{1}}\left( \bm{\xi } \right),{{\Phi }_{2}}\left( \bm{\xi } \right),\cdots ,{{\Phi }_{M}}\left( \bm{\xi } \right) \right\}$, the adaptive aPC in Eq.(\ref{adaptive_PCE}) satisfies the following two properties for $\bm{\theta} \to \bm{\theta}^{*}$, i.e.,
\begin{equation}\label{aaPC_property}
\begin{aligned}
& \left | E\left[ {{{\hat{y}}}^{(p)}}\left( \bm{\xi } \right) \right] - E\left[ {\mathcal{C}_{1}}\left( \bm{\xi};\bm{\theta} \right) \right] \right | <  {\varepsilon}_3 , \\
& \left | D\left[ {{{\hat{y}}}^{(p)}}\left( \bm{\xi } \right) \right] - \sum\limits_{i=2}^{M}{{{\left\{ E\left[ {\mathcal{C}_{i}}\left( \bm{\xi };\bm{\theta} \right) \right] \right\}}^{2}}} \right | <  {\varepsilon}_4 ,
\end{aligned}
\end{equation}
where $ \forall {\varepsilon}_3 >0 $ and $ \forall {\varepsilon}_4 >0 $.

\section{Deep aPCE method for uncertainty quantification}\label{sec4}
\subsection{Proposed Deep aPCE method}\label{sec41}
\begin{figure}[htbp]
	\centering
	{\includegraphics[scale=0.69]{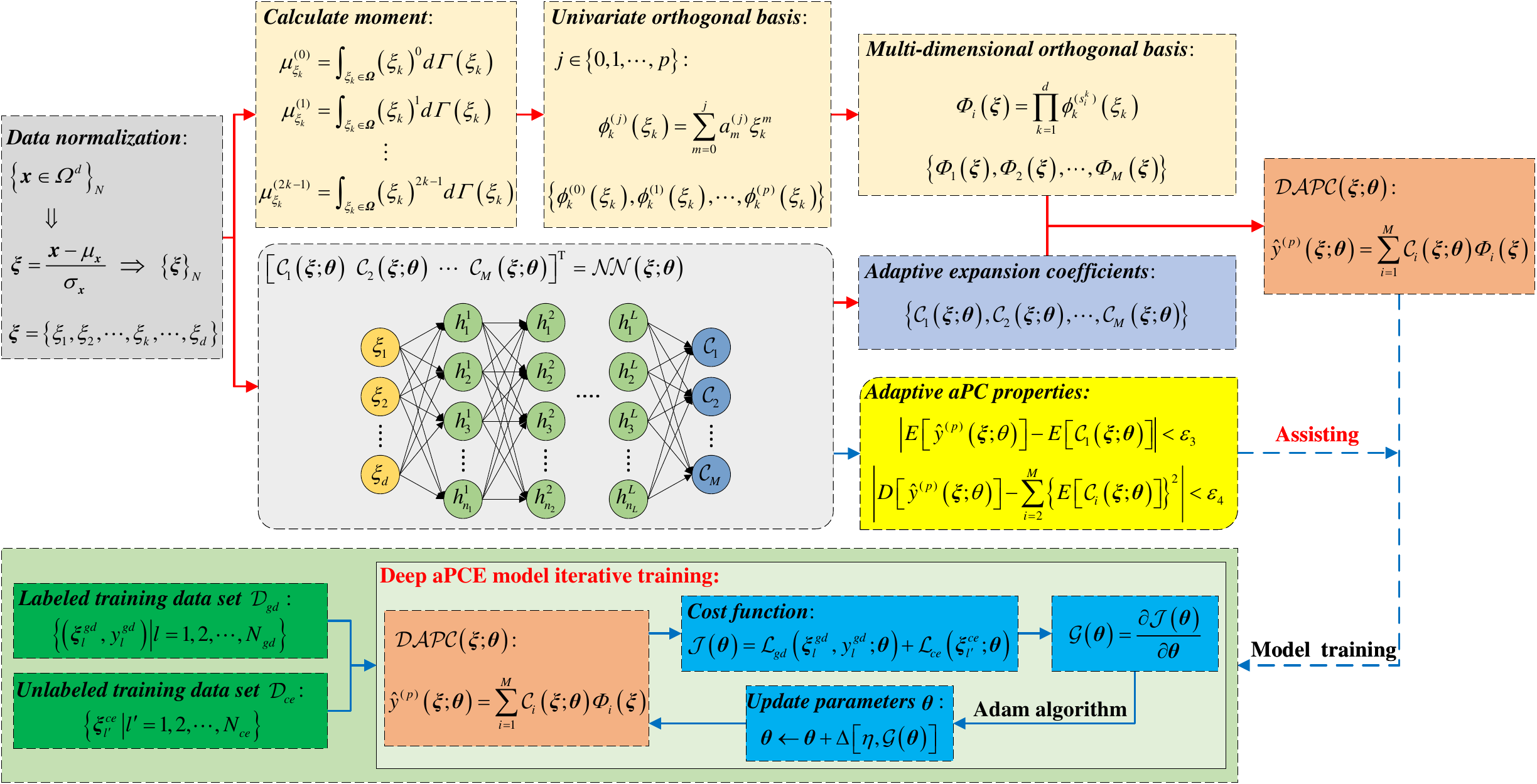}}
	\caption{The framework of the Deep aPCE method}\label{framework}
\end{figure}

According to the researches by Hornik et al. \cite{Hornik1989} and Cybenko \cite{Cybenko1989}, DNN is a universal approximator that can achieve any desired degree of accuracy by choosing a suitable DNN structure. Thus, DNN is used to construct the function $\bm{\mathcal{C}} \left (\bm{\xi};\bm{\theta}\right )$. In this paper, a semi-supervised Deep aPCE method is proposed to quantify the uncertainty of stochastic model $Y=f\left( \bm{\xi } \right)$ based on the DL and the adaptive aPC, and the framework of the Deep aPCE method is shown in Fig.\ref{framework}. Refer to Fig.\ref{framework}, the random input data ${{\left\{ \bm{x}\in {{\Omega }^{d}} \right\}}_{N}}$ are firstly normalized to be ${{\left\{ \bm{\xi} \right\}}_{N}}$ by Eq.(\ref{normlize}). On the one hand, the first $(2k-1)\text{th}$ moments of input data  ${{\left\{ \bm{\xi} \right\}}_{N}}$ are calculated by Eq.(\ref{j_moment}), and then the multi-dimensional orthogonal basis $\left\{ {{\Phi }_{1}}\left( \bm{\xi } \right),{{\Phi }_{2}}\left( \bm{\xi } \right),\cdots ,{{\Phi }_{M}}\left( \bm{\xi } \right) \right\}$ can be obtained by Eq.(\ref{multi_basis}). On the other hand, a suitable DNN $\mathcal{N}\mathcal{N}\left( \bm{\xi };\bm{\theta } \right)$ is chosen to be the function $\bm{\mathcal{C}} \left (\bm{\xi};\bm{\theta}\right )$ so that the adaptive expansion coefficients $\left \{ \mathcal{C}_i \left (\bm{\xi};\bm{\theta}\right )\mid i=1,2,\cdots , M \right \} $ are solved, i.e.,
\begin{equation}\label{ex_coeff}
{{\left[ 
\begin{matrix}
{{\mathcal{C}}_{1}}\left( \mathbf{\xi };\mathbf{\theta } \right) & {{\mathcal{C}}_{2}}\left( \mathbf{\xi };\mathbf{\theta } \right) & \cdots  & {{\mathcal{C}}_{M}}\left( \mathbf{\xi };\mathbf{\theta } \right)  \\
\end{matrix}
\right]}^{\text{T}}}
=\bm{\mathcal{C}} \left (\bm{\xi};\bm{\theta}\right )=\mathcal{N}\mathcal{N}\left( \bm{\xi };\bm{\theta } \right).
\end{equation}
To decrease the number of labeled training data, the properties of adaptive aPC in Eqs.(\ref{aaPC_property}) are used to assist in training the DNN $\mathcal{N}\mathcal{N}\left( \bm{\xi };\bm{\theta } \right)$. According to Eq.(\ref{adaptive_PCE}), the Deep aPCE model $\mathcal{D}\mathcal{A}\mathcal{P}\mathcal{C}\left( \bm{\xi };\bm{\theta } \right)$ can be built based on the multi-dimensional orthogonal basis $\left\{ {{\Phi }_{1}}\left( \bm{\xi } \right),{{\Phi }_{2}}\left( \bm{\xi } \right),\cdots ,{{\Phi }_{M}}\left( \bm{\xi } \right) \right\}$ and the adaptive expansion coefficients $\left \{ \mathcal{C}_i \left (\bm{\xi};\bm{\theta}\right )\mid i=1,2,\cdots , M \right \} $. Apparently, the Deep aPCE model $\mathcal{D}\mathcal{A}\mathcal{P}\mathcal{C}\left( \bm{\xi };\bm{\theta } \right)$ and the DNN $\mathcal{N}\mathcal{N}\left( \bm{\xi };\bm{\theta } \right)$ share the same parameters $\bm{\theta}=\left( \bm{W},\bm{b}\right)$. Finally, based on the proposed semi-supervised cost function $\mathcal{J}\left(\bm{\theta}\right)$ and two kinds of training data sets, the shared parameters $\bm{\theta}=\left( \bm{W},\bm{b}\right)$ are learned iteratively by the Adam algorithm \cite{Kingma2014}.

In summary, based on the normalized input data, the construction of Deep aPCE model $\mathcal{D}\mathcal{A}\mathcal{P}\mathcal{C}\left( \bm{\xi };\bm{\theta } \right)$ mainly includes the following three parts:
\begin{itemize}
	\item Constructing the multi-dimensional orthogonal basis $\left\{ {{\Phi }_{1}}\left( \bm{\xi } \right),{{\Phi }_{2}}\left( \bm{\xi } \right),\cdots ,{{\Phi }_{M}}\left( \bm{\xi } \right) \right\}$;
	\item Proposing a semi-supervised cost function $\mathcal{J}\left(\bm{\theta}\right)$ based on the properties of adaptive aPC;
	\item Training the Deep aPCE model $\mathcal{DAPC}\left( \bm{\xi };\bm{\theta } \right)$  to learn the shared parameters $\bm{\theta}=\left( \bm{W},\bm{b}\right)$.
\end{itemize}
For the first part, the multi-dimensional orthogonal basis $\left\{ {{\Phi }_{1}}\left( \bm{\xi } \right),{{\Phi }_{2}}\left( \bm{\xi } \right),\cdots ,{{\Phi }_{M}}\left( \bm{\xi } \right) \right\}$ can be constructed based on the univariate orthogonal basis $\left\{ \phi _{k}^{(0)}\left( {{\xi }_{k}} \right),\phi _{k}^{(1)}\left( {{\xi }_{k}} \right),\cdots ,\phi _{k}^{(p)}\left( {{\xi }_{k}} \right) \right\}$ and the multivariate index (See \ref{appendix_B}) by Eq.(\ref{multi_basis}). For the latter two parts, they are presented in sections \ref{sec42} and \ref{sec43}, respectively.

\subsection{Proposed semi-supervised cost function}\label{sec42}
This section proposes a semi-supervised cost function $\mathcal{J}\left(\bm{\theta} \right)$ for learning the shared parameters $\bm{\theta}=\left( \bm{W},\bm{b}\right)$ based on two kinds of training data sets, i.e.,
\begin{itemize}
	\item The labeled training data set ${{\mathcal{D}}_{gd}}$ with  $N_{gd}$ data pairs $\left( \bm{x }_{l}^{gd},y_{l}^{gd} \right)$, i.e., ${{\mathcal{D}}_{gd}}=\left\{ \left( \bm{x }_{l}^{gd},y_{l}^{gd} \right)\left| l=1,2,\cdots ,{{N}_{gd}} \right. \right\}$. Apparently, each input $\bm{x }_{l}^{gd}$ has a corresponding ground truth value $y_{l}^{gd}$.
	\item The unlabeled training data set ${{\mathcal{D}}_{ce}}$ with  $N_{ce}$ inputs $\bm{x }_{{{l}'}}^{ce}$, i.e., ${{\mathcal{D}}_{ce}}=\left\{ \bm{x }_{{{l}'}}^{ce}\left| {l}'=1,2,\cdots ,{{N}_{ce}} \right. \right\}$. Different with the labeled training data set ${{\mathcal{D}}_{gd}}$, each data of the unlabeled training data set ${{\mathcal{D}}_{ce}}$ only has the input $\bm{x }_{{{l}'}}^{ce}$.
\end{itemize}
In this paper, the proposed semi-supervised cost function $\mathcal{J}\left(\bm{\theta} \right)$ is constructed by the $L_1\text{norm}$. The detailed reason can refer to \ref{appendix_A}.

For the labeled training data set ${{\mathcal{D}}_{gd}}$, each input $ \bm{x }_{l}^{gd} $ is normalized to be $\bm{\xi }_{l}^{gd}$ by Eq.(\ref{normlize}). For $ l=1,2,\cdots ,{N}_{gd} $, and then the corresponding estimated value $\hat{y}_{l}^{gd}\left( \bm{\xi }_{l}^{gd};\bm{\theta } \right)$ can be obtained by the Deep aPCE model $\mathcal{D}\mathcal{A}\mathcal{P}\mathcal{C}\left( \bm{\xi };\bm{\theta } \right)$. Thus, the mean absolute error ${{\mathcal{L}}_{gd}}\left( \bm{\xi }_{l}^{gd},y_{l}^{gd};\bm{\theta } \right)$ between the estimated values $ \hat{y}_{l}^{gd}\left( \bm{\xi }_{l}^{gd};\bm{\theta } \right) $ and the ground truth values $ y_{l}^{gd} $ ($ l=1,2,\cdots ,{N}_{gd} $) is
\begin{equation}\label{MAE_gd}
\begin{aligned}
& {{\mathcal{L}}_{gd}}\left( \bm{\xi }_{l}^{gd},y_{l}^{gd};\bm{\theta } \right)=\frac{1}{{{N}_{gd}}}\sum\limits_{l=1}^{{{N}_{gd}}}{\left\| \hat{y}_{l}^{gd}\left( \bm{\xi }_{l}^{gd};\bm{\theta } \right)-y_{l}^{gd} \right\|}_1 \\ 
& \qquad \qquad \qquad \quad \, =\frac{1}{{{N}_{gd}}}\sum\limits_{l=1}^{{{N}_{gd}}}{\left\| \sum\limits_{i=1}^{M}{{{\mathcal{C}}_{i}}\left( \bm{\xi }_{l}^{gd};\bm{\theta } \right){{\Phi }_{i}}\left( \bm{\xi }_{l}^{gd} \right)}-y_{l}^{gd} \right\|}_1. \\ 
\end{aligned}
\end{equation}

For the unlabeled training data set ${{\mathcal{D}}_{ce}}$, each input $ \bm{x }_{{l}'}^{ce} $ is normalized to be $\bm{\xi }_{{l}'}^{ce}$ by Eq.(\ref{normlize}). For $ l'=1,2,\cdots ,{N}_{ce} $, the normalized input $ \bm{\xi }_{{l}'}^{ce} $ is input into the Deep aPCE model $\mathcal{D}\mathcal{A}\mathcal{P}\mathcal{C}\left( \bm{\xi };\bm{\theta } \right)$, and then $N_{ce}$ sets of expansion coefficients $\left \{ \mathcal{C}_i \left (\bm{\xi}^{ce}_{l'};\bm{\theta}\right )\mid i=1,2,\cdots , M \right \} $ are
\begin{equation}\label{ceff_ce}
\left\{ \begin{matrix}
\left\{ {{\mathcal{C}}_{1}}\left( \bm{\xi }_{1}^{ce};\bm{\theta } \right),{{\mathcal{C}}_{2}}\left( \bm{\xi }_{1}^{ce};\bm{\theta } \right),\cdots ,{{\mathcal{C}}_{M}}\left( \bm{\xi }_{1}^{ce};\bm{\theta } \right) \right\}  \\
\left\{ {{\mathcal{C}}_{1}}\left( \bm{\xi }_{2}^{ce};\bm{\theta } \right),{{\mathcal{C}}_{2}}\left( \bm{\xi }_{2}^{ce};\bm{\theta } \right),\cdots ,{{\mathcal{C}}_{M}}\left( \bm{\xi }_{2}^{ce};\bm{\theta } \right) \right\}  \\
\vdots   \\
\left\{ {{\mathcal{C}}_{1}}\left( \bm{\xi }_{{{l}'}}^{ce};\bm{\theta } \right),{{\mathcal{C}}_{2}}\left( \bm{\xi }_{{{l}'}}^{ce};\bm{\theta } \right),\cdots ,{{\mathcal{C}}_{M}}\left( \bm{\xi }_{{{l}'}}^{ce};\bm{\theta } \right) \right\}  \\
\vdots   \\
\left\{ {{\mathcal{C}}_{1}}\left( \bm{\xi }_{{{N}_{ce}}}^{ce};\bm{\theta } \right),{{\mathcal{C}}_{2}}\left( \bm{\xi }_{{{N}_{ce}}}^{ce};\bm{\theta } \right),\cdots ,{{\mathcal{C}}_{M}}\left( \bm{\xi }_{{{N}_{ce}}}^{ce};\bm{\theta } \right) \right\} . \\
\end{matrix} \right.
\end{equation}
Thereby, the means of expansion coefficients ${{\mathcal{C}}_{i}}\left( {{\bm{\xi }}^{ce}};\bm{\theta } \right)$ ($i=1,2,\cdots ,M$) are
\begin{equation}\label{coeff_mean_ce}
\left\{ \begin{matrix}
E\left[ {{\mathcal{C}}_{1}}\left( {{\bm{\xi }}^{ce}};\bm{\theta } \right) \right]=\frac{1}{{{N}_{ce}}}\sum\limits_{{l}'=1}^{{{N}_{ce}}}{{{\mathcal{C}}_{1}}\left( \bm{\xi }_{{{l}'}}^{ce};\bm{\theta } \right)}  \\
E\left[ {{\mathcal{C}}_{2}}\left( {{\bm{\xi }}^{ce}};\bm{\theta } \right) \right]=\frac{1}{{{N}_{ce}}}\sum\limits_{{l}'=1}^{{{N}_{ce}}}{{{\mathcal{C}}_{2}}\left( \bm{\xi }_{{{l}'}}^{ce};\bm{\theta } \right)}  \\
\vdots   \\
E\left[ {{\mathcal{C}}_{i}}\left( {{\bm{\xi }}^{ce}};\bm{\theta } \right) \right]=\frac{1}{{{N}_{ce}}}\sum\limits_{{l}'=1}^{{{N}_{ce}}}{{{\mathcal{C}}_{i}}\left( \bm{\xi }_{{{l}'}}^{ce};\bm{\theta } \right)}  \\
\vdots   \\
E\left[ {{\mathcal{C}}_{M}}\left( {{\bm{\xi }}^{ce}};\bm{\theta } \right) \right]=\frac{1}{{{N}_{ce}}}\sum\limits_{{l}'=1}^{{{N}_{ce}}}{{{\mathcal{C}}_{M}}\left( \bm{\xi }_{{{l}'}}^{ce};\bm{\theta } \right)} . \\
\end{matrix} \right.
\end{equation}

In addition to the expansion coefficients $\left \{ \mathcal{C}_i \left (\bm{\xi}^{ce}_{l'};\bm{\theta}\right )\mid i=1,2,\cdots , M \right \} $, the corresponding estimated values $\hat{y}_{{l}'}^{ce}\left( \bm{\xi }_{{l}'}^{ce};\bm{\theta } \right)$ ($ l'=1,2,\cdots ,{N}_{ce} $) for the normalized input $ \bm{\xi }_{{l}'}^{ce} $ can be estimated by the Deep aPCE model $\mathcal{D}\mathcal{A}\mathcal{P}\mathcal{C}\left( \bm{\xi };\bm{\theta } \right)$. Thus, the mean and variance of the estimated values $\hat{y}_{l'}^{ce}\left( \bm{\xi }_{l'}^{ce};\bm{\theta } \right)$ ($ l'=1,2,\cdots ,{N}_{ce} $) are calculated respectively by
\begin{equation}\label{y_hat_ce_mean}
\begin{aligned}
& E\left[ {{{\hat{y}}}^{ce}}\left( {{\bm{\xi }}^{ce}};\bm{\theta } \right) \right]=\frac{1}{{{N}_{ce}}}\sum\limits_{{l}'=1}^{{{N}_{ce}}}{\hat{y}_{{{l}'}}^{ce}\left( \bm{\xi }_{{{l}'}}^{ce};\bm{\theta } \right)} \\ 
& \qquad \qquad \qquad =\frac{1}{{{N}_{ce}}}\sum\limits_{{l}'=1}^{{{N}_{ce}}}{\left[ \sum\limits_{i=1}^{M}{{{\mathcal{C}}_{i}}\left( \bm{\xi }_{{{l}'}}^{ce};\bm{\theta } \right){{\Phi }_{i}}\left( \bm{\xi }_{{{l}'}}^{ce} \right)} \right]}, \\ 
\end{aligned}
\end{equation}
and
\begin{equation}\label{y_hat_ce_var}
\begin{aligned}
& D\left[ {{{\hat{y}}}^{ce}}\left( \bm{\xi }^{ce};\bm{\theta } \right) \right]=\frac{1}{{{N}_{ce}}-1}\sum\limits_{{l}'=1}^{{{N}_{ce}}}{{{\left\{ {{{\hat{y}}}^{ce}}\left( \bm{\xi }_{{{l}'}}^{ce};\bm{\theta } \right)-E\left[ {{{\hat{y}}}^{ce}}\left( \bm{\xi }^{ce};\bm{\theta } \right) \right] \right\}}^{2}}} \\ 
& \qquad \qquad \qquad \quad=\frac{1}{{{N}_{ce}}-1}\sum\limits_{{l}'=1}^{{{N}_{ce}}}{{{\left\{ \sum\limits_{i=1}^{M}{{{\mathcal{C}}_{i}}\left( \bm{\xi }_{{{l}'}}^{ce};\bm{\theta } \right){{\Phi }_{i}}\left( \bm{\xi }_{{{l}'}}^{ce} \right)}-\frac{1}{{{N}_{ce}}}\sum\limits_{{l}'=1}^{{{N}_{ce}}}{\left[ \sum\limits_{i=1}^{M}{{{\mathcal{C}}_{i}}\left( \bm{\xi }_{{{l}'}}^{ce};\bm{\theta } \right){{\Phi }_{i}}\left( \bm{\xi }_{{{l}'}}^{ce} \right)} \right]} \right\}}^{2}}} . \\ 
\end{aligned}
\end{equation}

According to the property in Eq.(\ref{aaPC_property}), the absolute error $\mathcal{L}_{ce}^{1}\left( \bm{\xi }_{{{l}'}}^{ce};\bm{\theta } \right)$ between the mean of the first adaptive expansion coefficient $ {{\mathcal{C}}_{1}}\left( {{\bm{\xi }}^{ce}};\bm{\theta } \right) $ and the mean of estimated values $\hat{y}_{{l}'}^{ce}\left( \bm{\xi }_{{l}'}^{ce};\bm{\theta } \right)$ ($ l'=1,2,\cdots ,{N}_{ce} $) is 
\begin{equation}\label{Lce_1}
\begin{aligned}
& \mathcal{L}_{ce}^{1}\left( \bm{\xi }_{{{l}'}}^{ce};\bm{\theta } \right)=\left\| E\left[ {{{\hat{y}}}^{ce}}\left( {{\bm{\xi }}^{ce}};\theta  \right) \right]-E\left[ {{\mathcal{C}}_{1}}\left( {{\bm{\xi }}^{ce}};\bm{\theta } \right) \right] \right\|_1 \\ 
& \qquad \qquad \;\;\, =\left\| \frac{1}{{{N}_{ce}}}\sum\limits_{{l}'=1}^{{{N}_{ce}}}{\hat{y}_{{{l}'}}^{ce}\left( \bm{\xi }_{{{l}'}}^{ce};\bm{\theta } \right)}-\frac{1}{{{N}_{ce}}}\sum\limits_{{l}'=1}^{{{N}_{ce}}}{{{\mathcal{C}}_{1}}\left( \bm{\xi }_{{{l}'}}^{ce};\bm{\theta } \right)} \right\|_1 \\ 
& \qquad \qquad \;\;\, =\frac{1}{{{N}_{ce}}}\left\| \sum\limits_{{l}'=1}^{{{N}_{ce}}}{\left[ \hat{y}_{{{l}'}}^{ce}\left( \bm{\xi }_{{{l}'}}^{ce};\bm{\theta } \right)-{{\mathcal{C}}_{1}}\left( \bm{\xi }_{{{l}'}}^{ce};\bm{\theta } \right) \right]} \right\|_1 \\ 
& \qquad \qquad \;\;\, =\frac{1}{{{N}_{ce}}}\left\| \sum\limits_{{l}'=1}^{{{N}_{ce}}}{\left[ \sum\limits_{i=1}^{M}{{{\mathcal{C}}_{i}}\left( \bm{\xi }_{{{l}'}}^{ce};\bm{\theta } \right){{\Phi }_{i}}\left( \bm{\xi }_{{{l}'}}^{ce} \right)}-{{\mathcal{C}}_{1}}\left( \bm{\xi }_{{{l}'}}^{ce};\bm{\theta } \right) \right]} \right\|_1 . \\ 
\end{aligned}
\end{equation}
According to the property in Eq.(\ref{aaPC_property}), the absolute error $\mathcal{L}_{ce}^{2M}\left( \bm{\xi }_{{{l}'}}^{ce};\bm{\theta } \right)$ between the sum of squares of $E\left[ {{\mathcal{C}}_{i}}\left( {{\bm{\xi }}^{ce}};\bm{\theta } \right) \right]$ ($i=2,\cdots ,M$) and the variance of estimated values $\hat{y}_{{l}'}^{ce}\left( \bm{\xi }_{{l}'}^{ce};\bm{\theta } \right)$ ($ l'=1,2,\cdots ,{N}_{ce} $) is 
\begin{equation}\label{Lce_2}
\begin{aligned}
& \mathcal{L}_{ce}^{2M}\left( \bm{\xi }_{{{l}'}}^{ce};\bm{\theta } \right)=\left\| D\left[ {{{\hat{y}}}^{ce}}\left( {{\bm{\xi }}^{ce}};\theta  \right) \right]-\sum\limits_{i=2}^{M}{{{\left\{ E\left[ {{\mathcal{C}}_{i}}\left( {{\bm{\xi }}^{ce}};\bm{\theta } \right) \right] \right\}}^{2}}} \right\|_1 \\ 
& \qquad \qquad \quad \, =\left\| \frac{1}{{{N}_{ce}}-1}\sum\limits_{{l}'=1}^{{{N}_{ce}}}{{{\left\{ {{{\hat{y}}}^{ce}}\left( \bm{\xi }_{{{l}'}}^{ce};\bm{\theta } \right)-E\left[ {{{\hat{y}}}^{ce}}\left( \bm{\xi }_{{{l}'}}^{ce};\bm{\theta } \right) \right] \right\}}^{2}}}-\sum\limits_{i=2}^{M}{{{\left\{ E\left[ {{\mathcal{C}}_{i}}\left( {{\bm{\xi }}^{ce}};\bm{\theta } \right) \right] \right\}}^{2}}} \right\|_1 \\ 
& \qquad \qquad \quad \, =\left\| \frac{1}{{{N}_{ce}}-1}\sum\limits_{{l}'=1}^{{{N}_{ce}}}{{{\left\{ \sum\limits_{i=1}^{M}{{{\mathcal{C}}_{i}}\left( \bm{\xi }_{{{l}'}}^{ce};\bm{\theta } \right){{\Phi }_{i}}\left( \bm{\xi }_{{{l}'}}^{ce} \right)}-\frac{1}{{{N}_{ce}}}\sum\limits_{{l}'=1}^{{{N}_{ce}}}{\left[ \sum\limits_{i=1}^{M}{{{\mathcal{C}}_{i}}\left( \bm{\xi }_{{{l}'}}^{ce};\bm{\theta } \right){{\Phi }_{i}}\left( \bm{\xi }_{{{l}'}}^{ce} \right)} \right]} \right\}}^{2}}} \right. \\ 
& \qquad \qquad \qquad \; \left. -\sum\limits_{i=2}^{M}{{{\left[ \frac{1}{{{N}_{ce}}}\sum\limits_{{l}'=1}^{{{N}_{ce}}}{{{\mathcal{C}}_{i}}\left( \bm{\xi }_{{{l}'}}^{ce};\bm{\theta } \right)} \right]}^{2}}} \right\|_1 . \\ 
\end{aligned}
\end{equation}
Therefore, the absolute error ${{\mathcal{L}}_{ce}}\left( \bm{\xi }_{{{l}'}}^{ce};\bm{\theta } \right)$ of the unlabeled training data set ${{\mathcal{D}}_{ce}}$ is
\begin{equation}\label{Lce}
{{\mathcal{L}}_{ce}}\left( \bm{\xi }_{{{l}'}}^{ce};\bm{\theta } \right)=\mathcal{L}_{ce}^{1}\left( \bm{\xi }_{{{l}'}}^{ce};\bm{\theta } \right)+\mathcal{L}_{ce}^{2M}\left( \bm{\xi }_{{{l}'}}^{ce};\bm{\theta } \right) + \mathcal{L}_{ce}^{var}\left( \bm{\xi }_{{{l}'}}^{ce};\bm{\theta } \right).
\end{equation}

In summary, the semi-supervised cost function $\mathcal{J}\left( \bm{\theta } \right)$ for learning the shared parameters $\bm{\theta}$ of the Deep aPCE model $\mathcal{D}\mathcal{A}\mathcal{P}\mathcal{C}\left( \bm{\xi };\bm{\theta } \right)$ is 
\begin{equation}\label{cost_fun_L}
\begin{aligned}
& \mathcal{J}\left( \bm{\theta } \right)={{\mathcal{L}}_{gd}}\left( \bm{\xi }_{l}^{gd},y_{l}^{gd};\bm{\theta } \right)+ \lambda {{\mathcal{L}}_{ce}}\left( \bm{\xi }_{{{l}'}}^{ce};\bm{\theta } \right) \\ 
& \qquad \; ={{\mathcal{L}}_{gd}}\left( \bm{\xi }_{l}^{gd},y_{l}^{gd};\bm{\theta } \right)+\lambda \mathcal{L}_{ce}^{1}\left( \bm{\xi }_{{{l}'}}^{ce};\bm{\theta } \right)+\lambda\mathcal{L}_{ce}^{2M}\left( \bm{\xi }_{{{l}'}}^{ce};\bm{\theta } \right),\\ 
\end{aligned}
\end{equation}
where $\lambda$ is a hyperparameter.

\subsection{Deep aPCE model training for uncertainty quantification}\label{sec43}
\begin{figure}[ht]
	\centering
	{\includegraphics[scale=0.8]{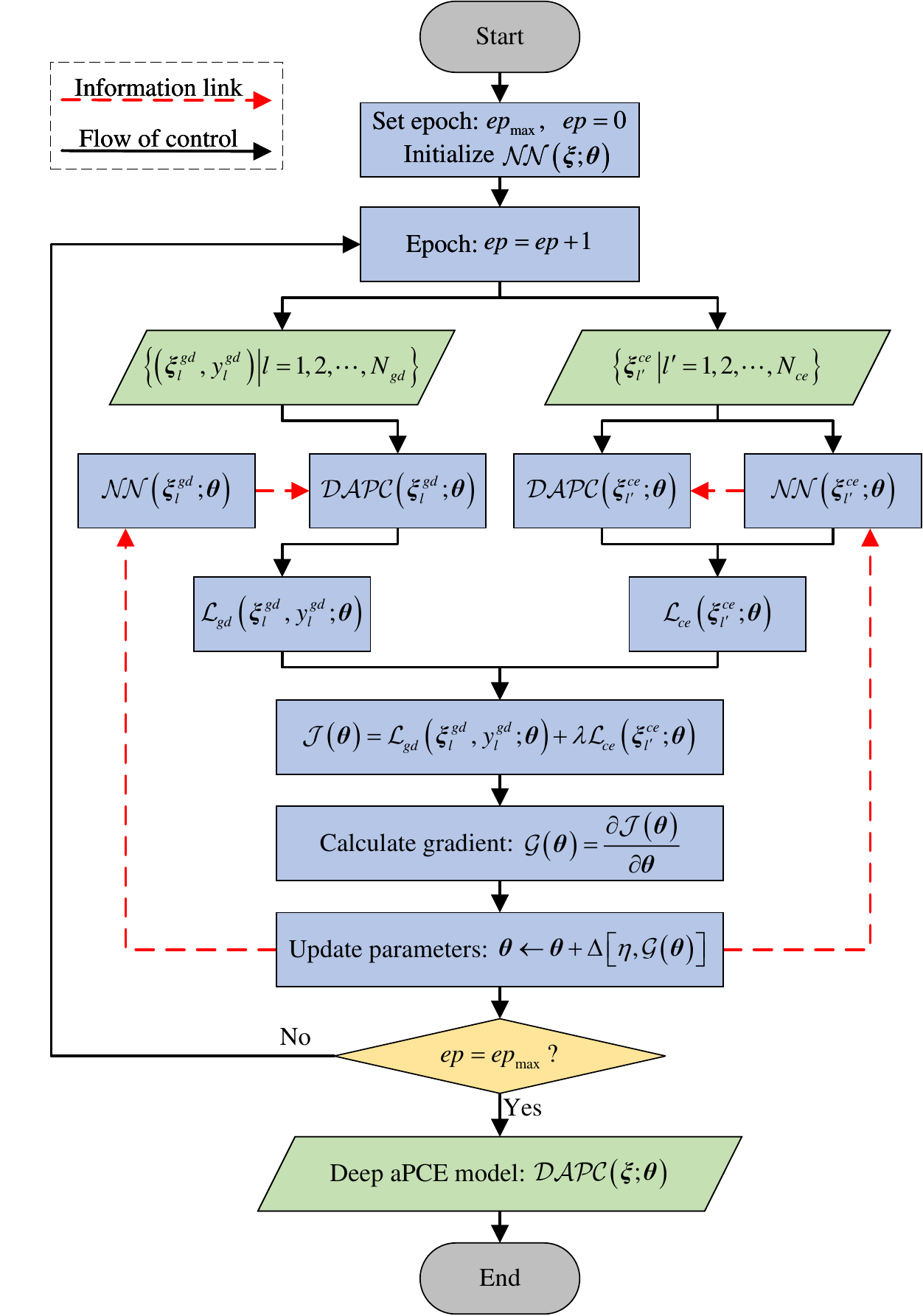}}
	\caption{The training flowchart of Deep aPCE model}\label{model_train}
\end{figure}

This section proposes an iterative training method for learning the shared parameters $\bm{\theta}=\left( \bm{W},\bm{b}\right)$ by minimizing the semi-supervised cost function $\mathcal{J}\left(\bm{\theta}\right)$ in Eq.(\ref{cost_fun_L}), i.e.,
\begin{equation}\label{mini_J_theta}
\bm{\theta }^{*} \gets \bm{\theta }=\underset{\bm{\theta }}{\mathop{\arg }}\,\min \mathcal{J}\left( \bm{\theta } \right)=\underset{\bm{\theta }}{\mathop{\arg }}\,\min \left[ {{\mathcal{L}}_{gd}}\left( \bm{\xi }_{l}^{gd},y_{l}^{gd};\bm{\theta } \right)+\lambda \mathcal{L}_{ce}^{1}\left( \bm{\xi }_{{{l}'}}^{ce};\bm{\theta } \right)+\lambda \mathcal{L}_{ce}^{2M}\left( \bm{\xi }_{{{l}'}}^{ce};\bm{\theta } \right) \right].
\end{equation}
The flowchart of the iterative training method is shown in Fig.\ref{model_train}. Firstly, setting the maximum training epoch $ ep_{max} $, and the theory-informed DNN $\mathcal{N}\mathcal{N}\left( \bm{\xi };\bm{\theta } \right)$ is initialized randomly so that the shared parameters $\bm{\theta}=\left( \bm{W},\bm{b}\right)$ have initial values. Then, given the labeled training data set ${{\mathcal{D}}_{gd}}$ and the unlabeled training data set ${{\mathcal{D}}_{ce}}$, the adaptive expansion coefficients can be obtained by the theory-informed DNN $\mathcal{N}\mathcal{N}\left( \bm{\xi };\bm{\theta } \right)$, based on which the Deep aPCE model $\mathcal{D}\mathcal{A}\mathcal{P}\mathcal{C}\left( \bm{\xi };\bm{\theta } \right)$ calculates the corresponding estimated values $\hat{y}\left( \bm{\xi };\bm{\theta } \right)$. Subsequently, the semi-supervised cost function $\mathcal{J}\left( \bm{\theta } \right)$ is calculated by Eq.(\ref{cost_fun_L}). In DL, the gradient $\mathcal{G}\left( \bm{\theta } \right)$ can be derived by the chain rule for differentiating compositions of functions using automatic differentiation \cite{Baydin2015}. Refer to section \ref{sec21}, the shared parameters $\bm{\theta}=\left( \bm{W},\bm{b}\right)$ are updated by Eq.(\ref{Param_update_MLP}). In this paper, the Adam algorithm \cite{Kingma2014} is selected as the optimization solution algorithm. Finally, the shared parameters $\bm{\theta}=\left( \bm{W},\bm{b}\right)$ of the Deep aPCE model $\mathcal{D}\mathcal{A}\mathcal{P}\mathcal{C}\left( \bm{\xi };\bm{\theta } \right)$ are updated iteratively until $ep=e{{p}_{\max }}$ as shown in Fig.\ref{model_train}. 

After obtaining the trained Deep aPCE model $\mathcal{D}\mathcal{A}\mathcal{P}\mathcal{C}\left( \bm{\xi };\bm{\theta } \right)$, it can estimate the output corresponding to the new random input $\bm{x}$. It is noteworthy that the new random input $\bm{x}$ should be normalized firstly by Eq.(\ref{normlize}). Thus, the uncertainty of stochastic model $Y=f\left( \bm{\xi } \right)$ can be quantified by performing MCS over the Deep aPCE model $\mathcal{D}\mathcal{A}\mathcal{P}\mathcal{C}\left( \bm{\xi };\bm{\theta } \right)$. In this paper, the kernel density estimation (KDE) \cite{Parzen1962, Chen2017} is used to estimate the probability density function of stochastic system response. Besides, the first fourth central moments (mean ${\mu}_Y$, standard deviation ${\sigma}_Y$, skewness ${\gamma}_Y$ and kurtosis ${\kappa}_Y$ are also calculated, i.e.,
\begin{equation}\label{4_moment}
\begin{aligned}
& {{\mu }_{Y}}=\frac{1}{{{N}_{MCS}}}\sum\limits_{l=1}^{{{N}_{MCS}}}{{{{\hat{y}}}^{(p)}}\left( {{\mathbf{\xi }}_{l}};\mathbf{\theta } \right)}, \\ 
& {{\sigma }_{Y}}=\frac{1}{{{N}_{MCS}}-1}\sum\limits_{l=1}^{{{N}_{MCS}}}{{{\left[ {{{\hat{y}}}^{(p)}}\left( {{\mathbf{\xi }}_{l}};\mathbf{\theta } \right)-{{\mu }_{Y}} \right]}^{2}}}, \\ 
& {{\gamma }_{Y}}=\frac{1}{{{N}_{MCS}}}\sum\limits_{l=1}^{{{N}_{MCS}}}{{{\left\{ {\left[ {{{\hat{y}}}^{(p)}}\left( {{\mathbf{\xi }}_{l}};\mathbf{\theta } \right)-{{\mu }_{Y}} \right]}/{{{\sigma }_{Y}}}\; \right\}}^{3}}}, \\ 
& {{\kappa }_{Y}}=\frac{1}{{{N}_{MCS}}}\sum\limits_{l=1}^{{{N}_{MCS}}}{{{\left\{ {\left[ {{{\hat{y}}}^{(p)}}\left( {{\mathbf{\xi }}_{l}};\mathbf{\theta } \right)-{{\mu }_{Y}} \right]}/{{{\sigma }_{Y}}}\; \right\}}^{4}}}. \\ 
\end{aligned}
\end{equation}

\section{Numerical examples}\label{sec5}
In this section, four numerical examples are used for verifying the effectiveness of the proposed semi-supervised Deep aPCE method. For each numerical example, the values of random input variables are sampled by Latin Hypercube Sampling, and a 2-order aPC model is used to construct the Deep aPCE model. Besides, the theory-informed neural network in the Deep aPCE model is built by PyTorch\footnote{https://pytorch.org}. $ReLU\left(x \right)$ \cite{Nair2010, Sun2014} is chosen to be the nonlinear activation function in each hidden layers' neurons for the first four examples, and \textit{Example} 5 uses $GELU\left(x \right)$ \cite{Hendrycks2016} to be the nonlinear activation function in each hidden layers' neurons. Especially, the solving process of the first numerical example will be introduced in detail to demonstrate the usage of the Deep aPCE method. The other four numerical examples' solving processes are similar to the first numerical example. The Gaussian process regression (GPR) method is used to solve \textit{Example 2} and \textit{Example 3} based on the scikit-learn \cite{scikit-learn2011}. Besides, to validate the effectiveness of two absolute errors $\mathcal{L}_{ce}^{1}\left( \bm{\xi }_{{{l}'}}^{ce};\bm{\theta } \right)$ and $\mathcal{L}_{ce}^{2M}\left( \bm{\xi }_{{{l}'}}^{ce};\bm{\theta } \right)$ in training the Deep aPCE model, the second numerical example will compare the accuracies of two Deep aPCE models, obtained by the cost function $\mathcal{J}\left( \bm{\theta } \right)$ and the mean absolute error ${{\mathcal{L}}_{gd}}\left( \bm{\xi }_{l}^{gd},y_{l}^{gd};\bm{\theta } \right)$, respectively. The hyperparameter $\lambda$ is 1.0 for the first four examples and 100.0 for \textit{Example 5}. The relevant codes of the Deep aPCE method by Python are available on this website\footnote{https://github.com/Xiaohu-Zheng/Deep-aPCE}.

\subsection{Example 1: Fortifier's clutch}
\begin{figure}[htbp]
	\centering
	{\includegraphics[scale=1]{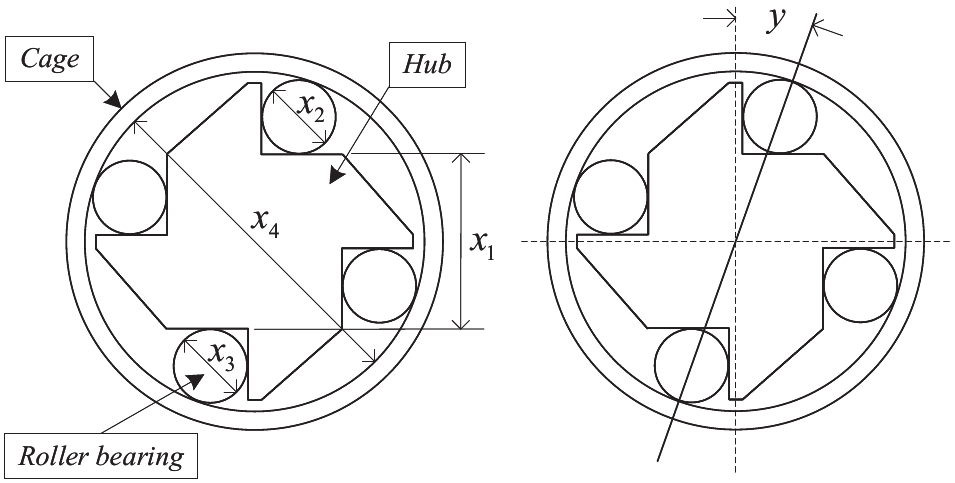}}
	\caption{Fortini's clutch}\label{Fortini}
\end{figure}

The first example is the contact angle $ y $ uncertainty analysis of Fortini's clutch \cite{Singh2007, Lee2009, Lee2019}, as shown in Fig.\ref{Fortini}.

The contact angle $ y $ can be calculated by four independent normally distributed variables $ X_1 $, $ X_2 $, $ X_3 $ and $ X_4 $ as follows:
\begin{equation}\label{Fortini_eq}
y=\arccos \left( \frac{{{X}_{1}}+0.5\left( {{X}_{2}}+{{X}_{3}} \right)}{{{X}_{4}}-0.5\left( {{X}_{2}}+{{X}_{3}} \right)} \right),
\end{equation}
where the distribution parameters of $ X_1 $, $ X_2 $, $ X_3 $ and $ X_4 $ are shown in Table \ref{Param_example_1}.

\begin{table}[htbp]
	\centering
	\caption{Distribution parameters of $ X_1 $, $ X_2 $, $ X_3 $ and $ X_4 $ in the Fortini's clutch example}
	\begin{tabular}{llll}
		\toprule
		Variable & mean (mm) & Variance ($\text{mm}^2$) & Distribution \\
		\midrule
		$X_1$ & 55.29 & $0.0793^2$ & Normal \\
		$X_2$ & 22.86 & $0.0043^2$ & Normal \\
		$X_3$ & 22.86 & $0.0043^2$ & Normal \\
		$X_4$ & 101.6 & $0.0793^2$ & Normal \\
		\bottomrule
	\end{tabular}
	\label{Param_example_1}
\end{table}

\paragraph{\textbf{Deep aPCE modeling}} For this example, the random input variable $\bm{X}=\left\{ {{X}_{1}},{{X}_{2}},{{X}_{3}},{{X}_{4}} \right\}$ is normalized to be $\bm{\xi}=\left\{ {\xi }_{1}, {\xi }_{2}, {\xi }_{3}, {\xi }_{4} \right\}$. Then, for $k=1,2,3,4$, the first three raw moments are calculated by Eq.(\ref{j_moment}), i.e.,
\begin{equation*}
{\mu}_{{\xi}_k}^{(1)}=0 \qquad {\mu}_{{\xi}_k}^{(2)}=1 \qquad {\mu}_{{\xi}_k}^{(3)}=0.
\end{equation*}
Thus, four univariate orthogonal bases (2-order) are constructed by Eq.(\ref{phi_xi_k}) and Table \ref{coeff_table}, i.e.,
\begin{equation*}
\begin{matrix}
\phi _{k}^{(0)}\left( {{\xi }_{k}} \right)=1 & \phi _{k}^{(1)}\left( {{\xi }_{k}} \right)={{\xi }_{k}} & \phi _{k}^{(2)}\left( {{\xi }_{k}} \right)=\xi _{k}^{2}-1 , \\
\end{matrix}
\end{equation*}
where $k=1,2,3,4$. Besides, according to Eqs.(\ref{s_i_k}) and (\ref{si_condition}), there are 15 multivariate index sets, i.e.,
\begin{equation*}
\begin{matrix}
{{s}_{1}}=\left\{ 0,0,0,0 \right\} & {{s}_{2}}=\left\{ 0,0,0,1 \right\} & {{s}_{3}}=\left\{ 0,0,0,2 \right\} & {{s}_{4}}=\left\{ 0,0,1,0 \right\} & {{s}_{5}}=\left\{ 0,0,1,1 \right\}  \\
{{s}_{6}}=\left\{ 0,0,2,0 \right\} & {{s}_{7}}=\left\{ 0,1,0,0 \right\} & {{s}_{8}}=\left\{ 0,1,0,1 \right\} & {{s}_{9}}=\left\{ 0,1,1,0 \right\} & {{s}_{10}}=\left\{ 0,2,0,0 \right\} \\
{{s}_{11}}=\left\{ 1,0,0,0 \right\} & {{s}_{12}}=\left\{ 1,0,0,1 \right\} & {{s}_{13}}=\left\{ 1,0,1,0 \right\} & {{s}_{14}}=\left\{ 1,1,0,0 \right\} & {{s}_{15}}=\left\{ 2,0,0,0 \right\}. \\
\end{matrix}
\end{equation*}
Based on the above results, four-dimensional orthogonal basis can be obtained by Eq.(\ref{multi_basis}), i.e.,
\begin{equation*}
\begin{matrix}
{{\Phi }_{1}}\left( \bm{\xi } \right)=1 & {{\Phi }_{2}}\left( \bm{\xi } \right)={{\xi }_{4}} & {{\Phi }_{3}}\left( \bm{\xi } \right)=(\xi _{4})^{2}-1 & {{\Phi }_{4}}\left( \bm{\xi } \right)={{\xi }_{3}} & {{\Phi }_{5}}\left( \bm{\xi } \right)={{\xi }_{3}}{{\xi }_{4}}  \\
{{\Phi }_{6}}\left( \bm{\xi } \right)=(\xi _{3})^{2}-1 & {{\Phi }_{7}}\left( \bm{\xi } \right)={{\xi }_{2}} & {{\Phi }_{8}}\left( \bm{\xi } \right)={{\xi }_{2}}{{\xi }_{4}} & {{\Phi }_{9}}\left( \bm{\xi } \right)={{\xi }_{2}}{{\xi }_{3}} & {{\Phi }_{10}}\left( \bm{\xi } \right)=(\xi _{2})^{2}-1  \\
{{\Phi }_{11}}\left( \bm{\xi } \right)={{\xi }_{1}} & {{\Phi }_{12}}\left( \bm{\xi } \right)={{\xi }_{1}}{{\xi }_{4}} & {{\Phi }_{13}}\left( \bm{\xi } \right)={{\xi }_{1}}{{\xi }_{3}} & {{\Phi }_{14}}\left( \bm{\xi } \right)={{\xi }_{1}}{{\xi }_{2}} & {{\Phi }_{15}}\left( \bm{\xi } \right)=(\xi _{1})^{2}-1,  \\
\end{matrix}
\end{equation*}
where $\bm{\xi }=\left\{{\xi}_{1}, {\xi}_{2}, {\xi}_{3}, {\xi}_{4} \right\} $. Therefore, the 2-order Deep aPCE model $\mathcal{D}\mathcal{A}\mathcal{P}\mathcal{C}\left( \bm{\xi };\bm{\theta } \right)$ is
\begin{equation*}
{{\hat{y}}^{(2)}}\left( \bm{\xi } \right)=\sum\limits_{i=1}^{15}{{{\mathcal{C}}_{i}}\left( \bm{\xi };\bm{\theta } \right){{\Phi }_{i}}\left( \bm{\xi } \right)}.
\end{equation*}
In this example, a DNN with 4 inputs, 5 hidden layers and 15 outputs is adopted to solve the adaptive expansion coefficients $\left \{ \mathcal{C}_i \left (\bm{\xi};\bm{\theta}\right )\mid i=1,2,\cdots , 15 \right \} $, where the neuron numbers of 5 hidden layers are 64, 128, 256, 128, and 64, respectively.

\paragraph{\textbf{Preparing training data set}} There are two kinds of training data sets, i.e., the labeled training data set ${{\mathcal{D}}_{gd}}=\left\{ \left( \bm{X }_{l}^{gd},y_{l}^{gd} \right)\left| l=1,2,\cdots ,{{N}_{gd}} \right. \right\}$ and the unlabeled training data set ${{\mathcal{D}}_{ce}}=\left\{ \bm{X }_{{{l}'}}^{ce}\left| {l}'=1,2,\cdots ,{{N}_{ce}} \right. \right\}$, where $y_{l}^{gd}$ is calculated by Eq.(\ref{Fortini_eq}) given $\bm{X }_{l}^{gd}$. The inputs $ \bm{X }_{l}^{gd} $ and $ \bm{X }_{{{l}'}}^{ce} $ are normalized to be $ \bm{\xi }_{l}^{gd} $ and $ \bm{\xi }_{{{l}'}}^{ce} $ by Eq.(\ref{normlize}), respectively. In this example, $10^5$ unlabeled training data make up the data set ${{\mathcal{D}}_{ce}}$. Besides, the number $ {N}_{gd} $ of labeled training data is 17, 30, and 40, respectively.

\paragraph{\textbf{Deep aPCE model training}} Based on the labeled training data set ${{\mathcal{D}}_{gd}}=\left\{ \left( \bm{\xi }_{l}^{gd},y_{l}^{gd} \right)\left| l=1,2,\cdots ,N_{gd} \right. \right\}$ ($ N_{gd}=17,30,40 $) and the unlabeled training data set ${{\mathcal{D}}_{ce}}=\left\{ \bm{\xi }_{{{l}'}}^{ce}\left| {l}'=1,2,\cdots ,10^5 \right. \right\}$, the Deep aPCE model is trained. The maximum training epoch $ep_{max}$ is set to be 7000. Besides, the initial learning rate $\eta $ is set to be 0.01. During the model training process, the learning rate $\eta $ is scaled by 0.8 times every 300 epochs.

\paragraph{\textbf{Results analysis}}  Based on the trained Deep aPCE model $\mathcal{D}\mathcal{A}\mathcal{P}\mathcal{C}\left( \bm{\xi };\bm{\theta } \right)$, the uncertainty analysis results of this example are shown in Table \ref{Fortini_table}. In Table \ref{Fortini_table}, the results by the MCS method ($10^6$ runs) are regarded as the true results for comparison. Besides, to validate the effectiveness of Deep aPCE method, this example is also solved by $5^n$ univariate dimension reduction method ($5^n$ UDR) \cite{Lee2009}, 2-order, 3-order, and 4-order PCE method \cite{Xiu2003}, respectively. Compared with the results of the MCS method, ${\varepsilon}_{Pr} $ denote the relative estimation error rate of $Pr\left(y<6 \text{deg} \right)$.

\begin{table}[htbp]
	\centering
	\caption{The uncertainty analysis results of the Fortini’s clutch example}
	\begin{tabular}{lccccccc}
		\toprule
		\multicolumn{1}{l}{Method} &  $N_{gd}$ & Mean    & Standard deviation     & Skewness     & Kurtosis & $Pr\left(y<6 \text{deg} \right)$ & ${\varepsilon}_{Pr} $ \\
		\midrule
		\multicolumn{1}{l}{MCS}                &   $10^6$    & 0.1219 & 0.0118 & $-0.3156$ & 3.2763 & 0.07881 & \multicolumn{1}{c}{-} \\
		$5^n$ UDR                              &     17      & 0.1219 & 0.0117 & $-0.1498$ & 3.0669 & 0.07452 & $5.4435\%$ \\
		2-order PCE                            &     81      & 0.1219 & 0.0119 & $-0.3062$ & 3.1324 & 0.08125 & $3.0961\%$ \\
		3-order PCE                            &     256     & 0.1219 & 0.0118 & $-0.3411$ & 3.3112 & 0.07858 & $0.2918\%$ \\
		4-order PCE                            &     625     & 0.1219 & 0.0118 & $-0.3164$ & 3.3361 & 0.07712 & $2.1444\%$ \\
		\midrule
		\multicolumn{1}{l}{\textbf{Deep aPCE}} & \textbf{17} & 0.1219 & 0.0118 & $-0.2737$ & 3.0950 & \textbf{0.07857} & $\bm{0.2969\%}$ \\
		                                       & \textbf{30} & 0.1219 & 0.0118 & $-0.2829$ & 3.1316 & \textbf{0.07872} & $\bm{0.1053\%}$ \\
		                                       & \textbf{40} & 0.1219 & 0.0118 & $-0.3204$ & 3.1809 & \textbf{0.07889} & $\bm{0.1015\%}$ \\
		\bottomrule
		\multicolumn{7}{l}{The units of mean and standard deviation are both 'rad'.}
	\end{tabular}
	\label{Fortini_table}
\end{table}

According to Table \ref{Fortini_table}, all methods can calculate the mean of contact angle $y$ correctly compared with the mean of the MCS method. When $N_{gd}=17$, the standard deviation calculated by the Deep aPCE method is 0.118, consistent with the MCS method's results. However, the standard deviation of the $5^n$ UDR is 0.117 rather than 0.118. Besides, the skewness and kurtosis of the Deep aPCE method are closer to the MCS method's results than the $5^n$ UDR method. Apparently, the accuracy of the Deep aPCE method is more exact than the $5^n$ UDR method when $N_{gd}=17$. When $N_{gd}=30$, the estimation error rate ${\varepsilon}_{Pr}=0.1053\% $ of the Deep aPCE method is already smaller than all other methods (The smallest ${\varepsilon}_{Pr}$ value of other methods is $0.2918\%$ calculated by 3-order PCE method (256 data)). When $N_{gd}=40$, the estimation error rate ${\varepsilon}_{Pr}=0.1015\% $ of the Deep aPCE method is the smallest among all methods. In summary, on the one hand, with the same amount of labeled training data, the Deep aPCE method's accuracy is higher than the existing PCE methods; on the other hand, the Deep aPCE method needs much less labeled training data to construct a high-precision surrogate model than the original PCE methods.

\subsection{Example 2: Elastic cantilever beam}\label{sec52}
\begin{figure}[htb]
	\centering
	{\includegraphics[scale=1]{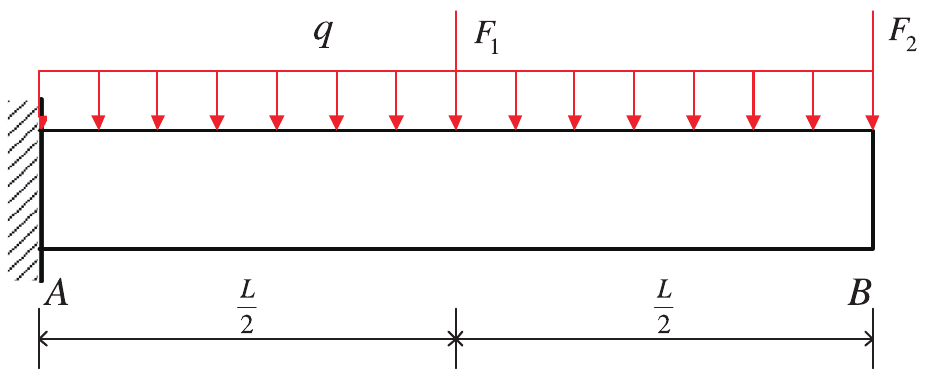}}
	\caption{Elastic cantilever beam}\label{cantilever}
\end{figure}

The second example considers an elastic cantilever beam \cite{Zhang2021, Fan2016} subjected to two concentrated loads and one distributed load as shown in Fig.\ref{cantilever}. In this example, the displacement of point $B$ is of interest, and its limit state function with respect to the displacement threshold ${{\Delta }_{\lim }}$ is
\begin{equation}\label{Example_eq_2}
G\left( q,{{F}_{1}},{{F}_{2}},E,I,L,{{\Delta }_{\lim }} \right)={{\Delta }_{\lim }}-\left( \frac{q{{L}^{4}}}{8EI}+\frac{5{{F}_{1}}{{L}^{3}}}{48EI}+\frac{{{F}_{2}}{{L}^{3}}}{3EI} \right),
\end{equation}
where the statistical properties of random variables $ q $, $ {F}_{1} $, $ {F}_{2} $, $ E $, $ I $, $ L $, and $ {\Delta }_{\lim } $ are shown in Table \ref{Param_example_2}.

\begin{table}[htb]
	\centering
	\caption{Statistical properties of random variables in the elastic cantilever beam example \cite{Zhang2021}}
	\begin{tabular}{lllll}
		\toprule
		Variable & Description & Mean  & C.O.V. & Distribution \\
		\midrule
		$q$ & Distributed load & 50.0 N/mm & 0.15  & Gumbel \\
		$F_1$ & Concentrated load & $7.0\times {{10}^{4}}$ N & 0.18  & Gumbel \\
		$F_2$ & Concentrated load & $1.0\times {{10}^{5}}$ N & 0.20  & Gumbel \\
		$E$ & Elastic modules & $2.6\times {{10}^{5}}$ MPa & 0.12  & Lognormal \\
		$I$ & Moment of inertia & $5.3594\times {{10}^{8}}$ $\text{mm}^4$ & 0.10  & Normal \\
		$L$ & Length of beam & $3.0\times {{10}^{3}}$ mm & 0.05  & Normal \\
		${\Delta}_{lim}$ & Threshold & 30.0 mm & 0.30  & Lognormal \\
		\bottomrule
		\multicolumn{5}{l}{C.O.V. = the coefficient of variation} \\
	\end{tabular}
	\label{Param_example_2}
\end{table}

\paragraph{\textbf{Solving Deep aPCE model}} In this elastic cantilever beam example, a DNN with 7 inputs and 36 outputs is adopted to solve the expansion coefficients $\left \{ \mathcal{C}_i \left (\bm{\xi};\bm{\theta}\right )\mid i=1,2,\cdots , 36 \right \} $, and the neuron numbers of 5 hidden layers are 64, 128, 256, 256, and 256, respectively. In this example, $10^5$ unlabeled training data make up data set ${{\mathcal{D}}_{ce}}$, and the number $ {N}_{gd} $ of labeled training data is 40, 50, 60, 70, 80, and 90, respectively. Besides, the maximum training epoch $ep_{max}$ is set to be 7000, and the initial learning rate $\eta $ is set to be 0.01. During the model training process, the learning rate $\eta $ is scaled by 0.7 times every 300 epochs.

\paragraph{\textbf{Results analysis}}  Based on the trained Deep aPCE model $\mathcal{D}\mathcal{A}\mathcal{P}\mathcal{C}\left( \bm{\xi };\bm{\theta } \right)$, the uncertainty analysis results of this example are shown in Table \ref{cantilever_results}. The results by MCS method ($10^7$ runs) are regarded as the truth results for comparison. Besides, to validate the effectiveness of the Deep aPCE method, this example is also solved by the 3-order CDA-DRM-SPCE (CD-SPCE) method \cite{Zhang2021}, the 3-order LARS-PCE method \cite{Zhang2021}, 3-order, the 4-order and 5-order CDA-DRM-FPCE (CD-FPCE) methods \cite{Zhang2021}, the 2-order OLS-PCE method \cite{Zhang2021}, and the GPR method, respectively.

\begin{table}[!htbp]
	\centering
	\caption{The uncertainty analysis results of the elastic cantilever beam example}
	\begin{tabular}{llllll}
		\toprule
		\multicolumn{1}{l}{Method} &  $N_{gd}$ & Mean (R.E.)    & S.D. (R.E.) & Skewness (R.E.) & Kurtosis (R.E.) \\
		\midrule
		MCS                  & $10^7$  &   18.0946    &   9.5305    &   0.7507    &  4.2713 \\
		3-order CD-SPCE &   88    &   18.1026 (0.044\%)    &   9.4971 (0.35\%)    &   0.7347 (2.13\%)    &  4.1541 (2.74\%) \\
		3-order LARS-PCE     &   98    &   18.1093 (0.081\%)   &   9.4744 (0.59\%)   &   0.7315 (2.56\%)   &  4.0851 (4.36\%) \\
		3-order CD-FPCE &   98    &   18.0909 (0.020\%)   &   9.5044 (0.27\%)   &   0.7318 (2.52\%)   &  4.1519 (2.80\%) \\
		4-order CD-FPCE &   168   &   18.0865 (0.045\%)   &   9.5047 (0.27\%)   &   0.7308 (2.65\%)   &  4.1533 (2.76\%) \\
		2-order OLS-PCE      &   168   &   18.0845 (0.056\%)   &   9.4593 (0.75\%)   &   0.6772 (9.79\%)   &  3.7715 (11.70\%) \\
		5-order CD-FPCE &   280   &   18.0865 (0.045\%)    &   9.5051 (0.27\%)    &   0.7306 (2.68\%)    &  4.1533 (2.76\%) \\
		\midrule
		GPR  &   40   &   18.1902 (0.53\%)   &   9.2662 (2.78\%)   &   0.8038 (7.07\%)   &  4.4325 (3.78\%) \\
		&   50    &   18.0865 (0.045\%)   &   9.4786 (0.54\%)   &   0.7790 (3.77\%)   &  4.3229 (1.21\%) \\
		&   60    &   18.2093 (0.63\%)   &   9.5076 (0.24\%)   &   0.7350 (2.09\%)   &  4.2327 (0.90\%) \\
		&   70    &   18.1395 (0.25\%)   &   9.4624 (0.71\%)   &   0.7699 (2.56\%)   &  4.2812 (0.23\%) \\
		&   80    &   18.1508 (0.31\%)   &   9.4726 (0.61\%)   &   0.7727 (2.92\%)   &  4.2852 (0.33\%) \\
		&   90    &   18.1119 (0.10\%)   &   9.4851 (0.48\%)   &   0.7608 (1.34\%)   &  4.2775 (0.15\%) \\
		\midrule
		\textbf{Deep aPCE}   &   40    &   18.0546 (0.22\%)   &   9.5270 (0.037\%)   &   0.7345 (2.16\%)   &  4.2370 (0.80\%) \\
		&   50    &   18.0799 (0.081\%)    &   9.5093 (0.22\%)    &   0.7570 (0.84\%)    &  4.2706 (0.016\%) \\
		&   60    &   18.0642 (0.17\%)    &   9.5256 (0.051\%)   &   0.7505 (0.026\%)   &  4.2633 (0.19\%) \\
		&   \textbf{70}   &   \textbf{18.0965} (0.010\%)   &   \textbf{9.5020} (0.30\%)   &   \textbf{0.7614} (1.43\%)   &  \textbf{4.2837} (0.29\%) \\
		&   \textbf{80}    &   \textbf{18.1046} (0.055\%)   &   \textbf{9.5188} (0.12\%)   &   \textbf{0.7589} (1.08\%)   &  \textbf{4.2712} (0.0013\%) \\
		&  \textbf{90}   &   \textbf{18.0914} (0.018\%)   &   \textbf{9.5175} (0.14\%)   &   \textbf{0.7478} (0.39\%)   &  \textbf{4.2511} (0.47\%) \\
		\bottomrule
		\multicolumn{6}{l}{R.E. = Relative error. \qquad S.D. = Standard deviation \qquad The units of mean and S.D. are both 'mm'.} \\
	\end{tabular}
	\label{cantilever_results}
\end{table}

Refer to Table \ref{cantilever_results}, compared with the results of the MCS method, the relative errors of the Deep aPCE method are 0.22\% on mean, 0.037\% on standard deviation, 2.16\% on skewness and 0.80\% on kurtosis for $N_{gd}=40$, and the relative errors of 3-order CD-SPCE method are 0.044\% on mean, 0.35\% on standard deviation, 2.13\% on skewness and 2.74\% on kurtosis for $N_{gd}=88$. Except that the relative error on mean and skewness of the Deep aPCE method is slightly larger than the 3-order CD-SPCE method, the relative errors of the Deep aPCE method are far less than the 3-order CD-SPCE method. What's more, it is noteworthy that the amount of labeled training data used by the Deep aPCE method (40 data) are less than half that of the 3-order CD-SPCE method (88 data). As the amount of labeled training data increases, the accuracy of the Deep aPCE method is improved gradually. Especially when $N_{gd}=70$, the Deep aPCE method's relative errors are 0.010\% on mean, 0.30\% on standard deviation, 1.43\% on skewness, and 0.29\% on kurtosis. For 5-order CD-FPCE method ($N_{gd}=280$), the relative errors are 0.045\% on mean, 0.27\% on standard deviation, 2.68\% on skewness and 2.76\% on kurtosis. The relative error on the standard deviation of the Deep aPCE method is only 0.03\% larger than the 3-order CD-SPCE method. However, the relative errors of the Deep aPCE method are much smaller than the 5-order CD-FPCE method for mean, skewness, and kurtosis. Besides, the most important is that the amount of labeled training data used by the Deep aPCE method is only a quarter of that of the 5-order CD-FPCE method. In summary, the Deep aPCE method needs much less labeled training data to construct a high-precision surrogate model than the original PCE methods in this elastic cantilever beam example.

\begin{figure}[!htb]
	\centering
	{\includegraphics[scale=0.65]{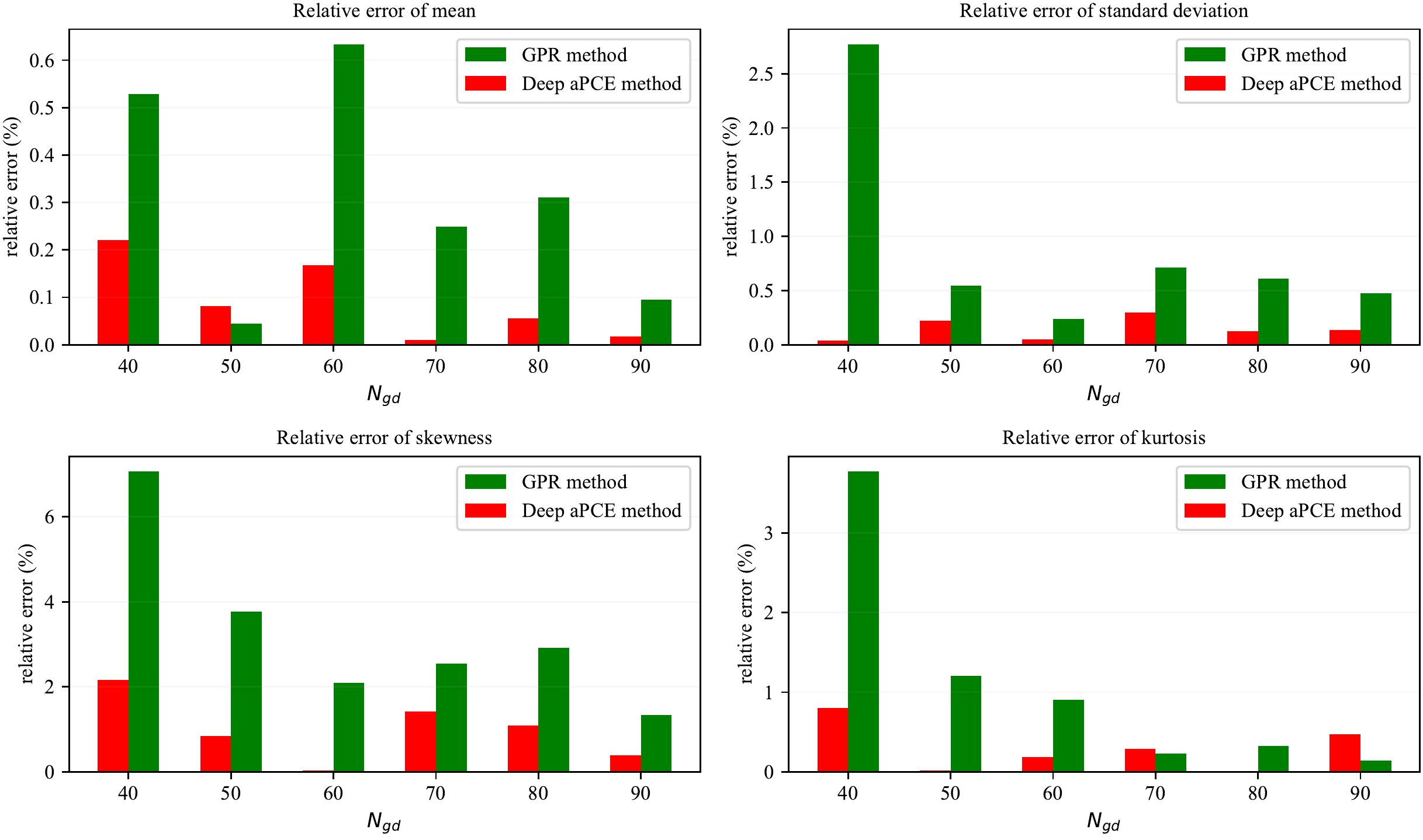}}
	\caption{Compared with the result of the MC method, the relative errors of four statistical moments by the Deep aPCE method and the GPR method for $N_{gd}=40, 50, 60, 70, 80, 90$ in the elastic cantilever beam example.}\label{error_cantilever}
\end{figure}

\begin{figure}[!htb]
	\centering
	{\includegraphics[scale=0.8]{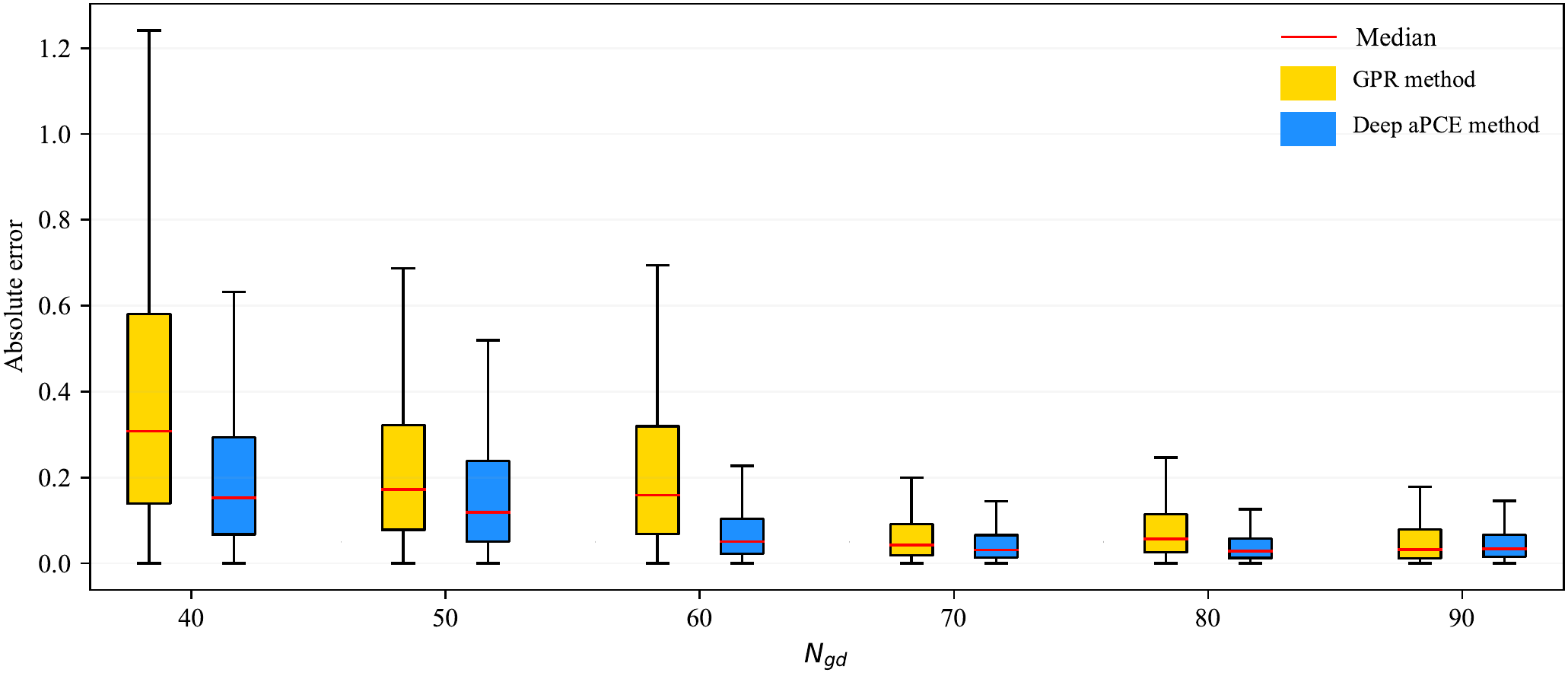}}
	\caption{The absolute error boxplots of the Deep aPCE method and the GPR method in the elastic cantilever beam example.}\label{Error_box_canti}
\end{figure}

\begin{figure}[!htb]
	\centering
	{\includegraphics[scale=0.75]{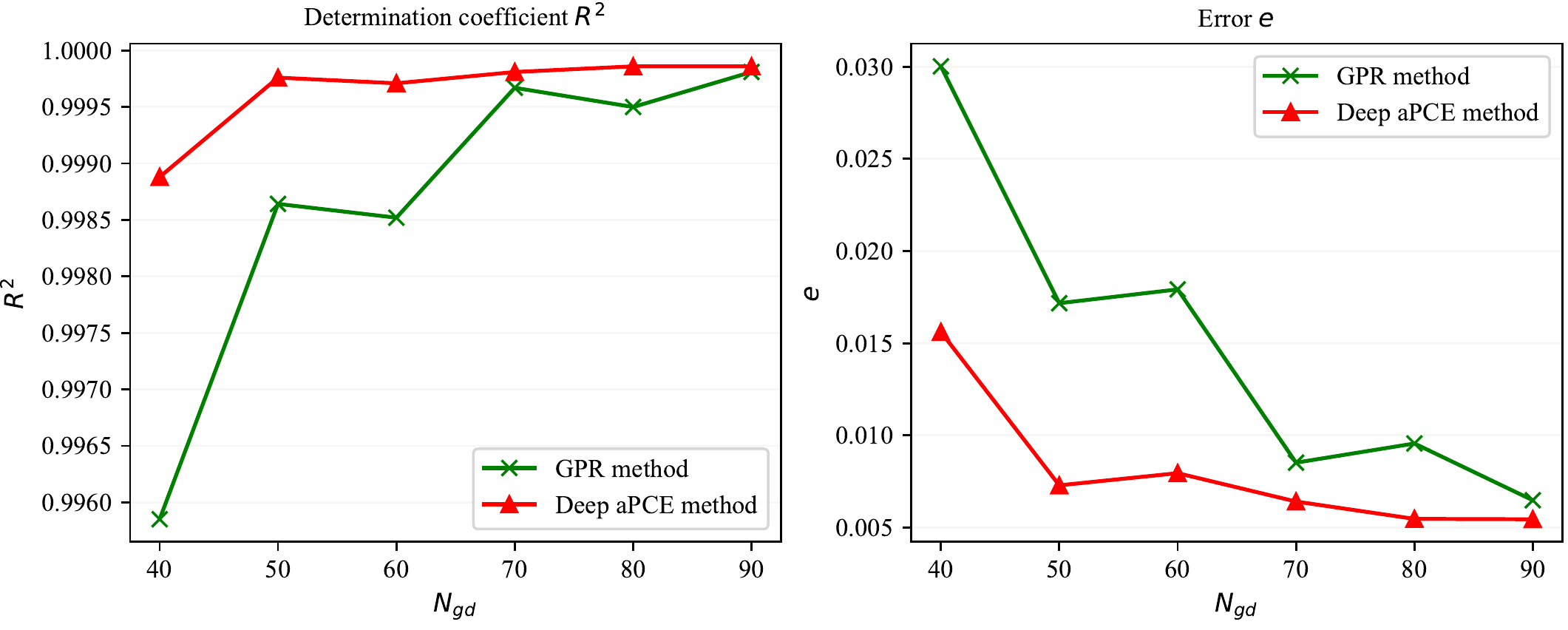}}
	\caption{The determination coefficients $R^2$ and the errors $e$ of the Deep aPCE method and the GPR method in the elastic cantilever beam example.}\label{R2_e_cantilever}
\end{figure}

\paragraph{\textbf{Performance comparison between the Deep aPCE method and the GPR method}} Compared with the results of MCS, the relative errors of four statistical moments by the Deep aPCE method are smaller than the GPR method, as shown in Fig.\ref{error_cantilever} (in addition to the kurtosis for $N_{gd}=\{70,90\}$ and the mean for $N_{gd}=50$). Besides, the absolute error boxplots of the Deep aPCE method and the GPR method are shown in Fig.\ref{Error_box_canti}. For $N_{gd}=40, 50, 60, 70, 80, 90$, the determination coefficients $R^2$ and the errors $e$ of the Deep aPCE method and the GPR method are shown in Fig.\ref{R2_e_cantilever}. Apparently, the determination coefficients $R^2$ of the Deep aPCE method are always closer to 1 than the GPR method, and the errors $e$ of the Deep aPCE method are always closer to 0 than the GPR method. Combining  Fig.\ref{error_cantilever}, Fig.\ref{Error_box_canti} and Fig.\ref{R2_e_cantilever}, the proposed Deep aPCE method can construct a more accurate surrogate model than the GPR method in the elastic cantilever beam example.

\begin{figure}[!htb]
	\centering
	{\includegraphics[scale=0.75]{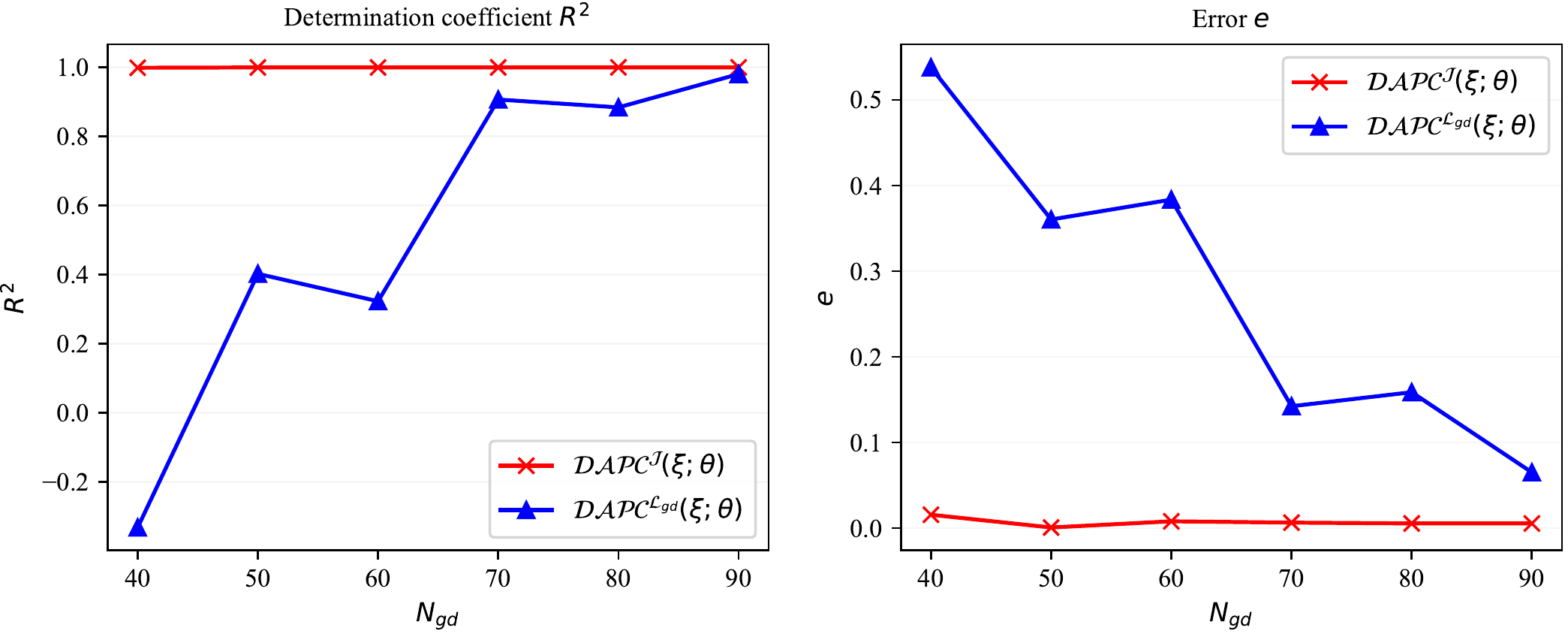}}
	\caption{The determination coefficients $R^2$ and the errors $e$ of the Deep aPCE models $\mathcal{DAPC}^{\mathcal{J}}\left( \bm{\xi };\bm{\theta } \right)$ and $\mathcal{DAPC}^{\mathcal{L}_{gd}}\left( \bm{\xi };\bm{\theta } \right)$ in the elastic cantilever beam example}\label{R2_e}
\end{figure}

\begin{figure}[!h]
	\centering
	{\includegraphics[scale=0.58]{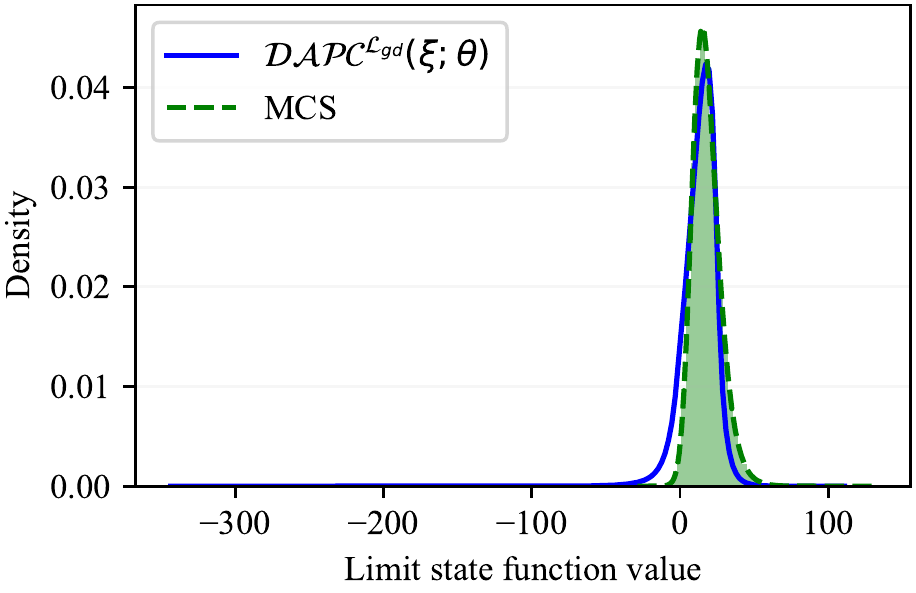}\label{Cantilever_pdfL40}}
	{\includegraphics[scale=0.58]{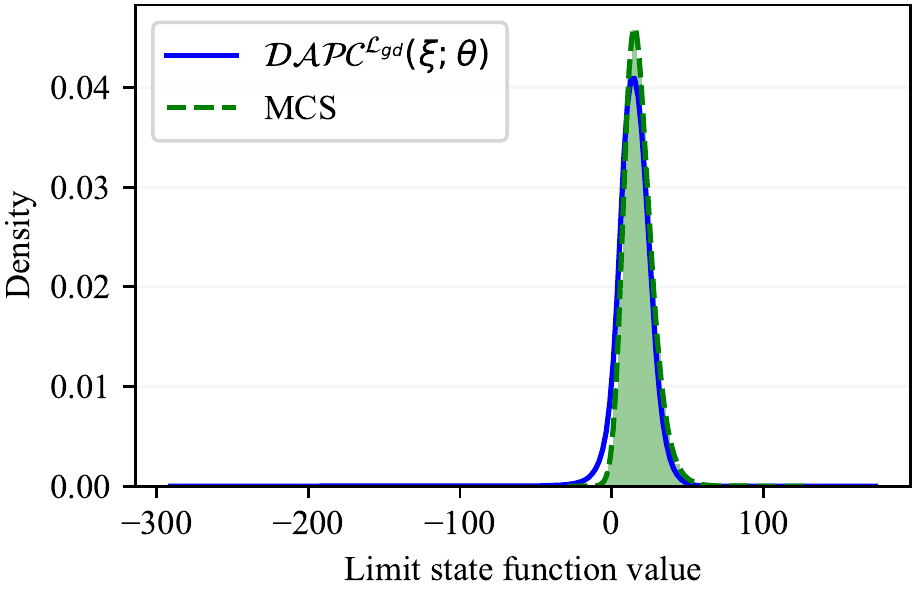}\label{Cantilever_pdfL50}}
	{\includegraphics[scale=0.58]{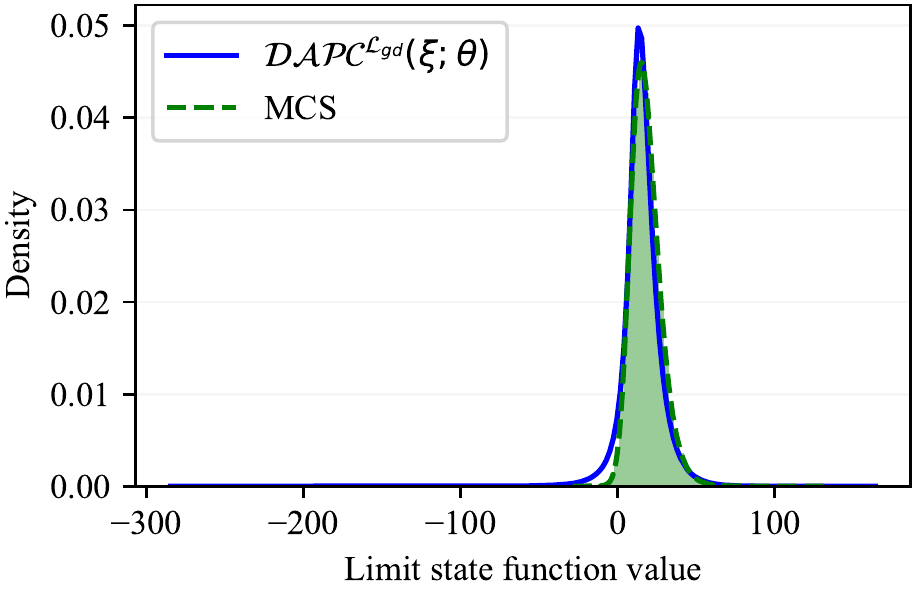}\label{Cantilever_pdfL60}}
	\subfigure [40 labeled training data]
	{\includegraphics[scale=0.58]{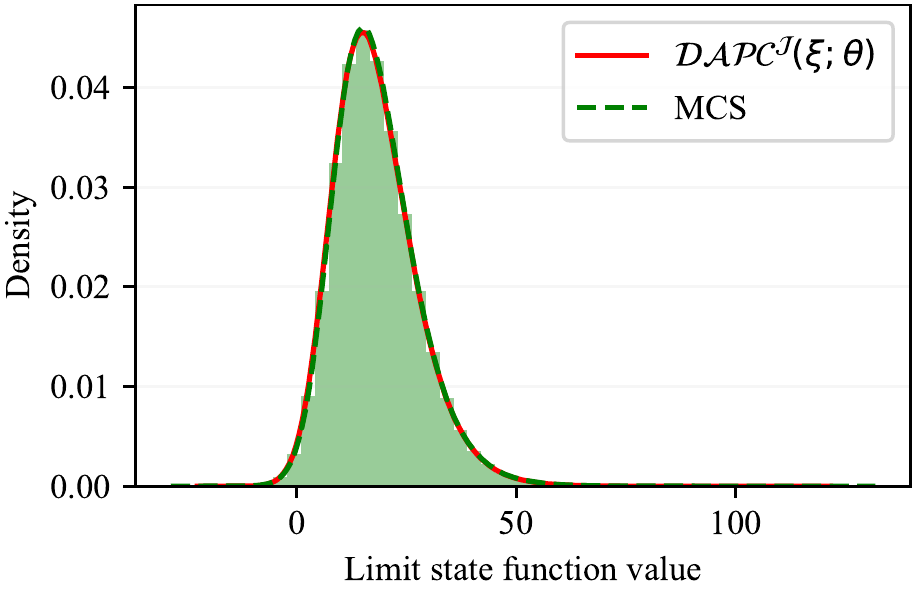}\label{Cantilever_pdfJ40}}
	\subfigure [50 labeled training data]
	{\includegraphics[scale=0.58]{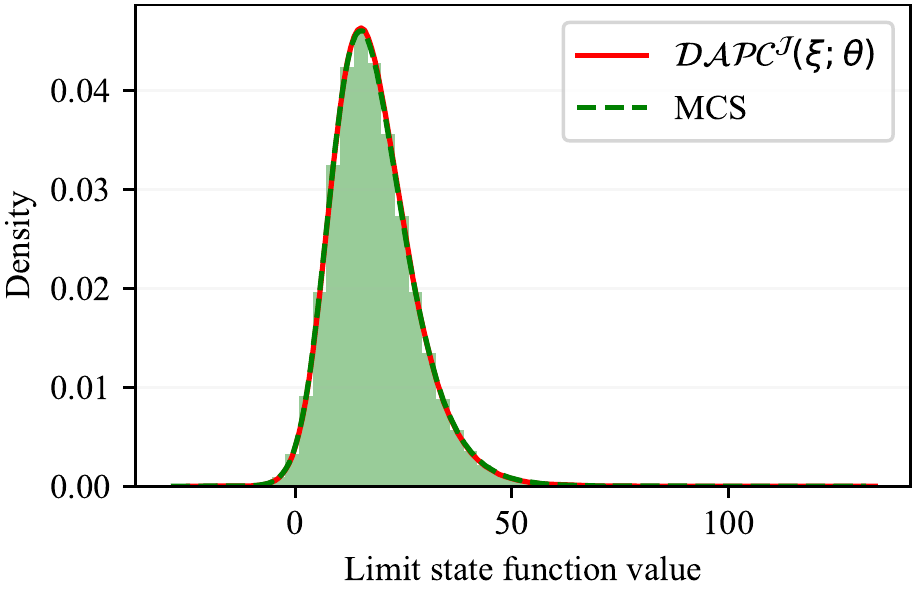}\label{Cantilever_pdfJ50}}
	\subfigure [60 labeled training data]
	{\includegraphics[scale=0.58]{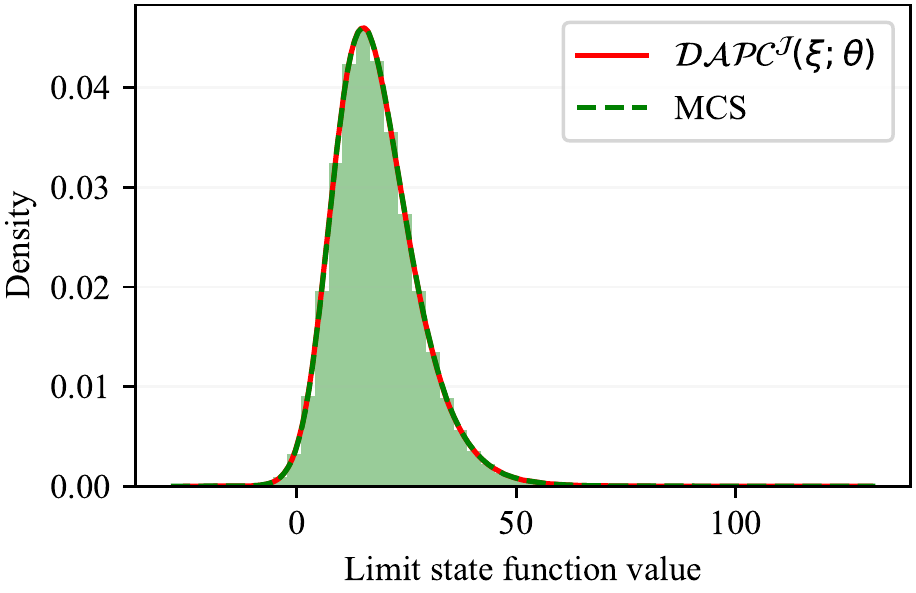}\label{Cantilever_pdfJ60}}
	\caption{The estimated probability density functions of the limit state function value by the Deep aPCE models $\mathcal{DAPC}^{\mathcal{J}}\left( \bm{\xi };\bm{\theta } \right)$ and $\mathcal{DAPC}^{\mathcal{L}_{gd}}\left( \bm{\xi };\bm{\theta } \right)$ (40, 50, and 60 labeled training data) and the MCS method for the elastic cantilever beam example. The blue lines in the first row's three figures are the results obtained by $\mathcal{DAPC}^{\mathcal{L}_{gd}}\left( \bm{\xi };\bm{\theta } \right)$, and the red lines in the second row's three figures are the results obtained by $\mathcal{DAPC}^{\mathcal{J}}\left( \bm{\xi };\bm{\theta } \right)$.}\label{Cantilever_J_L}
\end{figure}

\paragraph{\textbf{Effectiveness analysis of the proposed cost function}} To validate the effectiveness of two absolute errors $\mathcal{L}_{ce}^{1}\left( \bm{\xi }_{{{l}'}}^{ce};\bm{\theta } \right)$ and $\mathcal{L}_{ce}^{2M}\left( \bm{\xi }_{{{l}'}}^{ce};\bm{\theta } \right)$ in training the Deep aPCE model, this numerical example compares two Deep aPCE models $\mathcal{DAPC}^{\mathcal{J}}\left( \bm{\xi };\bm{\theta } \right)$ and $\mathcal{DAPC}^{\mathcal{L}_{gd}}\left( \bm{\xi };\bm{\theta } \right)$, where $\mathcal{DAPC}^{\mathcal{J}}\left( \bm{\xi };\bm{\theta } \right)$ is trained by the cost function $\mathcal{J}\left( \bm{\theta } \right)$ based on both the labeled training data and the unlabeled training data, and $\mathcal{DAPC}^{\mathcal{L}_{gd}}\left( \bm{\xi };\bm{\theta } \right)$ is trained by the mean absolute error ${{\mathcal{L}}_{gd}}\left( \bm{\xi }_{l}^{gd},y_{l}^{gd};\bm{\theta } \right)$ based on the labeled training data only. Compared with the results of MCS method, the relative errors of two models' uncertainty analysis results are shown in Table \ref{Cost_compare_results}. Apparently, for $N_{gd}=40,50,60,70,80,90$, the relative errors on mean, standard deviation, skewness and kurtosis of the model $\mathcal{DAPC}^{\mathcal{J}}\left( \bm{\xi };\bm{\theta } \right)$ are far less than the model $\mathcal{DAPC}^{\mathcal{L}_{gd}}\left( \bm{\xi };\bm{\theta } \right)$. Besides, the determination coefficients $R^2$ and the errors $e$ of the Deep aPCE models $\mathcal{DAPC}^{\mathcal{J}}\left( \bm{\xi };\bm{\theta } \right)$ and $\mathcal{DAPC}^{\mathcal{L}_{gd}}\left( \bm{\xi };\bm{\theta } \right)$ are shown in Fig.\ref{R2_e}. For $N_{gd}=40,50,60,70,80,90$, the determination coefficients $R^2$ of the Deep aPCE model $\mathcal{DAPC}^{\mathcal{J}}\left( \bm{\xi };\bm{\theta } \right)$ are always closer to 1 than the model $\mathcal{DAPC}^{\mathcal{L}_{gd}}\left( \bm{\xi };\bm{\theta } \right)$, and the errors $e$ of the Deep aPCE model $\mathcal{DAPC}^{\mathcal{J}}\left( \bm{\xi };\bm{\theta } \right)$ are always closer to 0 than the model $\mathcal{DAPC}^{\mathcal{L}_{gd}}\left( \bm{\xi };\bm{\theta } \right)$. Based on KDE, the estimated PDFs of limit state function value by the Deep aPCE models $\mathcal{DAPC}^{\mathcal{J}}\left( \bm{\xi };\bm{\theta } \right)$ and $\mathcal{DAPC}^{\mathcal{L}_{gd}}\left( \bm{\xi };\bm{\theta } \right)$ (40, 50, and 60 labeled training data) are shown in Fig.\ref{Cantilever_J_L}, where the green curves (dash line) are the results of the MCS method. According to Fig.\ref{Cantilever_J_L}, the PDF curves of the limit state function value by model $\mathcal{DAPC}^{\mathcal{J}}\left( \bm{\xi };\bm{\theta } \right)$ are closer to the MCS method's results than model $\mathcal{DAPC}^{\mathcal{L}_{gd}}\left( \bm{\xi };\bm{\theta } \right)$.

\begin{table}[htbp]
	\centering
	\caption{The uncertainty analysis results' relative errors of the Deep aPCE models $\mathcal{DAPC}^{\mathcal{J}}\left( \bm{\xi };\bm{\theta } \right)$ and $\mathcal{DAPC}^{\mathcal{L}_{gd}}\left( \bm{\xi };\bm{\theta } \right)$ in the elastic cantilever beam example}
	\begin{tabular}{lllcll}
		\toprule
		$N_{gd}$ & Model  & Mean    & Standard deviation & Skewness & Kurtosis \\
		\midrule
		40    &  $\mathcal{DAPC}^{\mathcal{L}_{gd}}\left( \bm{\xi };\bm{\theta } \right)$  &   $ 26.65\%$    &   $ 17.56\%$    &   $ 263.33\%$    &   $ 128.77\%$ \\
		      &  $\bm{\mathcal{DAPC}}^{\mathcal{J}}\left( \bm{\xi };\bm{\theta } \right)$       &   $ \textbf{0.22\%}$    &   $ \textbf{0.037\%}$    &   $ \textbf{2.16\%}$    &   $\textbf{0.80\%}$  \\
		\midrule
		50    &  $\mathcal{DAPC}^{\mathcal{L}_{gd}}\left( \bm{\xi };\bm{\theta } \right)$  &   $ 16.99\%$    &   $ 12.61\%$    &   $ 182.31\%$    &   $ 110.10\%$ \\
		      &  $\bm{\mathcal{DAPC}}^{\mathcal{J}}\left( \bm{\xi };\bm{\theta } \right)$       &   $ \textbf{0.081\%}$    &   $ \textbf{0.22\%}$    &   $ \textbf{0.84\%}$    &   $ \textbf{0.016\%}$ \\
		\midrule
		60    &  $\mathcal{DAPC}^{\mathcal{L}_{gd}}\left( \bm{\xi };\bm{\theta } \right)$  &   $ 14.40\%$    &   $ 23.49\%$    &   $ 138.87\%$    &   $ 121.10\%$ \\
		      &  $\bm{\mathcal{DAPC}}^{\mathcal{J}}\left( \bm{\xi };\bm{\theta } \right)$       &   $ \textbf{0.17\%}$    &   $ \textbf{0.051\%}$    &   $ \textbf{0.026\%}$    &   $ \textbf{0.19\%}$ \\
		\midrule
		70    &  $\mathcal{DAPC}^{\mathcal{L}_{gd}}\left( \bm{\xi };\bm{\theta } \right)$  &   $ 6.08\%$    &   $ 3.74\%$    &   $ 23.93\%$    &   $ 7.17\%$ \\
		      &  $\bm{\mathcal{DAPC}}^{\mathcal{J}}\left( \bm{\xi };\bm{\theta } \right)$       &   $ \textbf{0.010\%}$    &   $ \textbf{0.30\%}$    &   $ \textbf{1.43\%}$    &   $ \textbf{0.29\%}$ \\
		\midrule
		80    &  $\mathcal{DAPC}^{\mathcal{L}_{gd}}\left( \bm{\xi };\bm{\theta } \right)$  &   $ 7.86\%$    &   $ 3.25\%$    &   $ 31.30\%$    &   $ 14.96\%$ \\
		      &  $\bm{\mathcal{DAPC}}^{\mathcal{J}}\left( \bm{\xi };\bm{\theta } \right)$       &   $ \textbf{0.055\%}$    &   $ \textbf{0.12\%}$    &   $ \textbf{1.08\%}$    &   $ \textbf{0.0013\%}$ \\
		\midrule
		90    &  $\mathcal{DAPC}^{\mathcal{L}_{gd}}\left( \bm{\xi };\bm{\theta } \right)$  &   $ 2.88\%$    &   $ 1.51\%$    &   $ 10.53\%$    &   $ 10.48\%$ \\
		      &  $\bm{\mathcal{DAPC}}^{\mathcal{J}}\left( \bm{\xi };\bm{\theta } \right)$       &   $ \textbf{0.018\%}$    &   $ \textbf{0.14\%}$    &   $ \textbf{0.39\%}$    &   $ \textbf{0.47\%}$ \\
		\bottomrule
	\end{tabular}
	\label{Cost_compare_results}
\end{table}

In summary, according to the results of Table \ref{Cost_compare_results}, Fig.\ref{R2_e} and Fig.\ref{Cantilever_J_L}, the accuracy of the Deep aPCE model $\mathcal{DAPC}^{\mathcal{J}}\left( \bm{\xi };\bm{\theta } \right)$ trained by the cost function $\mathcal{J}\left( \bm{\theta } \right)$ are higher than the model $\mathcal{DAPC}^{\mathcal{L}_{gd}}\left( \bm{\xi };\bm{\theta } \right)$ trained by the mean absolute error ${{\mathcal{L}}_{gd}}\left( \bm{\xi }_{l}^{gd},y_{l}^{gd};\bm{\theta } \right)$. According to Eq.(\ref{mini_J_theta}), the cost function $\mathcal{J}\left( \bm{\theta } \right)$ is composed of ${{\mathcal{L}}_{gd}}\left( \bm{\xi }_{l}^{gd},y_{l}^{gd};\bm{\theta } \right)$, $\mathcal{L}_{ce}^{1}\left( \bm{\xi }_{{{l}'}}^{ce};\bm{\theta } \right)$ and $\mathcal{L}_{ce}^{2M}\left( \bm{\xi }_{{{l}'}}^{ce};\bm{\theta } \right)$, where $\mathcal{L}_{ce}^{1}\left( \bm{\xi }_{{{l}'}}^{ce};\bm{\theta } \right)$ and $\mathcal{L}_{ce}^{2M}\left( \bm{\xi }_{{{l}'}}^{ce};\bm{\theta } \right)$ are calculated by the unlabeled training data based on the properties of adaptive aPC. Therefore, only using a small amount of labeled training data, the applications of two absolute errors $\mathcal{L}_{ce}^{1}\left( \bm{\xi }_{{{l}'}}^{ce};\bm{\theta } \right)$ and $\mathcal{L}_{ce}^{2M}\left( \bm{\xi }_{{{l}'}}^{ce};\bm{\theta } \right)$ in training the Deep aPCE model can obtain an accurate surrogate model.

\subsection{Example 3: Three-bay six-storey planar frame}
\begin{figure}[!htb]
	\centering
	{\includegraphics[scale=1]{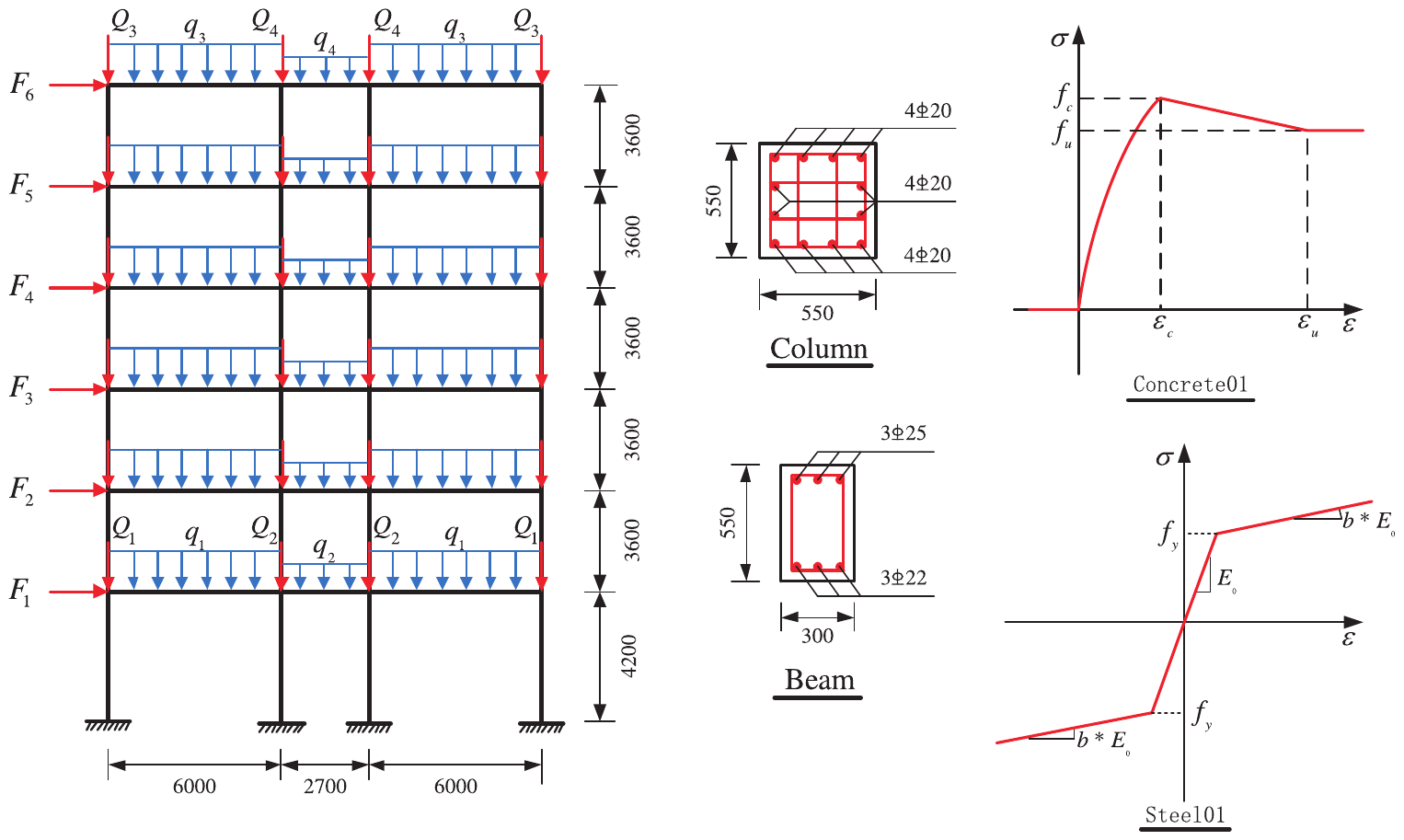}}
	\caption{Three-bay six-storey planar frame \cite{Xu2020}}\label{frame}
\end{figure}

As shown in Fig.\ref{frame}, the third example considers a three-bay six-storey planar reinforced concrete frame structure with nonlinear constitutive laws. Same as the setting of reference \cite{Xu2020, Zhang2021}, the lateral top displacement $\Delta \left( \bm{X} \right)$ of the three-bay six-storey planar reinforced concrete frame structure is calculated by the Opensees software. The uni-axial material Concrete01 and Steel01 describe the nonlinear constitutive laws of concrete and rebar, respectively, as shown in Fig.\ref{frame}. The limit state function with respect to the displacement threshold ${{\Delta }_{\lim }}$ is
\begin{equation}\label{Example_eq_3}
G\left( \bm{X} \right)={{\Delta }_{\lim }}-\Delta \left( \bm{X} \right),
\end{equation}
where ${{\Delta }_{\lim }}=0.06\text{m}$, and $ \bm{X} $ are the random variables (including 24 independent random variables) of which the statistical properties are shown in Table \ref{Param_example_3}.

\begin{table}[!htb]
	\centering
	\caption{Statistical properties of random variables in the three-bay six-storey planar frame}
	\begin{tabular}{lllll}
		\toprule
		Variable & Description & Mean  & C.O.V. & Distribution \\
		\midrule
		$f_{cc}$ & Confined concrete compressive strength & 35 MPa & 0.10  & Normal \\
		${\varepsilon}_{cc}$ & Confined concrete strain at maximum strength & 0.005 & 0.05  & Normal \\
		$f_{cu}$ & Confined concrete crushing strength & 25 MPa & 0.10  & Normal \\
		${\varepsilon}_{cu}$ & Confined concrete strain at crushing strength & 0.02 & 0.05  & Normal \\
		$f_{c}$ & Unconfined concrete compressive strength & 27 MPa & 0.10  & Normal \\
		$f_{u}$ & Unconfined concrete crushing strength & 0 MPa & -  & Deterministic \\
		${\varepsilon}_{c}$ & Unconfined concrete strain at maximum strength & 0.002 & 0.05  & Normal \\
		${\varepsilon}_{u}$ & Unconfined concrete strain at crushing strength & 0.006 & 0.05  & Normal \\
		$f_{y}$ & Yield strength of rebar & 400 MPa & 0.10  & Normal \\
		$E_{0}$ & Initial elastic modulus of rebar & 200 GPa & 0.10  & Normal \\
		$b$ & Strain-hardening ratio of rebar & 0.007 & 0.05  & Normal \\
		$q_1$ & Uniform load & 21.41 kN/m & 0.25  & Lognormal \\
		$q_2$ & Uniform load & 11.48 kN/m & 0.25  & Lognormal \\
		$q_3$ & Uniform load & 22.68 kN/m & 0.25  & Lognormal \\
		$q_4$ & Uniform load & 12.18 kN/m & 0.25  & Lognormal \\
		$Q_1$ & External force & 42.43 kN & 0.25  & Lognormal \\
		$Q_2$ & External force & 68.25 kN & 0.25  & Lognormal \\
		$Q_3$ & External force & 44.03 kN & 0.25  & Lognormal \\
		$Q_4$ & External force & 71.35 kN & 0.25  & Lognormal \\
		$F_1$ & External force & 10 kN & 0.25  & Lognormal \\
		$F_2$ & External force & 20 kN & 0.25  & Lognormal \\
		$F_3$ & External force & 30 kN & 0.25  & Lognormal \\
		$F_4$ & External force & 40 kN & 0.25  & Lognormal \\
		$F_5$ & External force & 50 kN & 0.25  & Lognormal \\
		$F_6$ & External force & 60 kN & 0.25  & Lognormal \\
		\bottomrule
	\end{tabular}
	\label{Param_example_3}
\end{table}

\paragraph{\textbf{Solving Deep aPCE model}} In this three-bay six-storey planar frame example, a DNN with 24 inputs and 325 outputs is adopted to solve the adaptive expansion coefficients $\left \{ \mathcal{C}_i \left (\bm{\xi};\bm{\theta}\right )\mid i=1,2,\cdots , 325 \right \} $, and the neuron numbers of 5 hidden layers are 64, 128, 256, 512, and 512, respectively. In this example, $6\times10^5$ unlabeled training data make up the data set ${{\mathcal{D}}_{ce}}$, and the number $ {N}_{gd} $ of labeled training data is 350, 370, and 390, respectively. Besides, the maximum training epoch $ep_{max}$ is set to be 7000, and the initial learning rate $\eta $ is set to be 0.01. During the model training process, the learning rate $\eta $ is scaled by 0.7 times every 300 epochs.

\paragraph{\textbf{Results analysis}} Based on the trained Deep aPCE model $\mathcal{D}\mathcal{A}\mathcal{P}\mathcal{C}\left( \bm{\xi };\bm{\theta } \right)$, the uncertainty analysis results of this example are shown in Table \ref{frame3_results}. The results by the MCS method ($5\times10^5$ runs) are regarded as the truth results for comparison. Besides, to validate the effectiveness of the Deep aPCE method, this example is also solved by the CDA-DRM-FPCE (CD-FPCE) method \cite{Zhang2021} and the GPR method, respectively.

\begin{table}[htbp]
	\centering
	\caption{The uncertainty analysis results of the three-bay six-storey planar reinforced concrete frame example}
	\begin{tabular}{lllcll}
		\toprule
		\multicolumn{1}{l}{Method} &  $N_{gd}$ & Mean    & Standard deviation     & Skewness (R.E.)     & Kurtosis (R.E.) \\
		\midrule
		\multicolumn{1}{l}{MCS} &   $5\times10^5$    &   0.0209    &   0.0067    &   $-0.6644$    & 3.8514 \\
		CD-FPCE     &   398    &   0.0209    &   0.0067    &   $-0.6121$ (7.88\%)    & 3.7481 (2.68\%) \\
		\multicolumn{1}{l}{GPR} &   350    &   0.0209    &   0.0067    &   $-0.6234$ (6.17\%)   & 3.6884 (4.23\%) \\
		&   370    &   0.0209    &    0.0067   &   $-0.6336$ (4.64\%)    & 3.7117 (3.63\%) \\
		&   390    &   0.0209    &   0.0067    &   $-0.6379$ (3.80\%)    & 3.7201 (3.41\%) \\
		\midrule
		\multicolumn{1}{l}{\textbf{Deep aPCE}} &   350    &   0.0209    &   0.0067    &   $-0.6529$ (1.73\%)    & 3.7611 (2.34\%) \\
		&   370    &   0.0209    &    0.0067   &   $-0.6522$ (1.84\%)    & 3.7716 (2.07\%) \\
		&   390    &   0.0209    &   0.0067    &   $-0.6569$ (1.14\%)    & 3.7810 (1.83\%) \\
		\bottomrule
		\multicolumn{6}{l}{The units of mean and standard deviation are both 'm'.}
	\end{tabular}
	\label{frame3_results}
\end{table}

\begin{figure}[!htb]
	\centering
	\subfigure [350 labeled training data]
	{\includegraphics[scale=0.58]{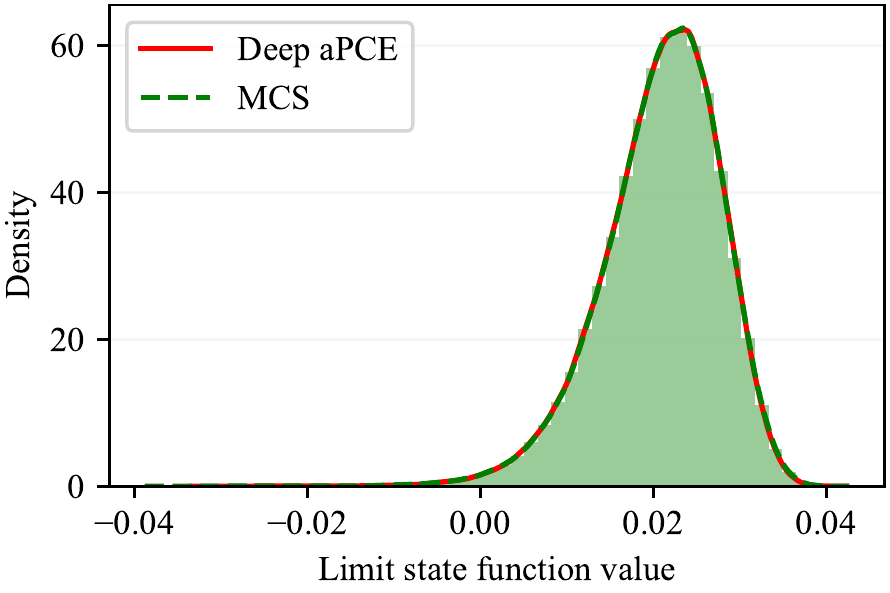}\label{Frame_pdf350}}
	\subfigure [370 labeled training data]
	{\includegraphics[scale=0.58]{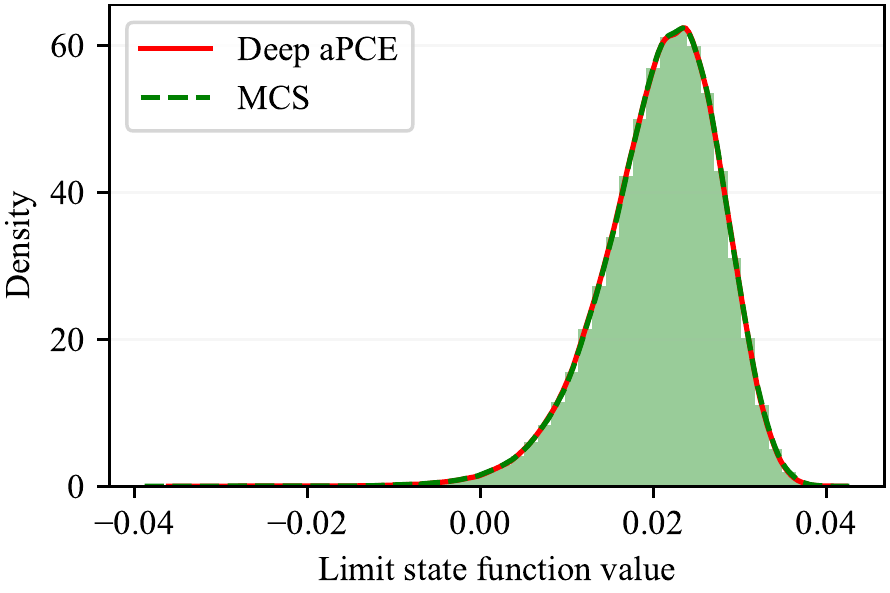}\label{Frame_pdf370}}
	\subfigure [390 labeled training data]
	{\includegraphics[scale=0.58]{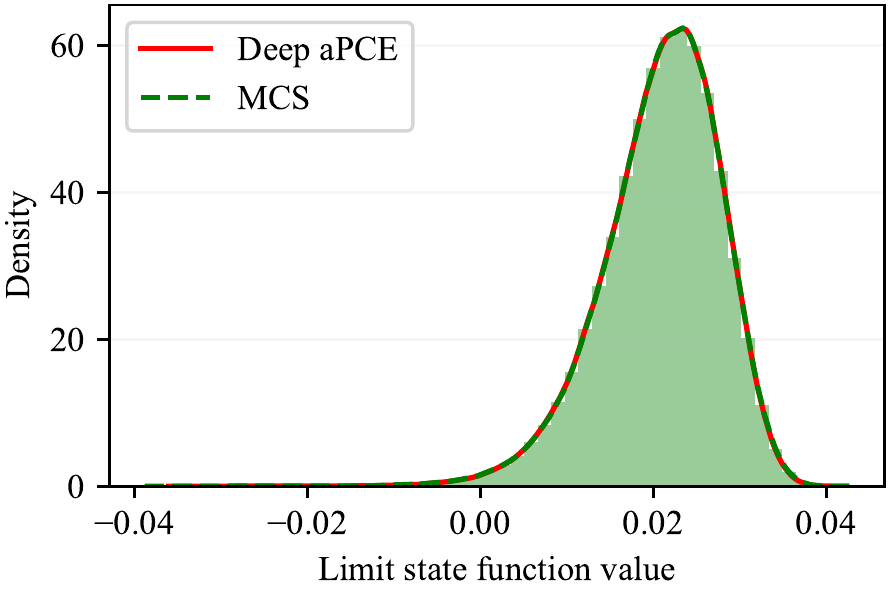}\label{Frame_pdf390}}
	\caption{The estimated probability density functions of the limit state function value by the Deep aPCE method (350, 370, and 390 labeled training data) and the MCS method for the three-bay six-storey planar frame example.}\label{Frame_pdf}
\end{figure}

Refer to Table \ref{frame3_results}, both the Deep aPCE method and the CD-FPCE method can calculate the mean and standard deviation of the limit state function value correctly for compared with the results of MCS. However, the relative errors of the Deep aPCE method are 1.73\% on skewness and 2.34\% on kurtosis for $N_{gd}=350$, the relative errors of CD-FPCE method are 7.88\% on skewness and 2.68\% on kurtosis for $N_{gd}=398$. Apparently, the Deep aPCE method uses only 350 labeled training data to obtain higher accuracy than the CD-FPCE method (398 labeled training data). What's more, the CD-FPCE method needs complex dimension-reduction and model decomposition operations while the proposed Deep aPCE method does not. As the amounts of the labeled training data increase, the accuracy of the Deep aPCE method is improved gradually. For $N_{gd}=390$, the relative errors of the Deep aPCE method are only 1.14\% on skewness and 1.83\% on kurtosis. Besides, based on KDE, the estimated PDFs of limit state function value by the Deep aPCE method (350, 370, and 390 labeled training data) and the MCS method are shown in Fig.\ref{Frame_pdf}. The red curves (solid line) are the results of the Deep aPCE method, and the green curves (dash line) are the results of the MCS method. According to Fig.\ref{Frame_pdf}, the PDF curves of the limit state function value by the Deep aPCE method are very closer to the MCS method's results for different amounts of labeled training data. Thus, the Deep aPCE method can accurately approximate the limit state function with less labeled training data than the CD-FPCE method.

\begin{figure}[!htb]
	\centering
	{\includegraphics[scale=0.6]{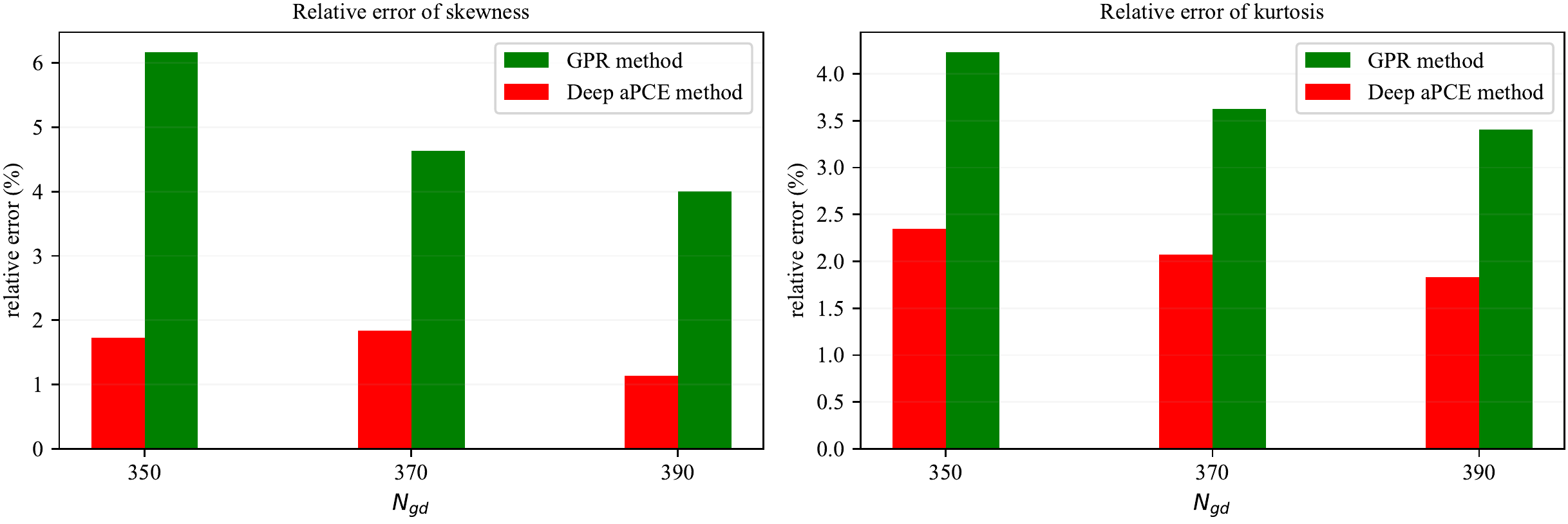}}
	\caption{Compared with the result of the MC method, the relative errors of skewness and kurtosis by the Deep aPCE method and the GPR method in the three-bay six-storey planar frame example}\label{error_frame}
\end{figure}

\begin{figure}[!htb]
	\centering
	{\includegraphics[scale=0.8]{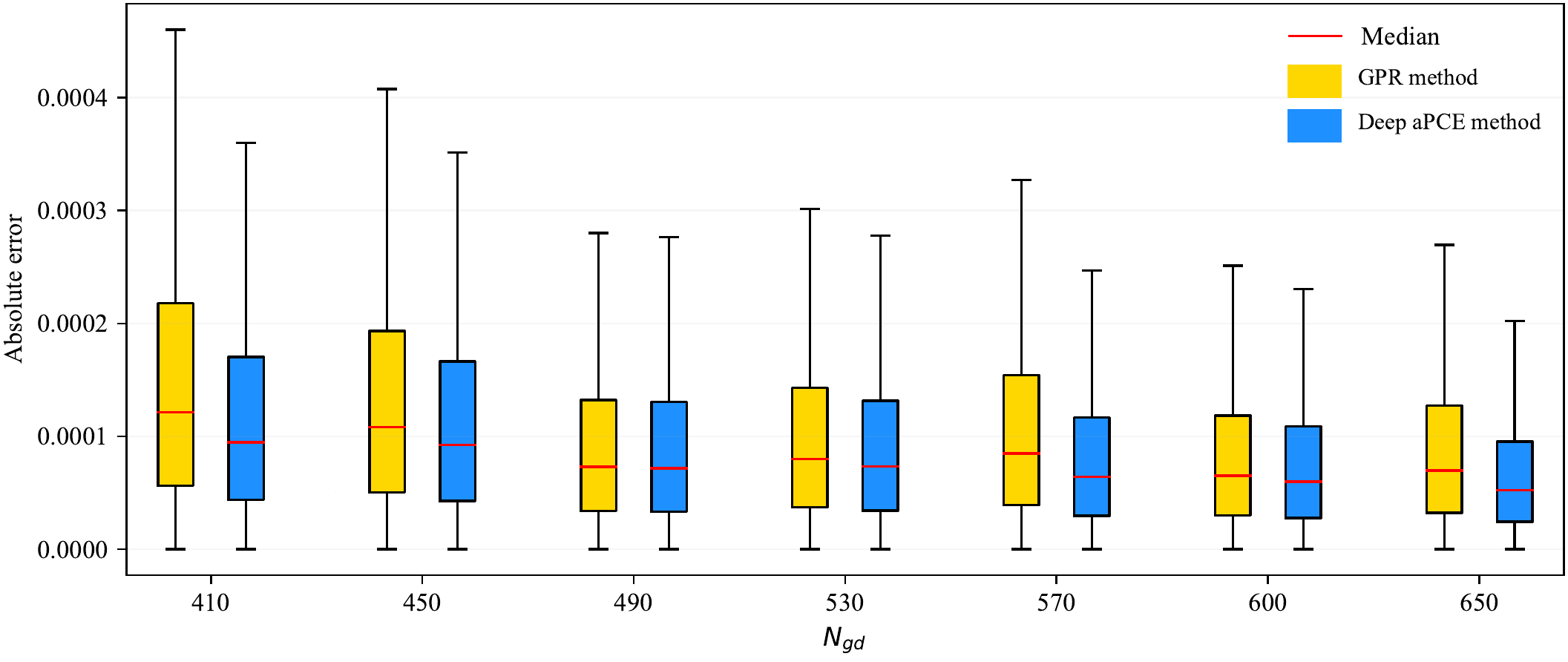}}
	\caption{The absolute error boxplots of the Deep aPCE method and the GPR method in the three-bay six-storey planar frame example.}\label{Error_box_frame}
\end{figure}

\begin{figure}[!htb]
	\centering
	{\includegraphics[scale=0.75]{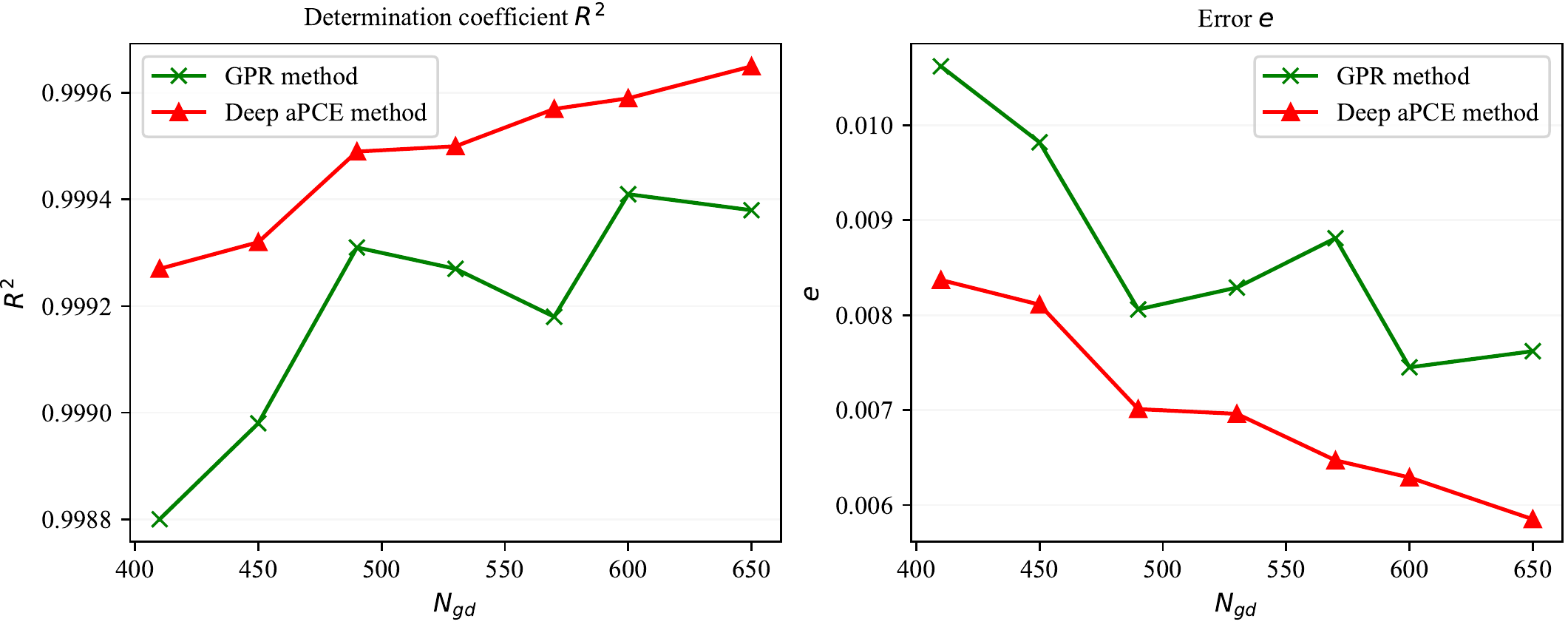}}
	\caption{The determination coefficients $R^2$ and the errors $e$ of the Deep aPCE method and the GPR method in the three-bay six-storey planar frame example.}\label{R2_e_frame}
\end{figure}

\paragraph{\textbf{Performance comparison between the Deep aPCE method and the GPR method}} Compared with the results of MCS, both the Deep aPCE method and the GPR method can calculate the mean and the standard deviation of the limit state function value correctly for $N_{gd}=\left\{350, 370, 390 \right\}$. However, the relative errors of skewness and kurtosis by the Deep aPCE method are smaller than the GPR method as shown in Fig.\ref{error_frame}. For $N_{gd}=410, 450, 490, 530, 570, 600, 650$, the absolute error boxplots of the Deep aPCE method and the GPR method are shown in Fig.\ref{Error_box_frame}. Besides, the determination coefficients $R^2$ and the errors $e$ of the Deep aPCE method and the GPR method are shown in Fig.\ref{R2_e_frame}. Apparently, the determination coefficients $R^2$ of the Deep aPCE method are always closer to 1 than the GPR method, and the errors $e$ of the Deep aPCE method are always closer to 0 than the GPR method. Combining Fig.\ref{error_frame}, Fig.\ref{Error_box_frame} and Fig.\ref{R2_e_frame}, the proposed Deep aPCE method can construct more accurate surrogate model than the GPR method in the three-bay six-storey planar frame example.

\subsection{Example 4: High-dimensional example}
The fourth example investigates the uncertainty quantification of a high-dimensional problem which was proposed by Rackwitz \cite{Rackwitz2001}. The analytical limit state function is
\begin{equation}\label{hi-dim}
G\left( {{x}_{1}},{{x}_{2}},\cdots ,{{x}_{n}} \right)=n+3\sigma \sqrt{n}-\sum\limits_{i=1}^{n}{{{x}_{i}}},
\end{equation}
where the random variables $x_i$ ($i=1,2, \cdots, n$) are independent, identically and lognormally distributed. In this paper, $n=\left\{40, 100\right\}$, and the means and standard deviations are ${{\mu }_{{{x}_{i}}}}=1$ and ${{\sigma }_{{{x}_{i}}}}=0.2$ ($i=1,2, \cdots, n$), respectively. In this example, if the value of the limit state function is less than 0, the failure will happen.

\paragraph{\textbf{Solving Deep aPCE model}} In this high-dimensional example, a DNN with 40 inputs and 861 outputs is adopted to solve the adaptive expansion coefficients $\left \{ \mathcal{C}_i \left (\bm{\xi};\bm{\theta}\right )\mid i=1,2,\cdots , 861 \right \} $, and the neuron numbers of 5 hidden layers are 100, 600, 800, 900, and 900, respectively. In this example, $5\times10^4$ unlabeled training data make up the data set ${{\mathcal{D}}_{ce}}$, and the number $ {N}_{gd} $ of labeled training data is 1200, 1300, 1400, 1500, 1600, 1700, 1800, 1900, and 2000, respectively. Besides, the maximum training epoch $ep_{max}$ is set to be 2000, and the initial learning rate $\eta $ is set to be 0.01. During the model training process, the learning rate $\eta $ is scaled by 0.6 times every 200 epochs.

\paragraph{\textbf{Results analyses for two situations}} To validate the effectiveness of the Deep aPCE method, this example is also solved by the DNN, the SS-SVM method \cite{Bourinet2011}, and the SIR-PCE method \cite{Pan2017}. The MCS method's results ($10^6$ runs) are regarded as the truth results for comparison. Compared with the MCS method's results, ${\varepsilon}_{P_f} $ denote the relative estimation error rate of failure probability $P_f$. The results analyses of two situations, i.e., $n=\left\{40, 100\right\}$, are as follows:
\begin{itemize}
	\item \textbf{Situation 1:} $n=40$
\end{itemize}

For $n=40$, the uncertainty analysis results of this example are shown in Table \ref{high_dim_results}. For $N_{gd}=1200$, the relative errors of the Deep aPCE method are 0.24\% on mean, no error on standard deviation, 0.93\% on skewness and 0.04\% on kurtosis, and the relative errors of the SIR-PCE method are 0.05\% on mean, 0.01\% on standard deviation, 15.67\% on skewness and 1.39\% on kurtosis. Except that the relative error on the mean of the Deep aPCE method is slightly larger than the SIR-PCE method, the relative errors of the Deep aPCE method are less than the SIR-PCE method. Especially, the relative error on the skewness of the Deep aPCE method (0.93\%) is much smaller than the SIR-PCE method (15.67\%). Besides, the estimation error rate ${\varepsilon}_{P_f} $ of the Deep aPCE method is 2.30\% less than 8.20\% of the SIR-PCE method. Compared with the MCS method's results, the estimation accuracy of mean, standard deviation, skewness, and kurtosis increases with the number $N_{gd}$ of labeled training data increasing for both Deep aPCE method and SIR-PCE method. However, the estimation accuracy of the Deep aPCE method is higher than the SIR-PCE method. Apparently, the accuracy of the Deep aPCE method based on 1300 labeled training data is higher than the SIR-PCE method based on 2000 labeled training data. For $N_{gd}=1900$, the Deep aPCE method has already no estimation error for the failure probability $P_f$. However, the SIR-PCE method still has an 0.50\% estimation error for the failure probability $P_f$ based on 2000 labeled training data.

\begin{table}[htb]
	\centering
	\caption{The uncertainty analysis results of the high-dimensional example for $n=40$}
	\begin{tabular}{lllcllll}
		\toprule
		\multicolumn{1}{l}{Method} &  $N_{gd}$ & Mean    & Standard deviation     & Skewness     & Kurtosis & ${\varepsilon}_{Pf} $ & $R^2$ \\
		\midrule
		\multicolumn{1}{l}{MCS} &   $10^6$    &   3.7944    &   1.2634    &   $-0.0964$    & 3.0077 & \multicolumn{1}{c}{-} & \multicolumn{1}{c}{-} \\
		SIR-PCE     &   1200    &   3.7962    &   1.2527    &   $-0.1115$    & 3.0495 & $8.20\%$ & 0.981000 \\
		            &   1600    &   3.7947    &   1.2566    &   $-0.0925$    & 3.0121 & $0.50\%$ & 0.980500 \\
		            &   2000    &   3.7948    &   1.2626    &   $-0.0915$    & 3.0234 & $0.50\%$ & 0.981700\\
		\midrule       
		\multicolumn{1}{l}{\textbf{Deep aPCE}} &   1200    &  3.7854    &  1.2634    &   $-0.0973$   & 3.0089 & $2.30\%$  & 0.999611 \\
		&   1300    &  3.7943    &  1.2635    &   $-0.0967$   & 3.0079 & $0.16\%$ & 0.999984 \\
		&   1400    &  3.7938    &  1.2632    &   $-0.0962$   & 3.0062 & $\textbf{0.05\%}$ & 0.999986 \\
		&   1500    &  3.7939    &  1.2632    &   $-0.0968$   & 3.0075 & $0.27\%$ & 0.999993 \\
		&   1600    &  3.7940    &  1.2633    &   $-0.0962$   & 3.0073 & $\textbf{0.05\%}$ & 0.999992 \\
		&   1700    &  3.7944    &  1.2637    &   $-0.0962$   & 3.0065 & $0.43\%$ & 0.999993 \\
		&   1800    &  3.7942    &  1.2631    &   $-0.0964$   & 3.0074 & $0.11\%$ & 0.999995 \\
		&   1900    &  3.7943    &  1.2633    &   $-0.0963$   & 3.0080 & \textbf{0.00}     & \textbf{0.999996} \\
		&   2000    &  3.7943    &  1.2633    &   $-0.0963$   & 3.0074 & \textbf{0.00}     & \textbf{0.999996} \\
		\bottomrule
	\end{tabular}
	\label{high_dim_results}
\end{table}

According to Table \ref{high_dim_results}, the determination coefficient $ R^2 $ of the Deep aPCE method has up to 0.999611 for $N_{gd}=1200$, but the determination coefficient $ R^2 $ of the SIR-PCE method is only 0.981000. Besides, as the amounts of labeled training data increase, the determination coefficient $ R^2 $ of the Deep aPCE method is improved gradually. For $N_{gd}=2000$, the determination coefficient $ R^2 $ of the Deep aPCE method is 0.999996 which is more than 0.981700 calculated by the SIR-PCE method. Thus, the Deep aPCE method can estimate the limit state function value more accurately than the SIR-PCE method. 

\begin{itemize}
	\item \textbf{Situation 2:} $n=100$
\end{itemize}

For $n=100$, Table \ref{high_dim_results_100} shows the uncertainty analysis results. Besides, the estimations of failure probability are shown in Table \ref{failure_probability_100}. DNN uses 50000 labeled training data to construct an accurate surrogate model. However, the proposed Deep aPCE method only needs 5300 labeled training data to a more accurate surrogate model. For $N_{gd}=\left\{5400, 5500\right\}$, the accuracy of the surrogate model can be further improved. Therefore, the proposed Deep aPCE method needs only about one-tenth (a little over) of the labeled data required by the DNN to build an accurate surrogate model. According to Table \ref{failure_probability_100}, the failure probability estimated by the proposed Deep aPCE method is the most accurate in all methods. The number of labeled training data required by the SS-SVM method (6036) is close to the proposed Deep aPCE method (5300). However, the relative estimation error rate ${\varepsilon}_{P_f}$ of the former is 1.88\%, which is more than 0.11\% of the latter. Although the SIR-PCE method only needs 3000 labeled training data to build a surrogate model, its relative estimation error rate ${\varepsilon}_{P_f}$ is 9.94\%, far greater than 0.11\% of the proposed Deep aPCE method. In summary, the proposed Deep aPCE method can directly construct accurate surrogate models of the high dimensional stochastic systems without complex dimension-reduction and model decomposition operations.

\begin{table}[htb]
	\centering
	\caption{The uncertainty analysis results of the high-dimensional example for $n=100$}
	\begin{tabular}{llllll}
		\toprule
		\multicolumn{1}{l}{Method} &  $N_{gd}$ & Mean (R.E.)    & S.D. (R.E.)     & Skewness (R.E.)     & Kurtosis (R.E.) \\
		\midrule
		\multicolumn{1}{l}{MCS} &   $10^6$    &   6.0000    &   2.0036    &   $-0.0642$    & 3.0104 \\
		DNN     &   50000    &   5.9995 (0.0081\%)    &   2.0035 (0.0070\%)    &   $-0.0620$ (3.4594\%)    & 3.0076 (0.0948\%) \\
		\multicolumn{1}{l}{\textbf{Deep aPCE}} &   5300    &  5.9999 (0.0013\%)    &  2.0034 (0.0119\%)   &   $-0.0637$ (0.82492\%)  & 3.0102 (0.0071\%) \\
		&   5400    &  6.0000 (0.0093\%)   &  2.0037 (0.0022\%)    &   $-0.0643$ (0.1721\%)   & 3.0104 (0.00173\%) \\
		&   5500    &  6.0000 (0.0016\%)   &  2.0038 (0.0081\%)    &   $-0.0643$ (0.0940\%)   & 3.0104 (0.0008\%) \\
		\bottomrule
		\multicolumn{6}{l}{R.E. = Relative error. \qquad S.D. = Standard deviation.} \\
	\end{tabular}
	\label{high_dim_results_100}
\end{table}

\begin{table}[htb]
	\centering
	\caption{The estimation of failure probability in the high-dimensional example for $n=100$}
	\begin{tabular}{llllll}
		\toprule
		\multicolumn{1}{l}{Method} & MCS  &   DNN   & SS-SVM & SIR-PCE & Deep aPCE  \\
		\midrule
		$N_{gd}$ &   $10^6$    &  50000  &  6036  &  3000 & 5300    \\
		$P_f$    &  $1.81\times10^{-3}$  &  $1.77\times10^{-3}$  &  $1.74\times10^{-3}$  & $1.63\times10^{-3}$ &  $1.81\times10^{-3}$ \\
		${\varepsilon}_{P_f}$    & - &  1.88\% & 3.87\%  &  9.94\% & 0.11\% \\
		\bottomrule
	\end{tabular}
	\label{failure_probability_100}
\end{table}

\begin{figure}[!htb]
	\centering
	\subfigure [5300 labeled training data]
	{\includegraphics[scale=0.58]{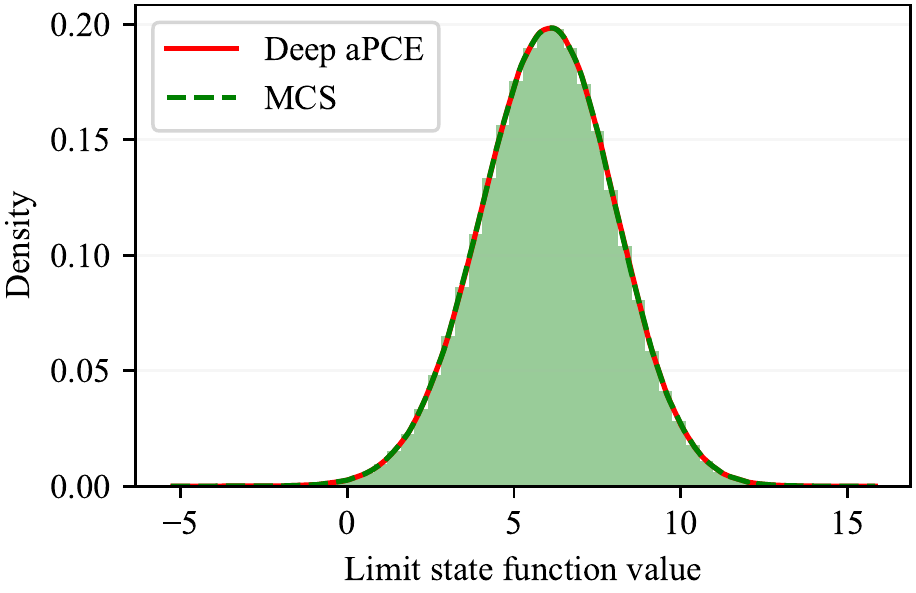}}
	\subfigure [5400 labeled training data]
	{\includegraphics[scale=0.58]{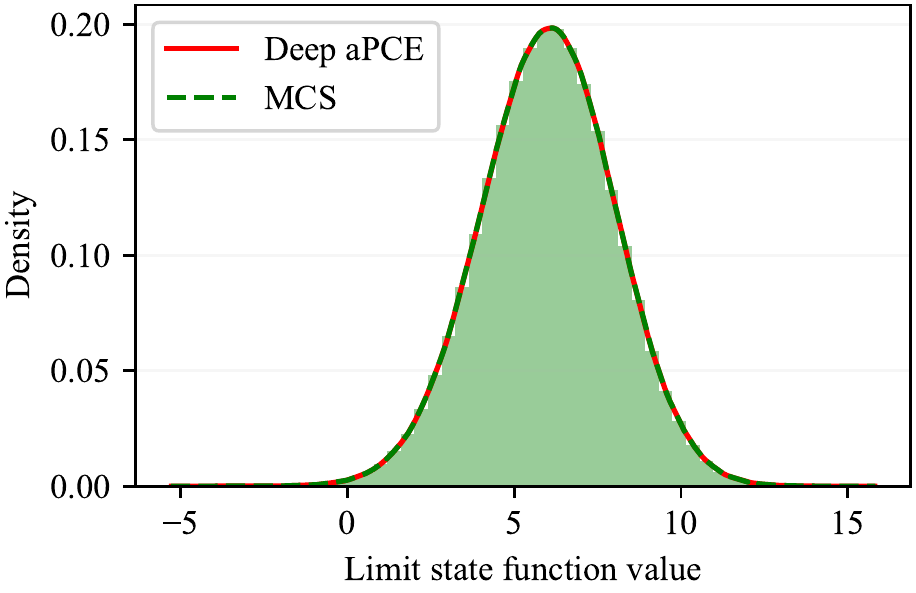}}
	\subfigure [5500 labeled training data]
	{\includegraphics[scale=0.58]{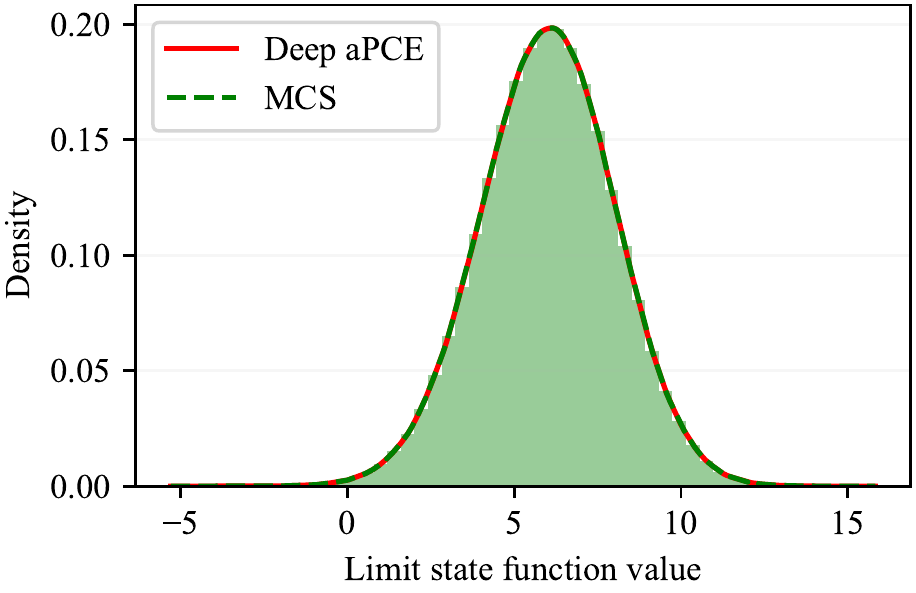}}
	\caption{The estimated probability density functions of the limit state function value by the Deep aPCE method (1200, 1300, and 1400 labeled training data) and the MCS method for the high-dimensional example ($n=100$).}\label{Rackwitz_pdf}
\end{figure}

To show the accuracy of the Deep aPCE method more intuitively, based on KDE, the estimated PDFs of limit state function value by the Deep aPCE method (5300, 5400, and 5500 labeled training data) and the MCS method are shown in Fig.\ref{Rackwitz_pdf}. The red curves (solid line) are the results of the Deep aPCE method, and the green curves (dash line) are the results of the MCS method. According to Fig.\ref{Rackwitz_pdf}, the PDF curves of limit state function value by the Deep aPCE method are very closer to the MCS method's results for different amounts of labeled training data.

\subsection{Example 5: Multimodal probability distribution example}
\begin{figure}[!htb]
	\centering
	{\includegraphics[scale=1.3]{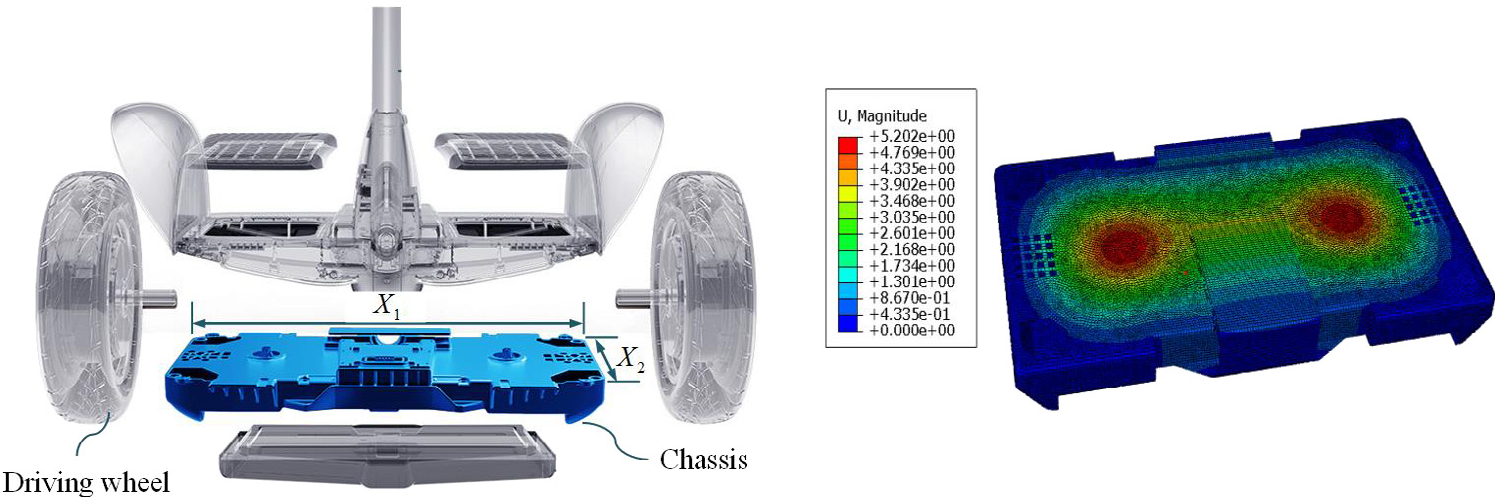}}
	\caption{The chassis deformation analysis of a balance vehicle \cite{Meng2020}}\label{B_car}
\end{figure}

\begin{figure}[!htb]
	\centering
	\subfigure [Length $X_1$]
	{\includegraphics[scale=0.75]{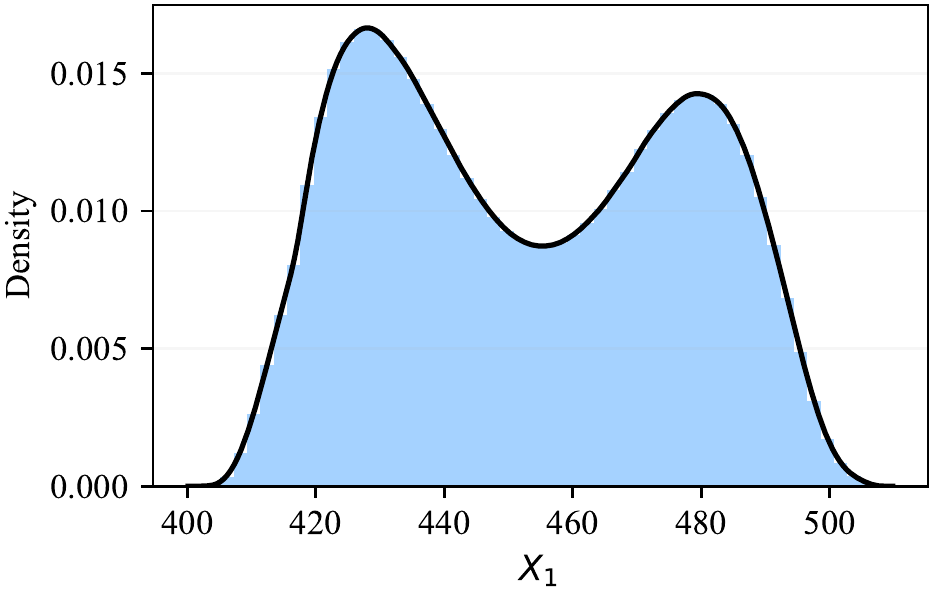}\label{X1}}
	\subfigure [Width $X_2$]
	{\includegraphics[scale=0.75]{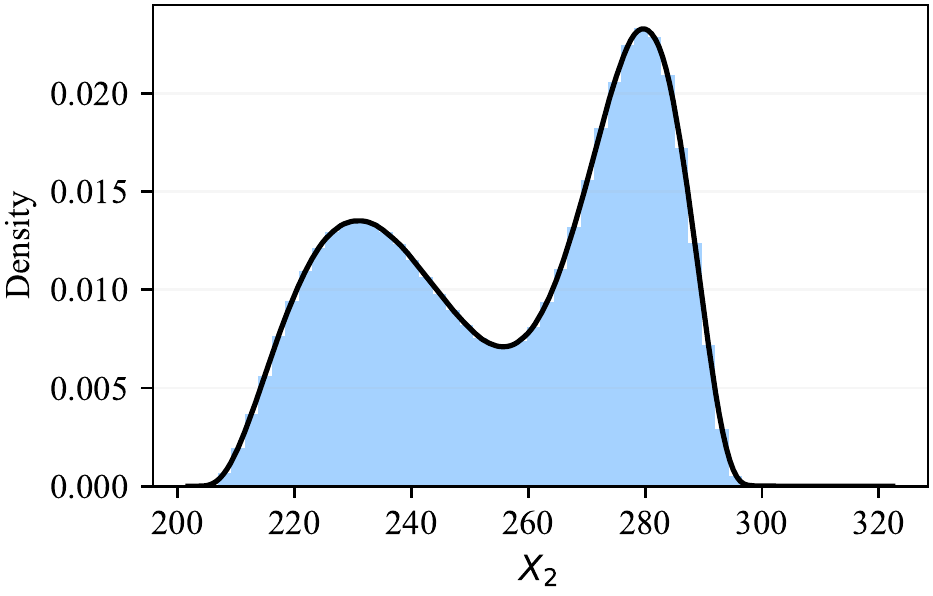}\label{X2}}
	\subfigure [Material elastic modulus $E$]
	{\includegraphics[scale=0.75]{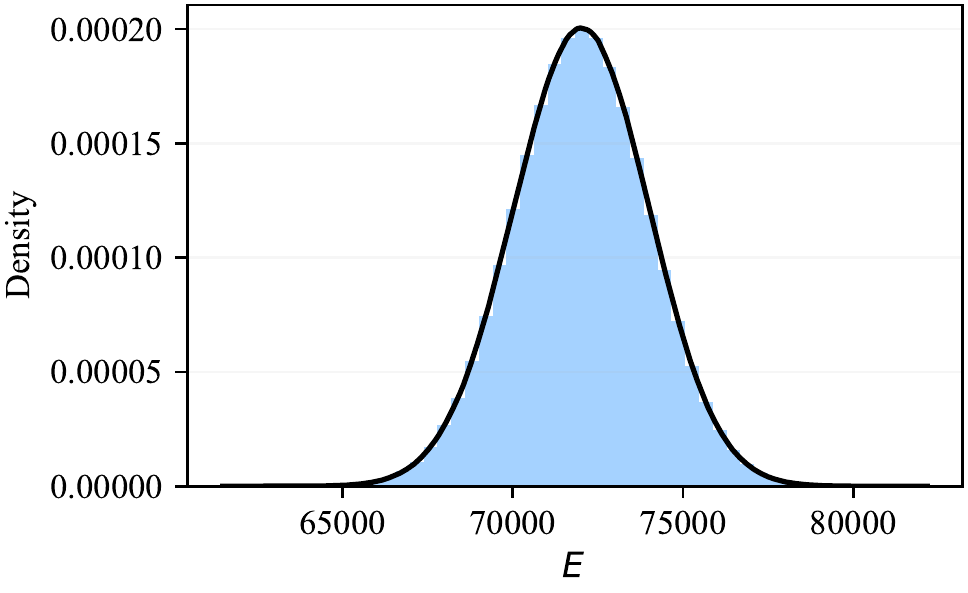}\label{E}}
	\subfigure [Horizontal load $F$]
	{\includegraphics[scale=0.75]{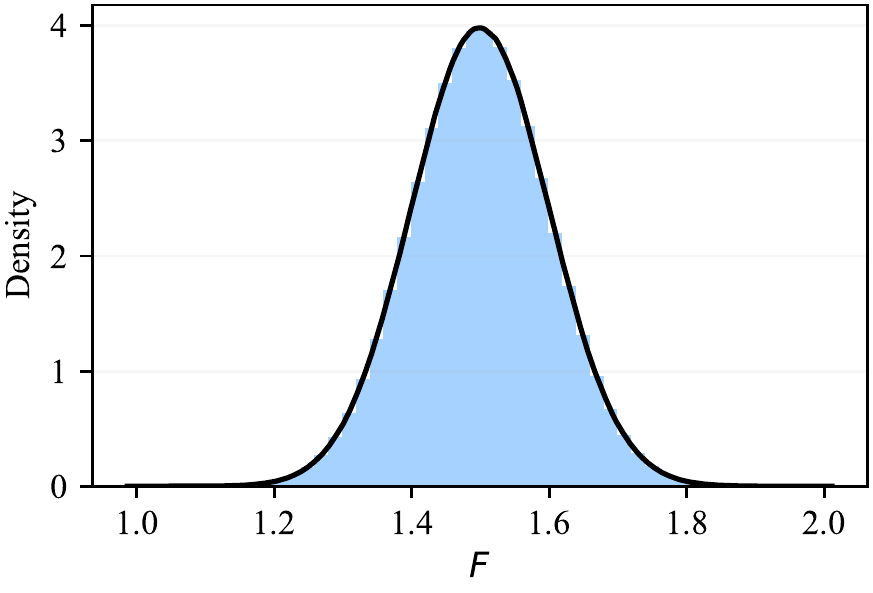}\label{F}}
	\caption{The probability density functions of the balance car's four random variables.}\label{four_variables}
\end{figure}

The fifth example is the chassis deformation analysis of a balance vehicle \cite{Meng2020} as shown in Fig.\ref{B_car}. The chassis includes four random variables, i.e., length $ X_1 $ (mm), width $ X_2 $ (mm), the material elastic modulus $ E $ (MPa), and the horizontal load $F$ (MPa), where $X_1$ and $X_2$
are multimodally distributed, and E and F are normally distributed as shown in Fig.\ref{four_variables}.

\paragraph{\textbf{Solving Deep aPCE model}} In this multimodal probability distribution example, a DNN with 4 inputs and 15 outputs is adopted to solve the adaptive expansion coefficients $\left \{ \mathcal{C}_i \left (\bm{\xi};\bm{\theta}\right )\mid i=1,2,\cdots , 15 \right \} $, and the neuron numbers of 5 hidden layers are 64, 128, 256, 128 and 64, respectively. In this example, $10^6$ unlabeled training data make up the data set ${{\mathcal{D}}_{ce}}$, and the numbers $ {N}_{gd} $ of labeled training data are 20, 30, 40, 50, and 60, respectively. Besides, the maximum training epoch $ep_{max}$ is set to be 7000, and the initial learning rate $\eta $ is set to be 0.01. During the model training process, the learning rate $\eta $ is scaled by 0.8 times every 300 epochs.

\begin{figure}[!htb]
	\centering
	{\includegraphics[scale=0.65]{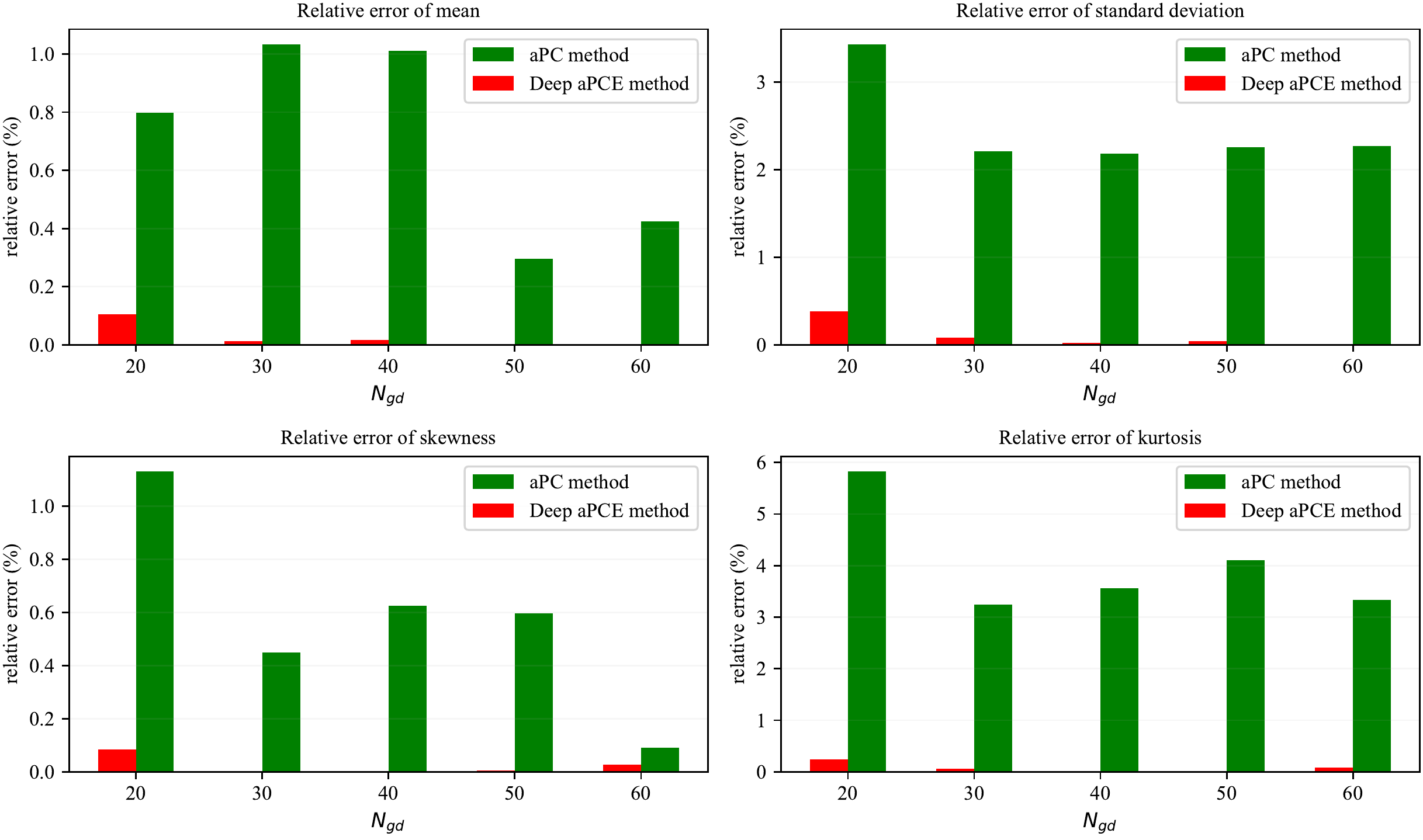}}
	\caption{Compared with the result of MC method, the relative errors of four statistical moments by the Deep aPCE method and the aPC method for $N_{gd}=20, 30, 40, 50, 60$ in the balance vehicle example.}\label{error_Bcar4}
\end{figure}

\begin{figure}[!htb]
	\centering
	{\includegraphics[scale=0.8]{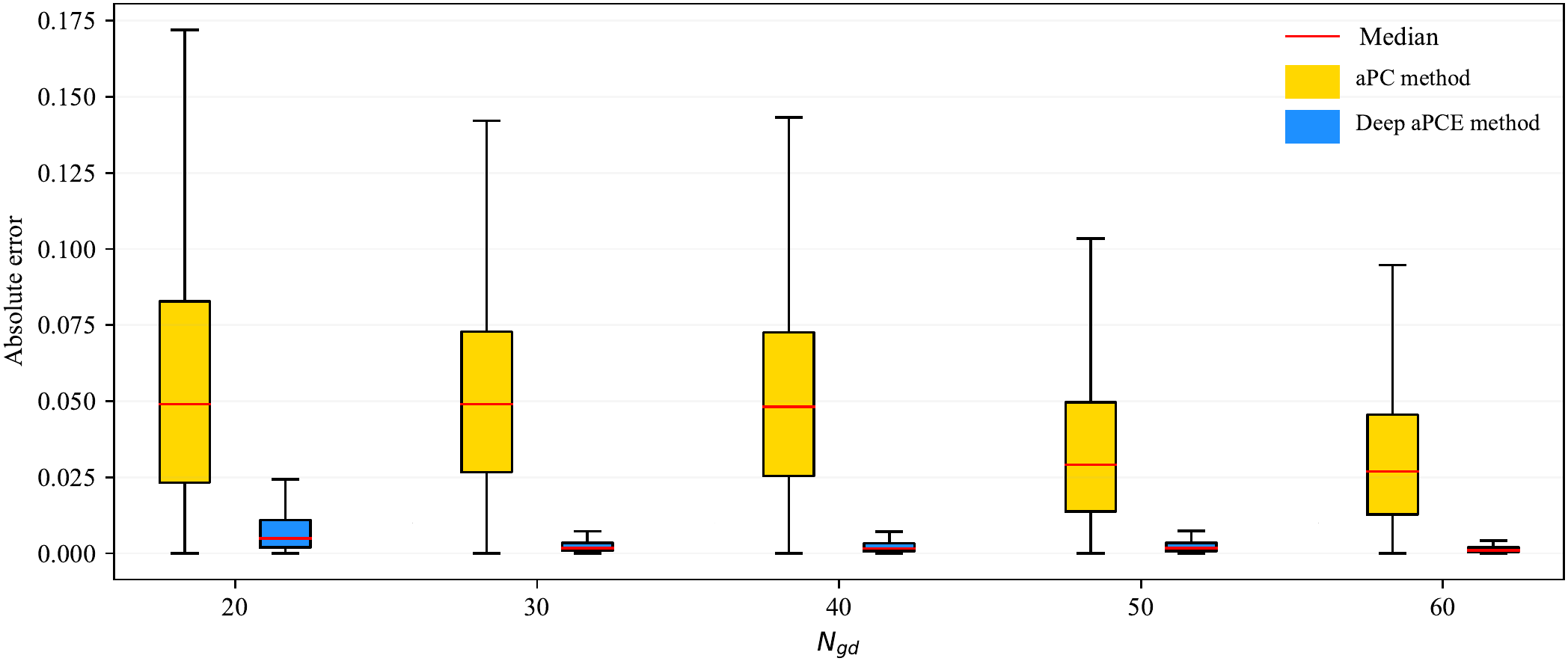}}
	\caption{The absolute error boxplots of the Deep aPCE method and the aPC method in the balance vehicle example.}\label{Error_box_Bcar}
\end{figure}

\paragraph{\textbf{Results analysis}} Based on the trained Deep aPCE model $\mathcal{D}\mathcal{A}\mathcal{P}\mathcal{C}\left( \bm{\xi };\bm{\theta } \right)$, the uncertainty analysis results of this example are shown in Table \ref{multimodal_results}. In Table \ref{multimodal_results}, the MCS method's results ($10^6$ runs) are regarded as the truth results for comparison. To validate the effectiveness of the Deep aPCE method, this example is also solved by the aPC method \cite{Oladyshkin2012}. Compared with the results of MCS, the relative errors of four statistical moments by the Deep aPCE method are smaller than the aPC method as shown in Fig.\ref{error_Bcar4}. Besides, the absolute error boxplots of the Deep aPCE method and the aPC method are shown in Fig.\ref{Error_box_Bcar}. According to Table \ref{multimodal_results}, Fig.\ref{error_Bcar4} and Fig.\ref{Error_box_Bcar}, the accuracies of uncertainty analysis results by the aPC method is not satisfactory as that by the Deep aPCE method when $N_{gd}=20, 30, 40, 50, 60$. The aPC method needs 280 labeled training data to construct a relatively accurate surrogate model of chassis deformation. However, the Deep aPCE method only needs 50 labeled training data to build a high-precision surrogate model.

\begin{table}[!htb]
	\centering
	\caption{The uncertainty analysis results of the balance vehicle's chassis deformation}
	\begin{tabular}{lllcllll}
		\toprule
		\multicolumn{1}{l}{Method} &  $N_{gd}$ & Mean    & Standard deviation     & Skewness     & Kurtosis & $e$ & $R^2$ \\
		\midrule
		\multicolumn{1}{l}{MCS} &   $10^6$    &   4.7228    &   0.6788    &   $-0.3284$    & 2.3621 & \multicolumn{1}{c}{-} & \multicolumn{1}{c}{-} \\
		aPC     &   20    &   4.7605    &   0.6555    &   $-0.3247$    & 2.2245 & 0.0148 & 0.989109 \\
		&   30    &   4.6740    &   0.6938    &   $-0.3299$    & 2.4388 & 0.0126 & 0.992094 \\
		&   40    &   4.6751    &   0.6640    &   $-0.3263$    & 2.2779 & 0.0126 & 0.992129 \\
		&   50    &   4.7368    &   0.6635    &   $-0.3264$    & 2.2653 & 0.0090 & 0.995989 \\
		&   60    &   4.7429    &   0.6634    &   $-0.3287$    & 2.2834 & 0.0082 & 0.996679 \\
		&   150   &   4.7432    &   0.6779    &   $-0.3285$    & 2.3557 & 0.0043 & 0.999078 \\
		&   280   &   4.7282    &   0.6768    &   $-0.3289$    & 2.3615 & 0.0012 & 0.999926 \\
		\midrule
		\multicolumn{1}{l}{\textbf{Deep aPCE}} &   20    &  4.7179    &  0.6812    &   $-0.3281$   & 2.3564 & 0.0058  & 0.998318 \\
		&   30    &  4.7222    &  0.6793    &   $-0.3284$   & 2.3606 & 0.0020 & 0.999796 \\
		&   40    &  4.7221    &  0.6789    &   $-0.3284$   & 2.3618 & 0.0014 & 0.999906 \\
		&   \textbf{50}    &  \textbf{4.7227}    &  \textbf{0.6788}    &   $-\textbf{0.3284}$   & \textbf{2.3618} & \textbf{0.0011} & \textbf{0.999936} \\
		&   \textbf{60}    &  \textbf{4.7228}    &  \textbf{0.6788}    &   $-\textbf{0.3285}$   & \textbf{2.3601} & \textbf{0.0006} & \textbf{0.999980} \\
		\bottomrule
	\end{tabular}
	\label{multimodal_results}
\end{table}

\begin{figure}[!htb]
	\centering
	{\includegraphics[scale=0.75]{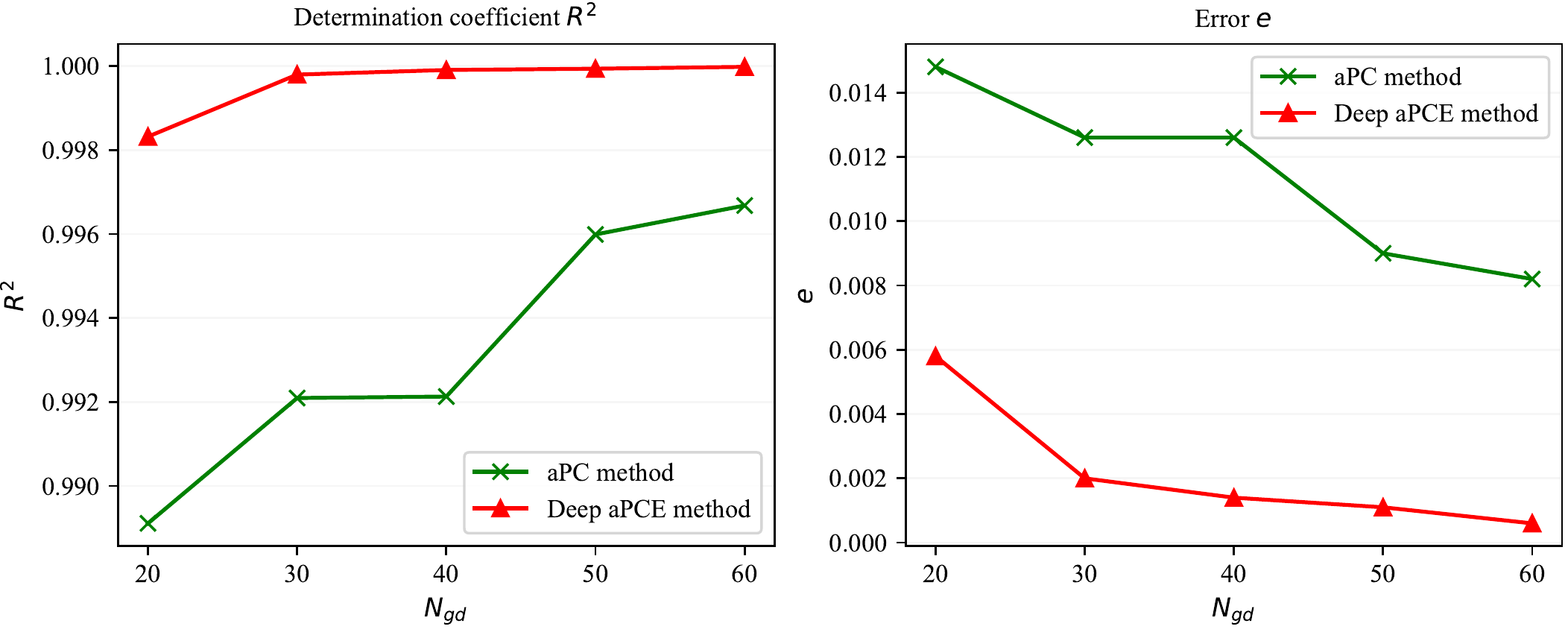}}
	\caption{The determination coefficients $R^2$ and the errors $e$ of the Deep aPCE method and the aPC method in the chassis deformation analysis of the balance vehicle example.}\label{B_car_R2e}
\end{figure}

\begin{figure}[!htb]
	\centering
	{\includegraphics[scale=0.58]{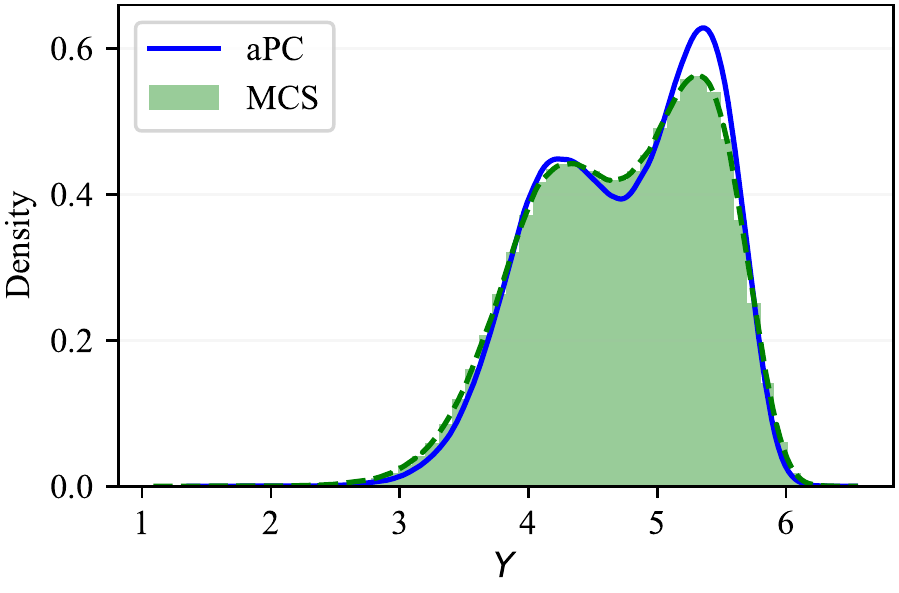}}
	{\includegraphics[scale=0.58]{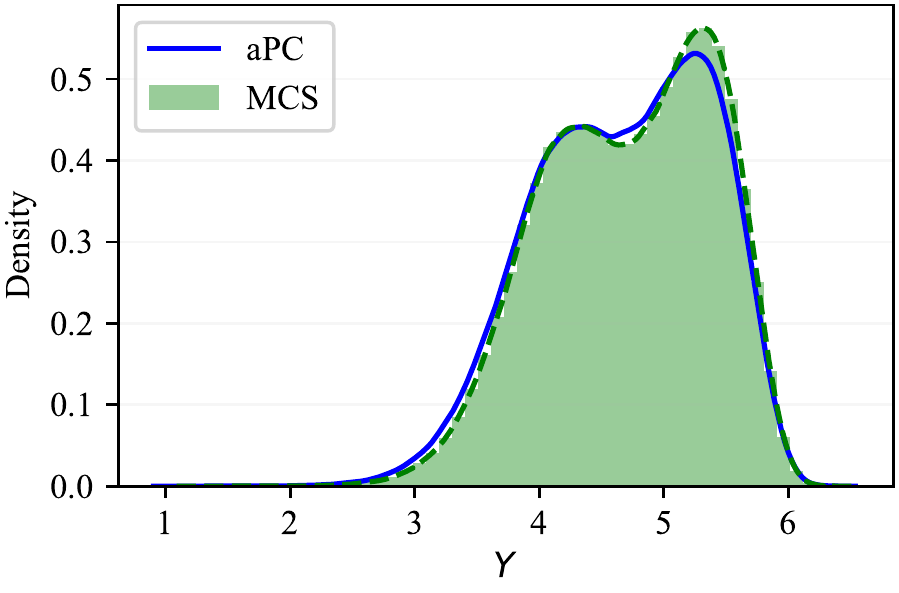}}
	{\includegraphics[scale=0.58]{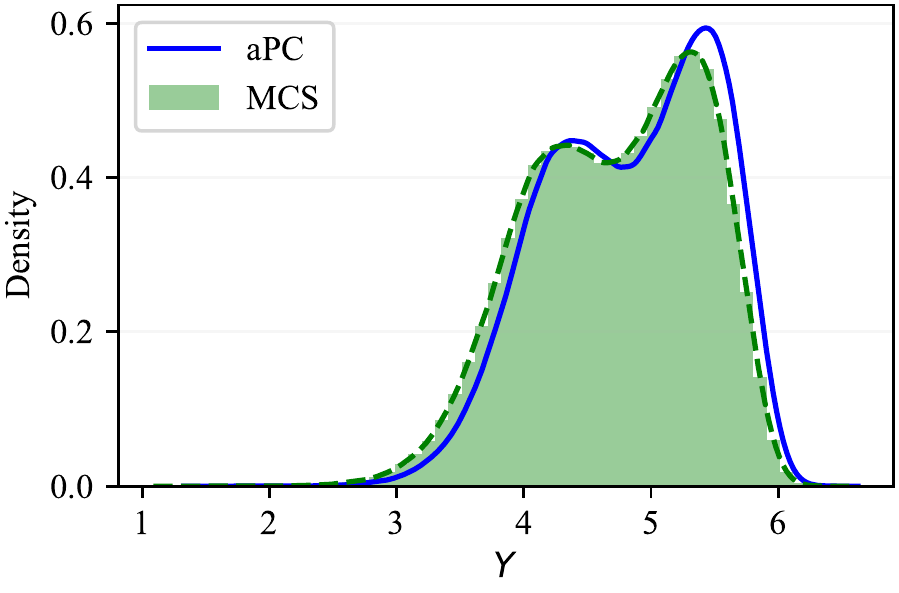}}
	\subfigure [20 labeled training data]
	{\includegraphics[scale=0.58]{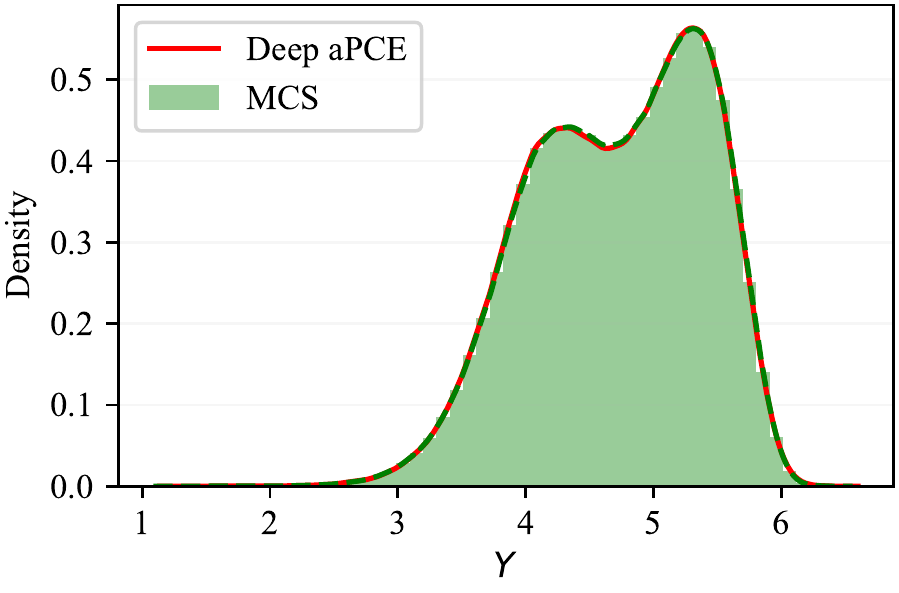}\label{BCar_20}}
	\subfigure [30 labeled training data]
	{\includegraphics[scale=0.58]{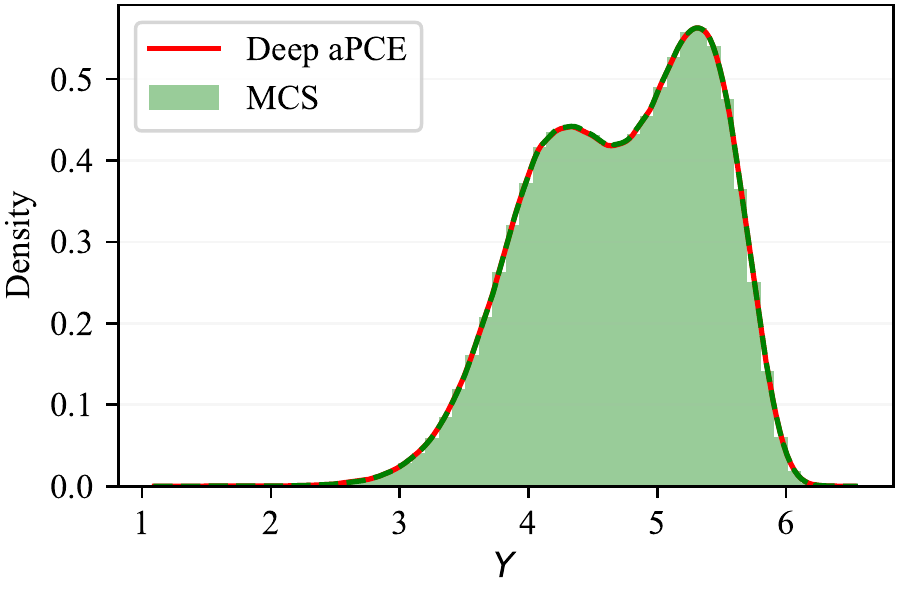}\label{BCar_30}}
	\subfigure [40 labeled training data]
	{\includegraphics[scale=0.58]{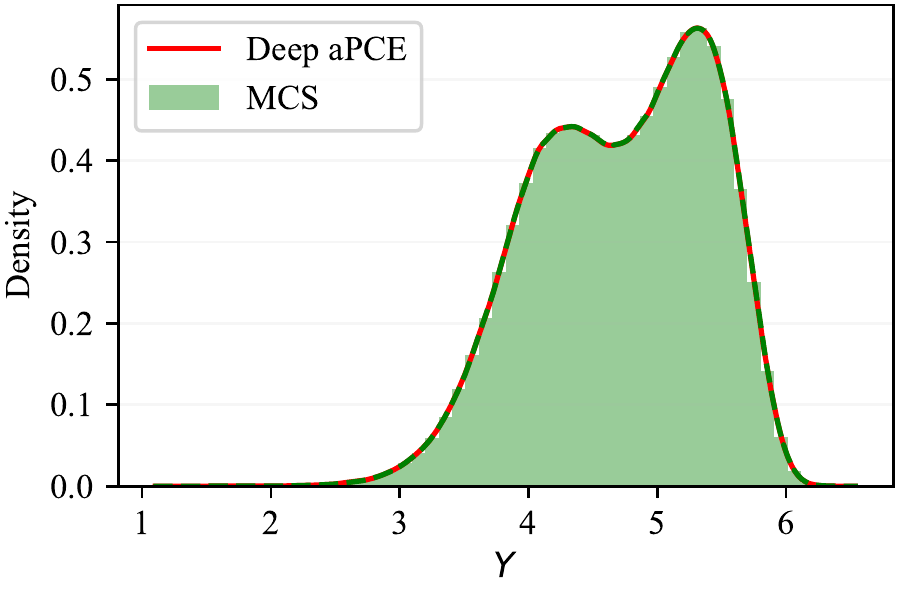}\label{BCar_40}}
	\caption{Compared with the results of MCS method, the estimated probability density functions of the horizontal deformation $Y$ by the Deep aPCE method and the aPC method (20, 30, and 40 labeled training data) for the chassis deformation analysis of a balance vehicle. The blue lines in the first row's three figures are the results obtained by the aPC method, and the red lines in the second row's three figures are the results obtained by the Deep aPCE method.}\label{BCar_pdf}
\end{figure}

Further, the determination coefficients $R^2$ and the errors $e$ of the Deep aPCE method and the aPC method are shown in Fig.\ref{B_car_R2e} for $N_{gd}=20, 30, 40, 50, 60$. Apparently, the determination coefficients $R^2$ of the Deep aPCE method are always closer to 1 than the aPC method, and the errors $e$ of the Deep aPCE method are always closer to 0 than the aPC method. Based on KDE, the estimated PDFs of the horizontal deformation $Y$ by the Deep aPCE method and the aPC method (20, 30, and 40 labeled training data) are shown in Fig.\ref{BCar_pdf}, where the green curves (dash line) are the results of the MCS method. According to Fig.\ref{BCar_pdf}, the PDF curves of the horizontal deformation $Y$ by the Deep aPCE method are closer to the MCS method's results than the aPC method.

In summary, on the one hand, with the same amount of labeled training data, the Deep aPCE method's accuracy is higher than the aPC method; on the other hand, the Deep aPCE method needs much less labeled training data to construct a high-precision surrogate model than the aPC method.

\section{Engineering application}\label{sec6}
\subsection{Engineering background}\label{sec61}
\begin{figure}[!htb]
	\centering
	{\includegraphics[scale=1.3]{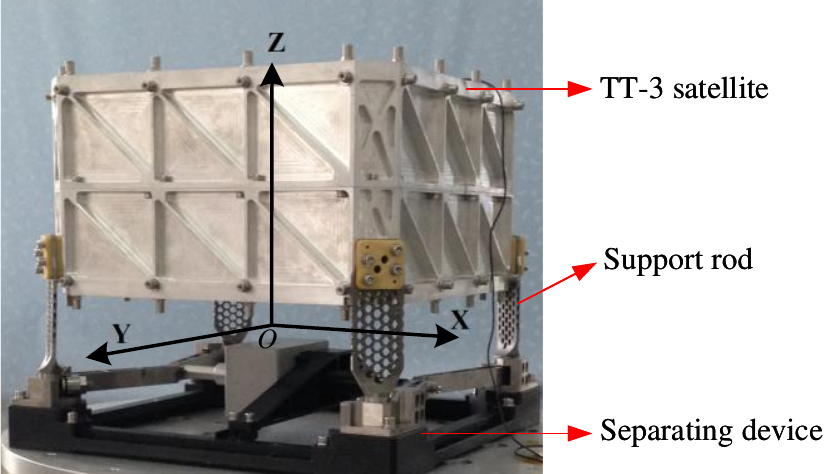}}
	\caption{Micro-satellite TT-3 frame structure}\label{satellite_frame}
\end{figure}

In this section, a practical engineering problem, i.e., the first-order frequency uncertainty analysis of the micro-satellite TianTuo-3 (TT-3) frame structure (Fig.\ref{satellite_frame}), is used to validate the effectiveness of the proposed Deep aPCE method. As shown in Fig.\ref{satellite_frame}, the TT-3 satellite frame structure includes three parts, i.e., the TT-3 satellite, four support rods and one separating device.

The TT-3 satellite frame structure's first-order frequency must be more significant than 81.0 Hz to avoid resonance during launch. For TT-3's frame structure, there are mainly six random variables (Table \ref{Param_example_engineering}) that affect the first-order frequency calculation. Due to the existence of uncertainty factors, as shown in Table \ref{Param_example_engineering}, the engineers need to quantify the first-order frequency uncertainty to guide the design of the TT-3's frame structure. Generally, a high-fidelity FEA model is used to calculate the first-order frequency. However, the FEA model has to be called many times to quantify the first-order frequency's uncertainty, leading to high computational costs. Therefore, the proposed Deep aPCE model is adopted to quantify the first-order frequency's uncertainty of TT-3's frame structure to improve the computational efficiency.

\begin{table}[htb]
	\centering
	\caption{Uncertainty parameters of the TT-3 satellite frame structure}
	\begin{tabular}{lllll}
		\toprule
		Variable & Description & True value  & Uncertainty source & V.R. \\
		\midrule
		$\rho_1$ & Aluminum alloy density & 2.69 g/$\text{cm}^{3}$ & M.M.C.  & $\pm 1\%$ \\
		$\rho_2$ & Spring steel density & 7.85 g/$\text{cm}^{3}$ & M.M.C.  & $\pm 1\%$ \\
		$\rho_3$ & Titanium alloy density & 4.43 g/$\text{cm}^{3}$ & M.M.C.  & $\pm 1\%$ \\
		$E_1$ & Aluminum elastic modulus & 68.9 GPa & E.T.C.  & $\pm 10\%$ \\
		$E_2$ & Spring steel elastic modulus & 200 GPa & E.T.C.  & $\pm 3\%$ \\
		$E_3$ & Titanium alloy elastic modulus & 113.8 GPa & E.T.C.  & $\pm 5\%$ \\
		\bottomrule
		\multicolumn{5}{l}{V.R. = the variation range. \qquad M.M.C. = the material molding conditions.} \\
		\multicolumn{5}{l}{E.T.C. = the environmental temperature change.} \\
	\end{tabular}
	\label{Param_example_engineering}
\end{table}

\subsection{Constructing surrogate model}
\begin{figure}[htb]
	\centering
	{\includegraphics[scale=0.34]{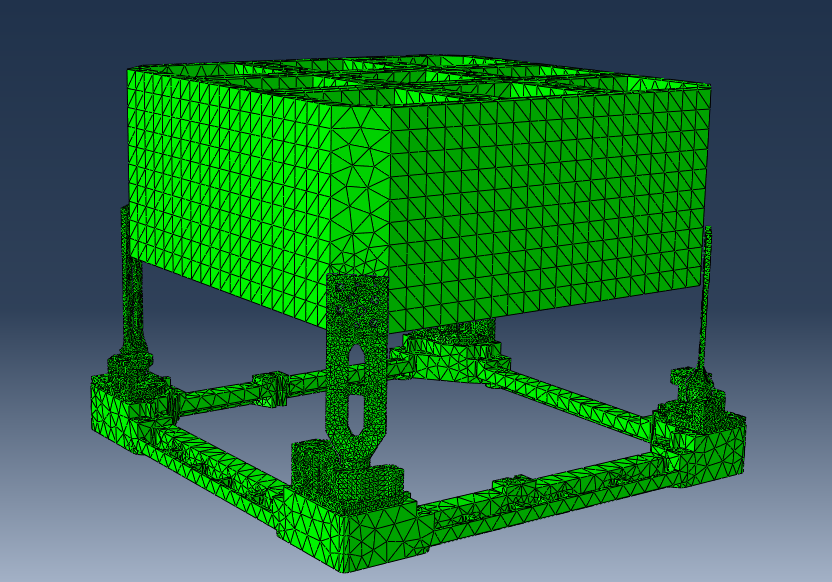}}
	\caption{The FEA model of the TT-3 satellite frame structure}\label{sat_FEA}
\end{figure}

\paragraph{\textbf{Preparing data}} According to engineering experience, each random variable is described by the normal distribution, and the corresponding mean and standard deviation are determined by '$3\sigma$' rule based on the variation range of the random variable. Thus, the means and standard deviations of six random variables can be obtained as shown in Table \ref{Param_engineering_example_cal}. Then, the values of six random variables are sampled by Latin Hypercube Sampling. Based on the ABAQUS software, the corresponding first-order frequency is calculated by the FEA model of TT-3's frame structure, as shown in Fig.\ref{sat_FEA}. There are two kinds of training data sets and a test data set, i.e., the labeled training data set ${{\mathcal{D}}_{gd}}$, the unlabeled training data set ${{\mathcal{D}}_{ce}}$ and the test data set ${{\mathcal{D}}_{test}}$. In this example, $10^5$ unlabeled training data make up the data set ${{\mathcal{D}}_{ce}}$. Besides, the number $ {N}_{gd} $ of labeled training data is 30, 40, 50, 60, and 70, respectively. Due to the computationally expensive FEA model (about 5 minutes for one single FEA simulation), 600 data are used to be the test data in this paper.

\begin{table}[htb]
	\centering
	\caption{Statistical properties of random variables in the TT-3 satellite frame structure}
	\begin{tabular}{llll}
		\toprule
		Variable & mean  & Standard deviation & Distribution \\
		\midrule
		$\rho_1$ & 2.69 g/$\text{cm}^{3}$ & $8.97\times10^{-3}$ g/$\text{cm}^{3}$  & Normal \\
		$\rho_2$ & 7.85 g/$\text{cm}^{3}$ & $2.617\times10^{-2}$ g/$\text{cm}^{3}$  & Normal \\
		$\rho_3$ & 4.43 g/$\text{cm}^{3}$ & $1.477\times10^{-2}$ g/$\text{cm}^{3}$  & Normal \\
		$E_1$ & $6.89\times10^4$ MPa & $2.29667\times10^3$ MPa & Normal \\
		$E_2$ & $2.00\times10^5$ MPa & $2.00\times10^3$ MPa & Normal \\
		$E_3$ & $1.138\times10^5$ MPa & $1.89667\times10^3$ MPa & Normal \\
		\bottomrule
	\end{tabular}
	\label{Param_engineering_example_cal}
\end{table}

\paragraph{\textbf{Deep aPCE model of TT-3 satellite frame structure}}
The random input variable $\bm{X}=\left\{ {{\rho}_{1}},{{\rho}_{2}},{{\rho}_{3}},{{E}_{1}},{{E}_{2}},{{E}_{3}} \right\}$ is normalized to be $\bm{\xi}=\left\{ {\xi }_{1}, {\xi }_{2}, {\xi }_{3}, {\xi }_{4}, {\xi }_{5}, {\xi }_{6} \right\}$. Thus, a 2-order Deep aPCE model $\mathcal{D}\mathcal{A}\mathcal{P}\mathcal{C}\left( \bm{\xi };\bm{\theta } \right)$ is used to be the surrogate model of the first-order frequency of TT-3's frame structure. In this example, a DNN with 6 inputs, 5 hidden layers and 28 outputs is adopted to solve the adaptive expansion coefficients $\left \{ \mathcal{C}_i \left (\bm{\xi};\bm{\theta}\right )\mid i=1,2,\cdots , 28 \right \} $, where the neuron numbers of 5 hidden layers are 64, 128, 128, 128 and 128, respectively. Besides, $ReLU\left(x \right)$ \cite{Nair2010, Sun2014} is chosen to be the nonlinear activation function for each hidden layers' neurons.

\paragraph{\textbf{Deep aPCE model training}} Based on the labeled training data set ${{\mathcal{D}}_{gd}}$ ($ N_{gd}=30,40,50,60,70$) and the unlabeled training data set ${{\mathcal{D}}_{ce}}$, the Deep aPCE model is trained. The maximum training epoch $ep_{max}$ is set to be 8000. Besides, the initial learning rate $\eta $ is set to be 0.05. During the model training process, the learning rate $\eta $ is scaled by 0.7 times every 300 epochs.

\subsection{Results analysis and discussion}
\begin{figure}[!htb]
	\centering
	{\includegraphics[scale=0.8]{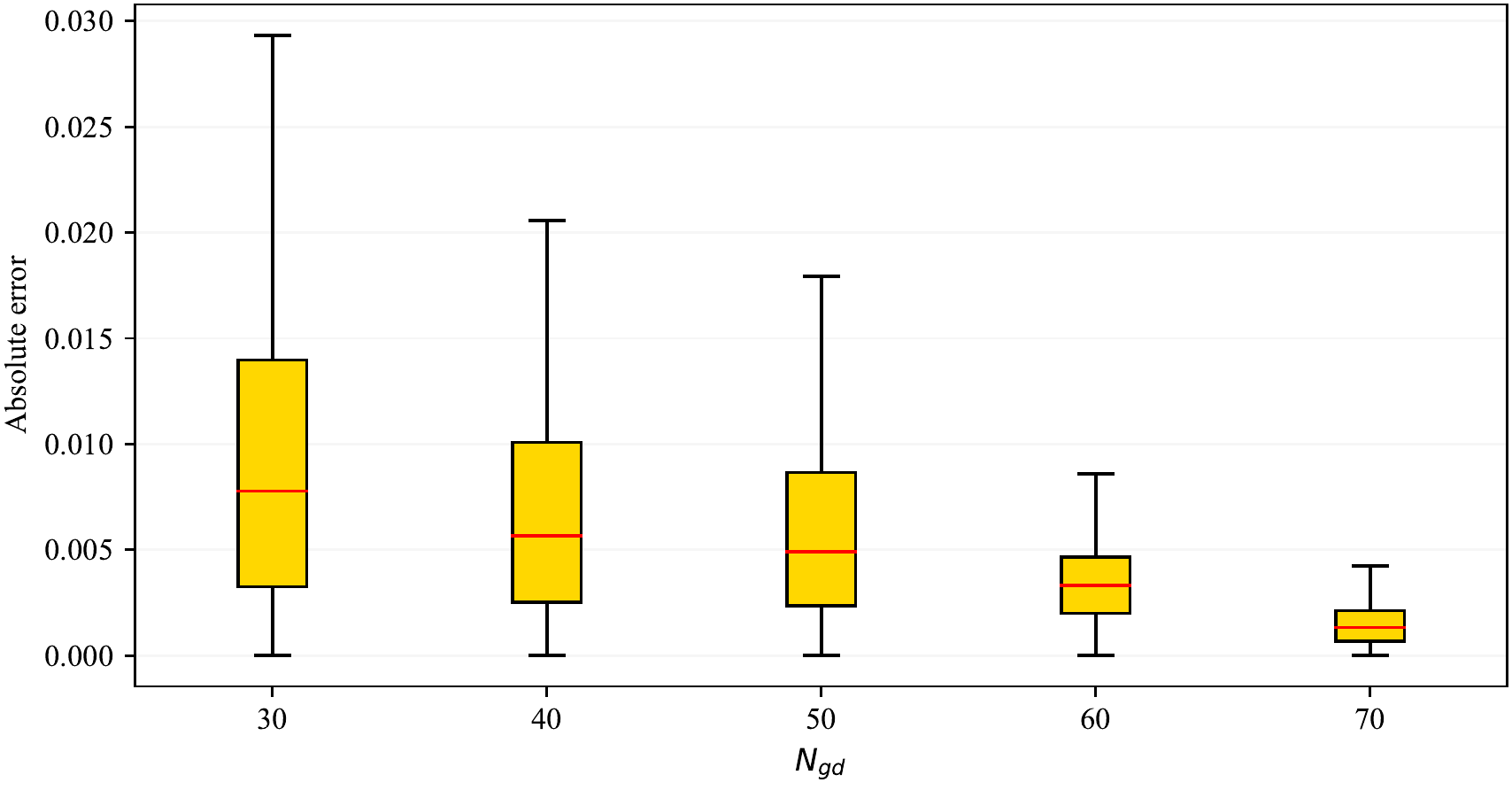}}
	\caption{The boxplots of the absolute errors between the Deep aPCE method ($N_{gd}=30,40,50,60,70$) and the FEA-MCS method in the TT-3 satellite frame structure example.}\label{Error_3}
\end{figure}

Based on the trained Deep aPCE model $\mathcal{D}\mathcal{A}\mathcal{P}\mathcal{C}\left( \bm{\xi };\bm{\theta } \right)$, the uncertainty analysis results are shown in Table \ref{sat_results}. When $N_{gd}=30$, the relative errors of the Deep aPCE method are no error on mean, 1.32\% on standard deviation, 7.71\% on skewness, and 0.038\% on kurtosis. Besides, the error $e$ and the determination coefficient $ R^2 $ are 0.00019 and 0.99881, respectively. Thus, based on 30 labeled training data, the Deep aPCE model can fit the first-order frequency model of TT-3's frame structure with only a bit of error. As shown in Table \ref{sat_results}, compared with the FEA-MCS method's results, the estimation accuracies of the Deep aPCE method's mean, standard deviation, skewness, and kurtosis increase with the number $N_{gd}$ of labeled training data. When $N_{gd}=50$, the Deep aPCE method can accurately calculate the mean, standard deviation, and kurtosis of the first-order frequency. Especially, based on 70 labeled training data, the error $e$ is only 0.00003, and the determination coefficient $ R^2 $ has arrived at 0.99998. For five Deep aPCE models ($N_{gd}=30,40,50,60,70$), the boxplot of the predictive absolute errors for 600 random inputs are shown in Fig.\ref{Error_3}, and their average absolute errors are only 0.01017, 0.00703, 0.00629, 0.00352 and 0.00165 as shown in Table \ref{sat_results}, respectively. Refer to Fig.\ref{Error_3}, the predictive absolute error decreases gradually with the number $N_{gd}$ of labeled training data increasing.

\begin{table}[!htbp]
	\centering
	\caption{The uncertainty analysis results and the average absolute errors of the first-order frequency of TT-3's frame structure}
	\begin{tabular}{lllclllll}
		\toprule
		\multicolumn{1}{l}{Method}             &  $N_{gd}$ & Mean      & Standard deviation     & Skewness     & Kurtosis & $e$ & $R^2$ & A.A.E. \\
		\midrule
		\multicolumn{1}{l}{FEA-MCS}            &   600    &   82.84    &   0.453    &   $-0.0726$    & 2.652   & -        & -       & - \\
		\multicolumn{1}{l}{\textbf{Deep aPCE}} &   30     &   82.84    &   0.447    &   $-0.0670$    & 2.651   & 0.00019  & 0.99881 & 0.01017 \\
		                                       &   40     &   82.84    &   0.447    &   $-0.0720$    & 2.650   & 0.00011  & 0.99958 & 0.00703 \\
		                                       &   50     &   82.84    &   0.453    &   $-0.0712$    & 2.652   & 0.00015  & 0.99926 & 0.00629 \\
		                                       &   60     &   82.84    &   0.453    &   $-0.0745$    & 2.653   & 0.00005  & 0.99992 & 0.00352 \\
	            	                           &   70     &   82.84    &   0.453    &   $-0.0713$    & 2.651   & 0.00003  & 0.99998 & 0.00165\\
		\bottomrule
		\multicolumn{9}{l}{A.A.E. = average absolute error. \qquad The units of mean and standard deviation are both 'Hz'.} \\
	\end{tabular}
	\label{sat_results}
\end{table}

Besides, the $k\text{-fold}$ cross-validation technique is also used to validate that the Deep aPCE model can fit the first-order frequency model of TT-3's frame structure perfectly. For $N_{gd}=40, 50,60,70$, each labeled training data set is randomly divided into 5 subsets equally. Then, the Deep aPCE model is trained by four subsets and the remaining one subset is used to test the trained Deep aPCE model, which is repeated five times. The average absolute errors of 5 models for four kinds of labeled training data sets are shown in Table \ref{AAE}. In Table \ref{AAE}, '32+8' denotes that 32 data and 8 data are respectively used to train and test the Deep aPCE model in each training, '40+10', '48+12' and '56+14' denote similar meaning. The final average absolute errors are only 0.00432, 0.00153, 0.00082, and 0.00072, respectively. In summary, the Deep aPCE model can fit the first-order frequency model of TT-3's frame structure well according to the results of Table \ref{sat_results}, Table \ref{AAE}, and Fig.\ref{Error_3}.

\begin{table}[!htb]
	\centering
	\caption{The average absolute errors of $5\text{-fold}$ cross-validation in the TT-3's frame structure example}
	\begin{tabular}{llllllc}
		\toprule
		 $N_{gd}$ & Model-1 &   Model-2   &   Model-3   &   Model-4   &   Model-5   & \textbf{Average value} \\
		\midrule
		40 (32+8)  & 0.00071  &   0.00107     &   0.00105     &   0.01783     &    0.00096    &   \textbf{0.00432} \\
		50 (40+10) & 0.00116  &   0.00371     &   0.00101     &   0.00104     &    0.00073    &   \textbf{0.00153} \\
		60 (48+12) & 0.00064  &   0.00103     &   0.00092     &   0.00080     &    0.00071    &   \textbf{0.00082} \\
		70 (56+14) & 0.00068  &   0.00054     &   0.00087     &   0.00063     &    0.00088    &   \textbf{0.00072} \\
		\bottomrule
	\end{tabular}
	\label{AAE}
\end{table}

According to the above analysis, the Deep aPCE model based on 70 labeled training data can accurately estimate the TT-3 satellite frame structure's first-order frequency. Thus, the MCS is performed $10^7$ times on the Deep aPCE model ($N_{gd}=70$). According to section \ref{sec61}, the TT-3 satellite frame structure will be damaged due to resonance if the first-order frequency is less than 81.0 Hz during launch. Based on the above $10^7$ results, the TT-3 satellite frame structure's failure probability is 0.0000151. Apparently, the failure probability is sufficiently small. Therefore, the design of the TT-3 satellite frame structure is reliable.

\begin{figure}[htb]
	\centering
	{\includegraphics[scale=0.8]{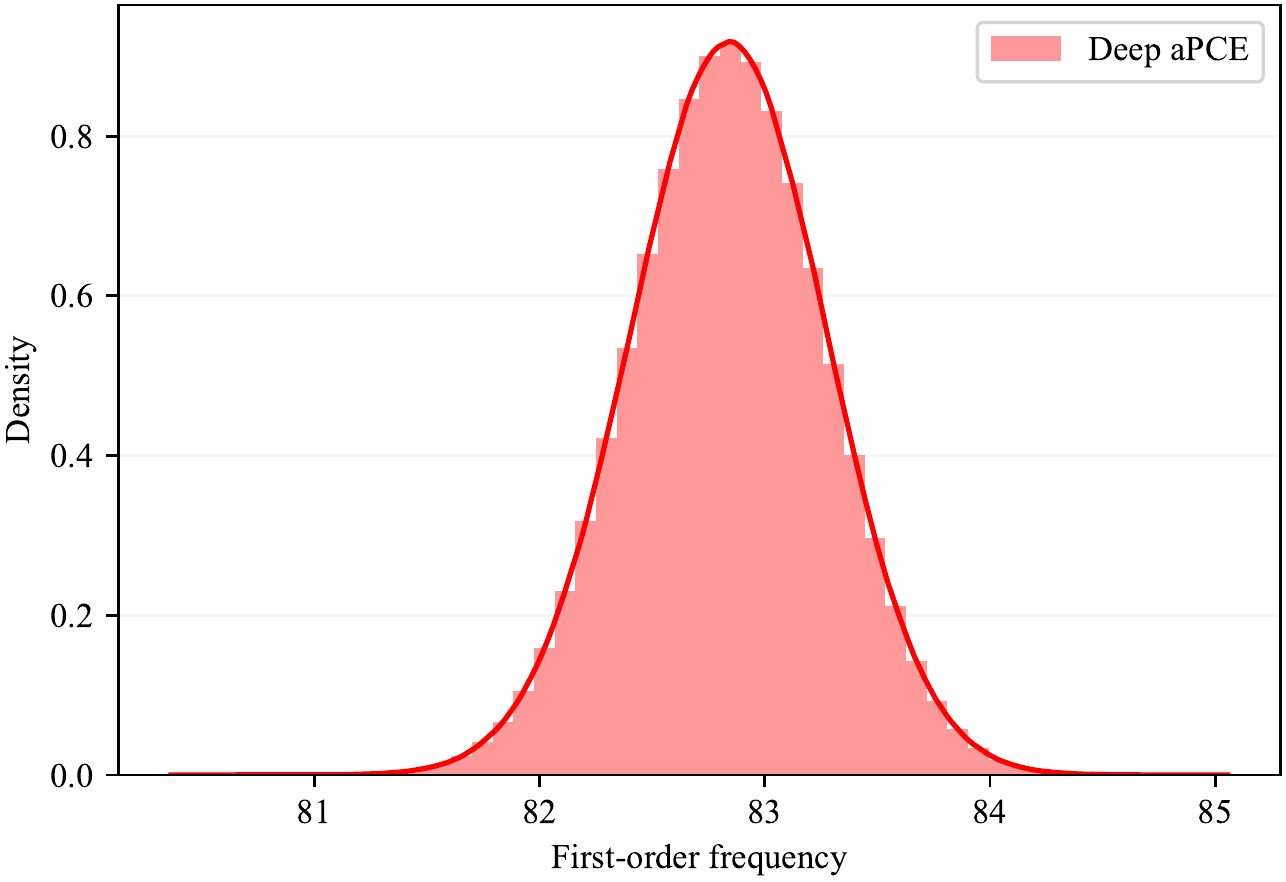}}
	\caption{The estimated probability density function of the first-order frequency of TT-3's frame structure based on the Deep aPCE model ($N_{gd}=70$).}\label{pdf70_sat}
\end{figure}

\section{Conclusions}
For quantifying the uncertainty of stochastic systems, this paper proposes the adaptive aPC and proves two properties about the adaptive expansion coefficients. Based on the adaptive aPC, a semi-supervised Deep aPCE method is proposed to reduce the training data cost and improve the surrogate model accuracy. By a small amount of labeled data and many unlabeled data, the Deep aPCE method uses two properties of the adaptive aPC to assist in training the DNN, significantly reducing the training data cost. Besides, the Deep aPCE method adopts the DNN to fine-tune the adaptive expansion coefficients dynamically, improving the Deep aPCE model accuracy with lower expansion order. Besides, the Deep aPCE method can directly construct accurate surrogate models of the high dimensional stochastic systems without complex dimension-reduction and model decomposition operations. This paper uses five numerical examples to verify the proposed Deep aPCE method's effectiveness. Using less amount of labeled training data, the accuracy of the Deep aPCE method is higher than that of the original PCE methods. Thus, the Deep aPCE method can significantly reduce the calling times of computationally expensive models and is suitable for low, medium, and high dimensional stochastic systems. Besides, the Deep aPCE method is applied to the uncertainty analysis of an actual engineering problem, and the results show that the micro-satellite TT-3 frame structure's design is reliable. For the super-high dimensional stochastic systems, both the Deep aPCE method and the original PCE methods may require more labeled training data to construct surrogate models. Therefore, the authors will explore how to reduce the number of labeled training data in constructing the Deep aPCE model of super-high dimensional stochastic systems in future research.

\section*{Acknowledgments}
This work was supported by the Postgraduate Scientific Research Innovation Project of Hunan Province (No.CX20200006) and the National Natural Science Foundation of China (Nos.11725211 and 52005505).

\appendix
\setcounter{figure}{0}
\setcounter{table}{0}
\section{The reason for choosing $L_1\text{-norm}$ rather than $L_2\text{-norm}$ to construct the proposed cost function $\mathcal{J}\left(\theta\right)$ }\label{appendix_A}
\begin{figure}[htb]
	\centering
	{\includegraphics[scale=0.8]{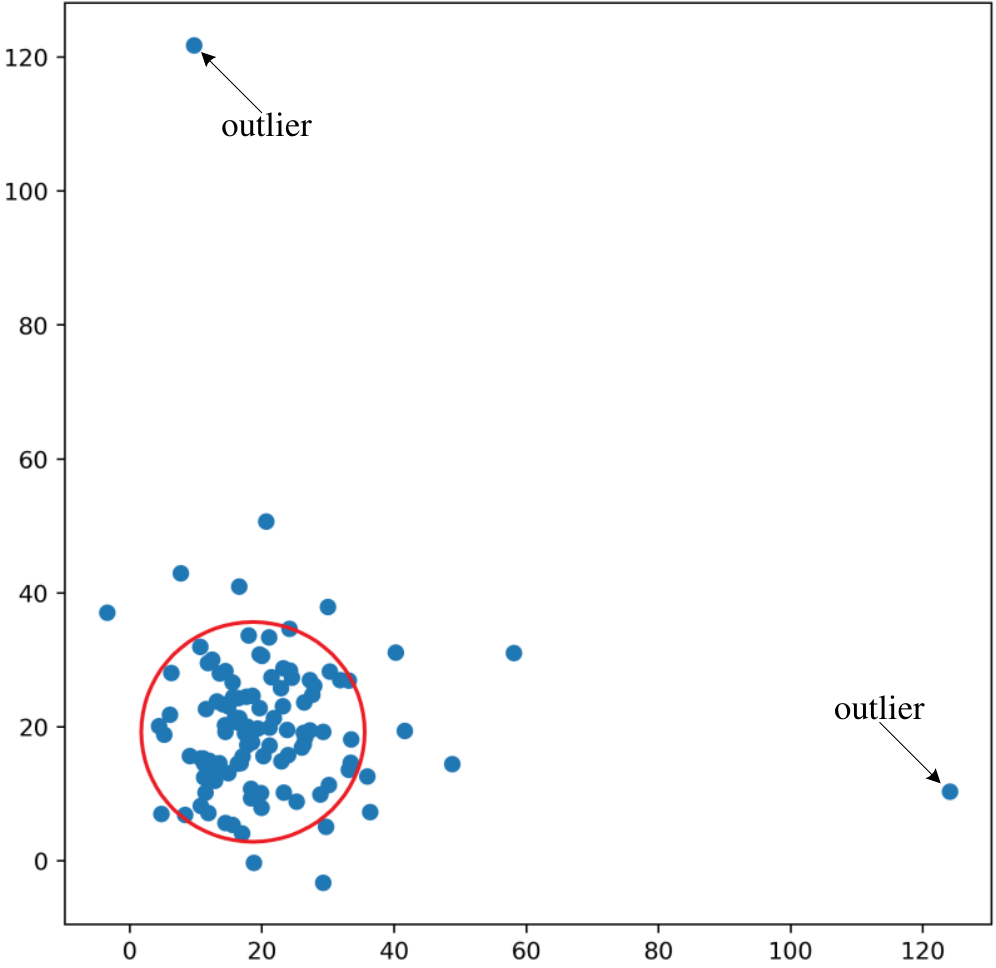}}
	\caption{The scatter plot of the displacement limit state value in the elastic cantilever beam example (Section \ref{sec52}).}\label{l1l2norm}
\end{figure}

Generally, the $L_2\text{-norm}$ is more sensitive to outliers than the $L_1\text{-norm}$. In preparing the labeled training data, the values of random input variables are sampled by Latin Hypercube Sampling, based on which the corresponding system response values can be obtained by the original complicated and computationally expensive model. The randomness of input variables leads to some system response values being outliers. For example, the scatter plot of the cantilever beam’s displacement limit state value is shown in Fig.\ref{l1l2norm}. Apparently, most points are concentrated in the red circle, and a few points (outliers) are distributed outside the red circle. If the $L_2\text{-norm}$ is used to measure the surrogate model prediction error in the Deep aPCE method, the outliers will reflect the surrogate model’s accuracy.

Further, the results of the elastic cantilever beam example (Section \ref{sec52}) are used to validate the above analysis. In the following experiment, the Deep aPCE model is trained by the cost function based on the $L_1\text{-norm}$ and the $L_2\text{-norm}$, respectively. The uncertainty analysis results’ relative errors are shown in Table \ref{table_l1l2}. According to Table \ref{table_l1l2}, the relative errors on mean, standard deviation, skewness, and kurtosis of the Deep aPCE model trained by the $L_2\text{-norm}$ are far less than that trained by the $L_1\text{-norm}$.

\begin{table}[htbp]
	\centering
	\caption{The relative errors of the Deep aPCE models trained by the cost function based on the $L_1\text{-norm}$ and the $L_2\text{-norm}$ in the elastic cantilever beam example.}
	\begin{tabular}{lllcll}
		\toprule
		$N_{gd}$ & Norm  & Mean    & Standard deviation & Skewness & Kurtosis \\
		\midrule
		40    &  $L_1\text{-norm}$       &   $ \textbf{0.22\%}$    &   $ \textbf{0.037\%}$    &   $ \textbf{2.16\%}$    &   $\textbf{0.80\%}$  \\
		&  $L_2\text{-norm}$  &   $ 24.47\%$    &   $ 8.32\%$    &   $ 167.56\%$    &   $ 16.58\%$ \\
		\midrule
		50    &  $L_1\text{-norm}$    &   $ \textbf{0.081\%}$    &   $ \textbf{0.22\%}$    &   $ \textbf{0.84\%}$    &   $ \textbf{0.016\%}$ \\
		&  $L_2\text{-norm}$  &   $ 16.68\%$    &   $ 20.25\%$    &   $ 83.21\%$    &   $ 18.35\%$ \\
		\midrule
		60    &  $L_1\text{-norm}$   &   $ \textbf{0.17\%}$    &   $ \textbf{0.051\%}$    &   $ \textbf{0.026\%}$    &   $ \textbf{0.19\%}$ \\
		&  $L_2\text{-norm}$  &   $ 18.74\%$    &   $ 2.99\%$    &   $ 185.52\%$    &   $ 173.20\%$ \\
		\midrule
		70    &  $L_1\text{-norm}$    &   $ \textbf{0.010\%}$    &   $ \textbf{0.30\%}$    &   $ \textbf{1.43\%}$    &   $ \textbf{0.29\%}$ \\
		&  $L_2\text{-norm}$  &   $ 8.28\%$    &   $ 0.69\%$    &   $ 16.59\%$    &   $ 10.51\%$ \\
		\midrule
		80    &  $L_1\text{-norm}$      &   $ \textbf{0.055\%}$    &   $ \textbf{0.12\%}$    &   $ \textbf{1.08\%}$    &   $ \textbf{0.0013\%}$ \\
		&  $L_2\text{-norm}$  &   $ 11.63\%$    &   $ 10.10\%$    &   $ 67.17\%$    &   $ 7.82\%$ \\
		\midrule
		90    &  $L_1\text{-norm}$      &   $ \textbf{0.018\%}$    &   $ \textbf{0.14\%}$    &   $ \textbf{0.39\%}$    &   $ \textbf{0.47\%}$ \\
		&  $L_2\text{-norm}$  &   $ 2.91\%$    &   $ 1.63\%$    &   $ 36.46\%$    &   $ 3.41\%$ \\
		\bottomrule
	\end{tabular}
	\label{table_l1l2}
\end{table}

In summary, the Deep aPCE method chooses the $L_1\text{-norm}$ rather than the $L_2\text{-norm}$ to construct the proposed cost function $\mathcal{J}\left(\theta\right)$.

\section{Multivariate index set generation method for constructing orthogonal basis}\label{appendix_B}
According to section \ref{sec22}, the univariate orthogonal basis $\left\{ \phi _{k}^{(0)}\left( {{\xi }_{k}} \right),\phi _{k}^{(1)}\left( {{\xi }_{k}} \right),\cdots ,\phi _{k}^{(p)}\left( {{\xi }_{k}} \right) \right\}$ is constructed by the raw moments of the random variable ${\xi}_k$. Then, the multi-dimensional orthogonal basis $$\left\{ {{\Phi }_{1}}\left( \bm{\xi } \right),{{\Phi }_{2}}\left( \bm{\xi } \right),\cdots ,{{\Phi }_{M}}\left( \bm{\xi } \right) \right\}$$ can be obtained by Eq.(\ref{multi_basis}). In this section, a simple multivariate index set generation method is proposed for determining the multivariate index set $ s_{i} $ of the multi-dimensional polynomial ${{\Phi }_{i}}\left( \bm{\xi } \right)$ in Eq.(\ref{multi_basis}).

Supposed that a $p$-order adaptive aPC needs to be modeled for the stochastic model $Y=f\left( \bm{\xi } \right)$ with the random input variable $\bm{\xi }=\left\{ {{\xi }_{k}}\left| k=1,2,\cdots ,d \right. \right\}$. Therefore, the number $Q$ of order combinations of all single variables $ {\xi }_{k} $ is 
\begin{equation}\label{combination_num}
Q={{\left( p+1 \right)}^{d}}.
\end{equation}
For $q=1,2,\cdots ,Q$, the $k\text{th}$ index $ \tilde{s}_{q}^{k} $ of the multivariate index set $ \tilde{s}_{q} $  is determined by
\begin{equation}\label{s_i_k}
\tilde{s}_{q}^{k}=\left\{ \begin{aligned}
& \left( q-1 \right)\bmod \left( p+1 \right) \qquad\qquad\quad k=d \\ 
& \left\lfloor \frac{q-1}{{{\left( p+1 \right)}^{d-k}}} \right\rfloor \bmod \left( p+1 \right) \qquad k=1,2,\cdots ,(d-1), \\ 
\end{aligned} \right.
\end{equation}
where $x\bmod y$ returns the remainder after division of $x$ by $y$, and $\left\lfloor x \right\rfloor $ rounds $x$ to the nearest integer less than or equal to $x$. Thus, the multivariate index set $ \tilde{s}_{q} $ can be obtained, i.e.
\begin{equation}\label{s_i_set}
{{\tilde{s}}_{q}}=\left\{ \tilde{s}_{q}^{1},\tilde{s}_{q}^{2},\cdots ,\tilde{s}_{q}^{k},\cdots ,\tilde{s}_{q}^{d} \right\}.
\end{equation}
Refer to Eq.(\ref{multi_basis}), the sum of multivariate index $ \tilde{s}_{q}^{k} $ in $ \tilde{s}_{q} $ should less than or equal to $p$, i.e.
\begin{equation}\label{si_condition}
\sum\limits_{k=1}^{d}{\tilde{s}_{q}^{k}}\le p.
\end{equation}
Therefore, only $M$ multivariate index sets meet the above conditions. Supposed that $M$ multivariate index sets are $\left\{ {{s}_{i}}|i=1,2,\cdots ,M \right\}$, where ${{s}_{i}}=\left\{ s_{i}^{k}|k=1,2,\cdots ,d \right\}$.

\bibliography{mybibfile}

\end{document}